\documentclass[Journal,SectionNumbers,10pt,InsideFigs,NoLineNumbers]{ascelike-new}
\usepackage[utf8]{inputenc}
\usepackage[T1]{fontenc}
\usepackage{lmodern}
\usepackage{graphicx}
\usepackage[figurename=Fig.,labelfont=bf,labelsep=period]{caption}
\usepackage{subcaption}
\usepackage{amsmath}
\usepackage{newtxtext,newtxmath}
\usepackage{multirow}

\usepackage{algorithm}
\usepackage{algpseudocode}
\usepackage{verbatim}
\usepackage{array}
\usepackage{listings}
\usepackage{tabularray}

\usepackage{tabularx}      % For 'X' column type
\usepackage[table]{xcolor} % For coloring cells
\usepackage{arydshln}      % For dashed lines in tables
\usepackage[colorlinks=true,citecolor=red,linkcolor=black,bookmarks=true,bookmarksopen=true,bookmarksopenlevel=1,pdfpagemode=UseOutlines]{hyperref}
\usepackage{xcolor}

\renewcommand\lstlistingname{Listing}
\renewcommand\lstlistlistingname{}

\begin{document}

% You will need to make the title all-caps
\title{Text2BIM: Generating Building Models Using a Large Language Model-based Multi-Agent Framework}

\author[1,2]{Changyu Du}
\author[1,2]{Sebastian Esser}
\author[1,2]{Stavros Nousias}
\author[1,2]{Andr{\'{e}} Borrmann}

\affil[1]{Chair of Computing in Civil and Building Engineering, Technical University of Munich, Munich, Germany.}
\affil[2]{TUM Georg Nemetschek Institute, Munich, Germany. Email: changyu.du@tum.de}

\maketitle

% Please include an abstract:
\begin{abstract}
The conventional BIM authoring process typically requires designers to master complex and tedious modeling commands in order to materialize their design intentions within BIM authoring tools. This additional cognitive burden complicates the design process and hinders the adoption of BIM and model-based design in the AEC (Architecture, Engineering, and Construction) industry. To facilitate the expression of design intentions more intuitively, we propose Text2BIM, an LLM-based multi-agent framework that can generate 3D building models from natural language instructions. This framework orchestrates multiple LLM agents to collaborate and reason, transforming textual user input into imperative code that invokes the BIM authoring tool's APIs, thereby generating editable BIM models with internal layouts, external envelopes, and semantic information directly in the software. Furthermore, a rule-based model checker is introduced into the agentic workflow, utilizing predefined domain knowledge to guide the LLM agents in resolving issues within the generated models and iteratively improving model quality. Extensive experiments were conducted to compare and analyze the performance of three different LLMs under the proposed framework. The evaluation results demonstrate that the proposed approach can effectively generate high-quality, structurally rational building models that are aligned with the abstract concepts specified by user input. Finally, an interactive software prototype was developed to integrate the framework into the BIM authoring software Vectorworks, showcasing the potential of modeling by chatting. The code is available at: \url{https://github.com/dcy0577/Text2BIM}
\end{abstract}

\section{Introduction}

Throughout the last decades, various digital representations and workflows have continuously emerged to represent the built environment. The notion of Building Information Modeling (BIM) comprises a holistic approach to reflect built assets with geometric and semantic information, which can be utilized across the entire life-cycle of a building and shared across different project stakeholders in dedicated representations \cite{Borrmann:2018:BIMIntro}. 
As tools for creating such representations, modern BIM authoring software integrates multidisciplinary design requirements, resulting in increasingly complex functions and commands. Designers must undergo extensive training to translate their design intent into software commands and complete the creation of building models. This makes the BIM design process cumbersome, complex, and challenging to learn, hindering its widespread adoption \cite{Du:2024:command_recommender}.

The recent applications of generative artificial intelligence (AI) in 3D modeling have demonstrated the potential to alleviate this additional cognitive burden through natural language-based human-machine interaction. Advanced 3D generative models like DreamFusion \cite{Poole2022DreamFusionTU} and Magic3D \cite{lin2023magic3d} can generate complex 3D models with rich textures directly from text descriptions, allowing designers to express design intent in natural language without tedious modeling commands. However, current Text-to-3D methods predominantly generate outputs based on geometric representations, such as voxels, point clouds, and meshes, which capture only the outer surface geometry of objects without modeling their possible internal contents or embedding semantic information. While existing studies focus on generating 3D geometric models for conceptual design from text, these purely geometric outputs lack the semantic richness and parametric detail required for seamless integration into BIM-based architectural design workflows.

This disconnect poses challenges: Designers cannot directly edit or modify these models in BIM authoring tools, and the absence of semantic information limits their utility in downstream tasks such as building simulation, analysis, and maintenance. Consequently, there is a clear need for (1) Research dedicated to generating 3D BIM models enriched with semantic information; (2) Development of efficient data representation format that integrates and represents the semantics, geometry, and topology of BIM models; and (3) Exploration of Large Language Models (LLMs) as an alternative generative method in 3D modeling.

Large language models (LLMs), such as generative pre-trained transformer (GPT) series \cite{openai2024gpt4technicalreport} are advanced large-scale deep neural networks pre-trained on extensive text corpora using unsupervised or semi-supervised learning strategies \cite{brown2020languagemodelsfewshotlearners}. 
These models have exhibited impressive capabilities across various natural language processing (NLP) tasks without explicit fine-tuning. 
This emergent capability \cite{wei2022emergentabilitieslargelanguage} is specifically reflected in the LLM's in-context and few-shots learning \cite{brown2020languagemodelsfewshotlearners} abilities, meaning that by carefully structuring the context of input prompts, e.g., accompanied by well-designed examples, one can customize the LLM's outputs without extensive parameter updates.
Research and techniques in this area are collectively referred to as prompt engineering \cite{10.1145/3411763.3451760}. In our field, such as automated compliance checking (ACC), many emerging studies \cite{fuchs2024usinglargelanguagemodels,YANG2024105817} have explored using these techniques for domain-specific adaptation of LLMs.

However, tailoring LLMs to generate semantic-rich 3D BIM models remains unexplored. Unlike building codes used in ACC that are intuitively suitable for LLM processing as 1D natural language, the key challenge here is designing a system capable of domain-adapting LLMs for complex 3D building modeling. Recent advances in LLM-based agent techniques \cite{Wang_2024} move beyond prompt engineering by enabling LLMs to access customized tools to interact with external environments and coordinate multiple LLMs collaboratively to perform real-world tasks beyond NLP in software engineering \cite{hong2023metagptmetaprogrammingmultiagent}, gaming \cite{xu2024surveygameplayingagents}, and societal simulations \cite{park2023generativeagentsinteractivesimulacra}, etc., providing a promising direction to address the challenge.

Inspired by these studies, this paper hypothesizes that an LLM-based multi-agent framework is capable of generating early-stage semantic-rich BIM models in sufficient quality (and reflecting user intent) from natural language instructions provided by AEC professionals.

The research questions explored by this paper can be summarized as follows:
\begin{itemize}
    \item From the algorithmic perspective, how can a multi-agent framework be designed to effectively condition and orchestrate LLMs, which were originally developed for daily conversation, to utilize domain-specific knowledge and generate meaningful 3D BIM models while alleviating hallucinations? 
    \item From the evaluation perspective, how can experiments and metrics be designed to comprehensively assess the generated BIM models?
    \item From the system integration perspective, how can software architecture be designed to facilitate the practical integration of the proposed method into engineering software?
    \item From the analytical perspective, how does the proposed method perform, what are its limitations, and which directions can future research focus on?   
\end{itemize}

To address these questions, we propose Text2BIM, an LLM-based multi-agent framework that converts natural language descriptions to 3D building models with external envelopes, internal layouts, and semantic information. By representing 3D BIM models as imperative code scripts invoking the BIM authoring software's Application Program Interfaces (APIs), the framework orchestrates multiple LLM agents to collaborate and autonomously generate executable code that ultimately produces native BIM models capable of further editing directly in the engineering software. The proposed framework utilizes specific prompts to guide LLMs in generating architectural-wise rational outcomes and automatically evaluates the generated BIM models against domain-specific building rules. It allows LLMs to iteratively improve the model quality through multiple feedback loops, incorporating domain knowledge from the rule-based model checker to autonomously resolve conflicts detected in the BIM model. A variety of automated methods and expert-based assessment were designed to comprehensively evaluate the framework's generation capabilities and the quality of the results. An interactive software prototype is implemented to integrate the proposed framework into the BIM authoring tool Vectorworks, demonstrating new possibilities for modeling-by-chatting during the design process. Finally, the research findings and directions for future work are discussed in detail.

%\begin{color}{blue}
\section{Scope}
The scope of this paper focuses on generating BIM models in the early design stage, detailing a feasibility study on the generation of reasonably laid-out, early-stage 3D buildings with a certain level of model quality from natural language descriptions, providing designers with a basis to further refine the designs in BIM authoring software.
The proposed approach aims to partially liberate designers from tedious and repetitive manual modeling and explore new ways of interacting with BIM authoring software. Designers can proceed to modify the resulting models in the BIM authoring tool at any time, ensuring a balanced level of automation and engineering autonomy. The generated BIM models only include essential building components such as interior and exterior walls, slabs, roofs, doors, and windows, along with representative semantic information like stories, spaces, and material definitions. 

The system's generative scope is limited by its available toolset. It does not aim to create building components that fall outside this toolset's capabilities, as defined in Appendix~\ref{app:tool_set}. For example, it cannot generate irregular, curved building models, nor can it produce BIM models with more engineering details required in more advanced design stages. The purpose of introducing the rule-based model checking is to improve the quality of the generated models, rather than to conduct building code compliance checks. Therefore, comprehensive consideration of building regulations is also beyond the scope of this paper. In this study, we present in detail the results of a scientific experiment. Despite the promising outcomes, we do not claim the development of a full-scale software product deployable in practice. The presented results encourage further refinements and developments by both academia and industry.

%\end{color}

% \subsection{Organization}

% The rest of the paper is organized into six sections. Section "RELATED WORK" presents the applications of generative AI in the field of BIM design and introduces background technologies. Section "METHODOLOGY" describes the proposed methodology in detail. Section "PROTOTYPE IMPLEMENTATION" presents the development and implementation of a software prototype based on the proposed methodology. Furthermore, section "EXPERIMENTS AND EVALUATION" presents the experimental evaluation process, and finally sections "DISCUSSION AND LIMITATION" and "CONCLUSION" discuss the findings and conclude this paper.

% The following part can be removed, doesn't seem to be common for this journal

% To assist with reading, we organized the paper into seven sections. Section two reviews the applications of generative AI in the field of BIM design and introduces background technologies such as LLM-based agents and BIM-based model checking. Section three presents the proposed Text2BIM method. Section four describes the implementation of the proposed framework and its integration with BIM authoring tool. Section five outlines the design of the experiments and demonstrates the validation results. The sixth section discusses and analyzes the experimental results, highlighting the limitations of the method. Finally, section seven concludes the paper and suggests potential future work.

\section{Background and related work}
In this section, we provide a comprehensive review of related work and introduce the theoretical background relevant to this study. 
Section \ref{sec:pre} introduces the LLM-relevant technical terms used in this study. 
Section \ref{sec:GenAI} reviews the latest advancements in generative AI in the field of 3D building design. Section \ref{sec:LLMagent} introduces LLM-based agents and explores their recent applications in automating and enhancing architectural design. Section \ref{sec:model_checking} presents the theoretical background of rule-based BIM model checking, which serves as the foundation for the model optimization loop and the experiments conducted in this research. Finally, Section \ref{sec:gaps} summarizes the limitations of existing work and promotes the identified research gaps.

\subsection{Preliminaries}
\label{sec:pre}
Large language models (LLMs), such as GPT-style architectures, are advanced large-scale deep neural networks designed for natural language processing (NLP) tasks. These models are trained on extensive text corpora, enabling them to develop a statistical understanding of language patterns and structures \cite{brown2020languagemodelsfewshotlearners}. By leveraging this learned representation, LLMs can perform diverse NLP tasks through autoregressive token prediction.

Prompt engineering is the practice of carefully crafting the text prompts given to an LLM to yield relevant and coherent outputs \cite{10.1145/3411763.3451760}. Because LLMs rely on contextual cues from the prompt to understand what task they should perform, prompt engineering has become essential for achieving high-quality, customized outputs without requiring extensive parameter updates to the model. Effective prompts often include clear instructions, illustrative examples, or constraints that steer the model's reasoning and output toward the desired goal \cite{10.1007/978-981-99-7962-2_30}. 
Prompt templates formalize this process by providing reusable structures where static background knowledge can be defined and context-specific information can be dynamically inserted, ensuring both consistency and efficiency \cite{schulhoff2024promptreportsystematicsurvey}. 
The Chain-of-Thought (CoT) is a special prompt engineering approach instructing an LLM to articulate its intermediate reasoning steps while generating a final answer \cite{10.5555/3600270.3602070}. This technique can enhance the reliability of model outputs, making it particularly effective for tasks requiring complex logical reasoning, such as our BIM modeling use case.
Self-reflection \cite{pan-etal-2024-automatically} in LLMs refers to a model's ability to evaluate its own generated responses, identify errors or inconsistencies, and adjust or refine its output accordingly. It typically involves generating an initial response, critiquing its correctness or coherence, and then revising it. This technique improves accuracy and reduces hallucinations.

In summary, these techniques collectively improve performance in applications built on LLMs.

\subsection{Generative AI in 3D building design}
\label{sec:GenAI}
The application of generative AI in the field of architectural design is gradually becoming a research hotspot. The key lies in constructing appropriate data representations for existing design data, experiential knowledge, and physical principles, and then training corresponding algorithms to intelligently generate new designs \cite{LIAO2024105187}. Current research and industrial applications primarily focus on generating 2D images or simple 3D volumes \cite{li2024generativeaiarchitecturaldesign}, utilizing Variational Autoencoders (VAEs), Generative Adversarial Networks (GANs), and diffusion models to create 2D architectural floor plans \cite{LUO2022104470,10204517,article}, building renderings \cite{inproceedings,buildings13071861,Archicad:2024}, architectural facade designs \cite{SUN2022108781,Zhanglei:2022}, or preliminary 3D conceptual forms \cite{XinweiZhuang:2023,Panagiota,tono22vitruvio}.
\citeN{Caparrini_2020} use connectivity vectors to represent different 3D mesh-like building geometries. By training a Variational Autoencoder (VAE) using these data, more building shapes can be generated by reconstructing interpolated positions within the learned distribution. Vitruvio \cite{tono22vitruvio} uses the occupancy field to describe the building shape by assigning binary values to each point in the 3D space, indicating whether the point is occupied by an object. The occupancy field is a continuous function learned and approximated by a modified occupancy network \cite{8953655} to represent the spatial occupancy of 3D objects, i.e., maps 3D coordinates to occupancy values. By properly encoding the sketch and the orientation into a finite latent embedding of the network, they sample shapes from the learned distribution, enabling the reconstruction of a 3D printable building mesh from a single perspective sketch. 

Other researchers are employing GANs to generate conceptual 3D buildings for the early design stage. \citeN{Panagiota} proposed using CPCGAN \cite{Yang_Wu_Zhang_Jin_2021} to generate point cloud representations of building geometries based on specific site rules. \citeN{Ennemoser:2023} decoded 3D voxels into 2D images to train a DCGAN \cite{Radford2015UnsupervisedRL}, and then used signed distance fields (SDF) \cite{Oleynikova2016SignedDF} to convert the generated images back into voxels. Although their method produces voxel models that can partially reconstruct the interior spaces of buildings, the generated results still suffer from issues such as geometric inconsistency and inaccuracy, making them far from practical architectural models. 

Overall, current research utilizing 3D generative algorithms based on pure geometric representations is still limited to generating conceptual architectural forms and has not yet been able to produce complex 3D building models with a high level of development (LOD) that exhibit both coherent exterior and interior geometry. Additionally, the results generated by these purely data-driven methods are difficult to constrain using text-based architectural rules. Better data representation methods need to be explored.

The rapidly advancing generative AI technologies, such as diffusion models and LLMs, have shown significant potential in the field of architectural design. The application of diffusion models is still primarily focused on tasks based on 2D images, such as generating architectural renderings from text \cite{li2024sketchtoarchitecturegenerativeaiaidedarchitectural} and replacing GANs to produce more robust structural designs \cite{He2023generative}. On the other hand, the application of LLMs is mainly centered on using natural language to retrieve data from BIM models \cite{ZHENG2023105067} and enhancing human-machine interaction in BIM authoring software \cite{Du:2024:Copilot_BIM,buildings14082499}. The concept of using LLMs to generate 3D building models is still in the early stages of exploration. One of the key challenges is representing 3D models as one-dimensional text data that LLMs can use \cite{LIAO2024105187}. This textual representation must concisely capture the features of the model, avoiding the verbosity caused by the overly detailed granularity typical of conventional model serialization files. \citeN{JIANG2024123431} proposed EPlus-LLM, which fine-tunes Flan-T5 to generate EnergyPlus Input Data Files (IDF) files from natural language inputs, enabling automated building energy modeling. This approach significantly reduces the manual effort required for modeling, while maintaining high accuracy and effectively handling user input errors. \citeN{JANG2024102532} proposed converting BIM models into XML format and then using LLM to process this structured text to add wall details. Finally, the modified XML is used to create more detailed wall instances which are subsequently integrated into the original BIM model in an approach that is termed "transplanting". 

The main difference from the approach presented in this paper is that here a method is introduced that enables the creation of early-stage building designs as coherent BIM models from scratch, while in \cite{JANG2024102532}, an existing, manually created BIM model is further refined. A secondary difference lies in the circumstance that the method introduced here uses imperative code invoking a BIM modeler's API instead of using declarative XML. By constructing and invoking high-level modeling APIs in the BIM authoring tool, we aim to express the geometric and semantic features of the model using the minimal and most flexible text format possible while also maximizing the benefits from the powerful code generation capabilities of LLMs.

In summary, our approach differs from the existing literature in terms of objectives and methods. We aim to generate BIM models with more information than purely geometric models derived from text descriptions. 
The literature reviewed in this section, along with their strengths and weaknesses, is summarized in Table \ref{tab:summary}.

\begin{table}[H]
\centering
\caption{Summary of Generative AI approaches in architectural design}
\label{tab:summary}
\footnotesize
\begin{tabular}{p{2.5cm} p{2.7cm} p{3.0cm} p{2.8cm} p{3.5cm}}
\hline
\hline
\textbf{Reference(s)} 
& \textbf{Approach / Algorithm} 
& \textbf{Data Representation} 
& \textbf{Task}
& \textbf{Advantages / Disadvantages}\\
\hline

\cite{LUO2022104470,10204517,article,inproceedings,buildings13071861,Archicad:2024,SUN2022108781,Zhanglei:2022}
& VAEs, GANs, or Diffusion models for 2D images 
& 2D floor plans, facade images, architectural renderings 
& 2D architectural image generation 
& \textbf{Advantages:} Quick visual results; well-developed pipelines for 2D tasks.\\
& & & & \textbf{Disadvantages:} 3D or BIM integration remains limited.\\
\hline

\cite{Caparrini_2020}
& VAE for 3D geometries 
& Connectivity vectors (mesh-like structures) 
& 3D geometric building form generation
& \textbf{Advantages:} Latent-space interpolation of building shapes.\\
& & & & \textbf{Disadvantages:} Does not guarantee interior fidelity; difficult to incorporate detailed design rules.\\
\hline

\cite{tono22vitruvio}
& Modified Occupancy Networks 
& Occupancy fields, mesh
& 3D geometric building form generation
& \textbf{Advantages:} Continuous 3D representation; reconstructs 3D meshes from sketches.\\
& & & & \textbf{Disadvantages:} Primarily conceptual forms; lacks component details like door or window.\\
\hline

\cite{Panagiota}
& CPCGAN 
& Point clouds of building geometries 
& 3D geometric building form generation
& \textbf{Advantages:} Incorporates site rules; flexible conceptual design.\\
& & & & \textbf{Disadvantages:} Low-resolution geometry; lacks interior details.\\
\hline

\cite{Ennemoser:2023}
& DCGAN + voxel/image conversion 
& 3D voxel grids, signed distance fields (SDF) 
& 3D geometric building form generation 
& \textbf{Advantages:} Can partially represent interior spaces using voxels.\\
& & & & \textbf{Disadvantages:} Prone to geometric inconsistency; not yet suitable for detailed modeling.\\
\hline

\cite{ZHENG2023105067,Du:2024:Copilot_BIM,buildings14082499}
& LLM-based retrieval and interaction 
& Text-based queries or commands 
& BIM data retrieval, human-machine interaction 
& \textbf{Advantages:} Automates queries; improves usability of BIM software.\\
& & & & \textbf{Disadvantages:} Focus on simple modeling tasks and human-machine interaction\\
\hline

\cite{JIANG2024123431}
& Fine-tuned Flan-T5 
& EnergyPlus IDF files (textual) 
& Building energy modeling 
& \textbf{Advantages:} Reduces manual input; maintains accuracy; handles user errors.\\
& & & & \textbf{Disadvantages:} Focused on energy simulation use-case\\
\hline

\cite{JANG2024102532}
& LLM for BIM model modification
& Converts BIM models to XML and processes the structured text to add wall details
& Automated wall detailing 
& \textbf{Advantages:} Structured text retains model info; facilitates rapid detailing.\\
& & & & \textbf{Disadvantages:} XML can be verbose; limited by predefined schema.\\
\hline

\textbf{Our method}
& LLM for BIM model generation
& Imperative code invoking high-level modeling APIs
& 3D BIM model generation
& \textbf{Advantages:} data representation format for BIM model that integrates geometry and semantics; Internal/external building layouts can be generated; leverages code generation.\\
& & & & \textbf{Disadvantages:} Depends on foundational LLMs and BIM tool function design.\\
\hline
\hline

\end{tabular}
\end{table}

\subsection{LLM-based agents}
\label{sec:LLMagent}

A Large Language Model-based agent refers to an autonomous system that utilizes an advanced language model to perform tasks involving perception, decision-making, and action \cite{Wang_2024}. These systems typically equip the LLM with tools to interact with the external environment, as well as memory modules to retain the thought processes, observations, and action records. 
The LLM-based agent uses the LLM as the "brain", leveraging its powerful in-context learning \cite{dong2024surveyincontextlearning} capabilities to synthesize information from various sources and deploy appropriate tools for different scenarios through linguistic reasoning, thereby enabling the system to behave, plan, and execute tasks like humans \cite{Du:2024:Copilot_BIM}. The agents can operate individually or in multi-agent systems where they collaborate, communicate, and specialize in distinct roles to solve more dynamic and complex problems using collective intelligence \cite{guo2024largelanguagemodelbased}. More specifically, an LLM-based multi-agent system is a distributed system composed of multiple agents that use LLMs as their core cognitive and decision-making mechanism. Such systems typically involve the systematic organization of multiple LLM-based agents to effectively manage communication within the framework and coordinate the agents in a structured way to achieve common goals or complete complex tasks. Large language models provide the agents with the ability to understand and generate natural language, enabling them to process unstructured data and exhibit advanced intelligent behavior in both human-machine interactions and communication between agents.

LLM-based multi-agent systems have growing applications in software development \cite{hong2023metagptmetaprogrammingmultiagent}, gaming \cite{xu2024surveygameplayingagents}, and societal simulations \cite{park2023generativeagentsinteractivesimulacra}, etc. However, their use in 3D design is still in the exploratory stage. 3D-GPT \cite{sun20243dgptprocedural3dmodeling} framework breaks down 3D modeling tasks into multiple steps, coordinating three LLM agents in a manner akin to a human team to create 3D assets in Blender that match textual descriptions. SceneCraft \cite{hu2024scenecraftllmagentsynthesizing} utilizes multiple LLM agents to convert textual descriptions into Python scripts executable in Blender, automatically rendering 3D scenes suitable for use in games and films. This system employs scene graphs to simulate spatial relationships between assets and iteratively refines scenes using visual language models. In the architecture domain, similar to the aforementioned research, \citeN{çelen2024idesignpersonalizedllminterior} enables multiple LLM agents to create scene graphs based on user instructions, and then uses a backtracking algorithm to place furniture, ultimately generating interior design scenes. \citeN{ZHANG2024114788} explore the integration of LLMs with Building Energy Modeling (BEM) software, focusing on automating and optimizing tasks such as input generation, output visualization, and knowledge extraction. A multi-agent system is developed, where a central LLM agent coordinates multiple specialized task agents to efficiently handle complex BEM tasks like generating or modifying EnergyPlus IDF. \citeN{mehta-etal-2024-improving} propose an interactive framework allowing human architects to collaborate with an LLM agent using natural language instructions to construct structures. The agent can place blocks, seek clarifications, and integrate human feedback within a Minecraft-like 3D simulation environment. \citeN{Du:2024:Copilot_BIM} integrate an LLM agent into BIM authoring software to answer software usage questions via Retrieval Augmentation Generation (RAG) \cite{10.5555/3495724.3496517} and perform simple modeling tasks based on natural language instructions. In conclusion, the existing literature within our field has only explored the application of LLM-based agents in straightforward modeling contexts. There is currently a dearth of studies exploring the potential of multi-agent systems to generate sophisticated 3D BIM models.

\subsection{BIM-based model checking}
\label{sec:model_checking}
%% From Sebastian - please review, extend, or delete
With the continuous adoption of BIM concepts in the design phase of built assets, the rich information base provides a sophisticated foundation for several downstream applications. 
The Industry Foundation Classes (IFC) \cite{ISO16739-1:2024} data model is well established to exchange digital representations of built assets comprising geometric and semantic information. 
These representations are perfectly tailored to automatically perform checks regarding the compliance of the envisioned design against various rules and guidelines. 
Such approaches have gained increasing interest from different stakeholders in the industry throughout the last few years. 
A comprehensive overview of opportunities and related challenges has been described by \citeN{Preidel2018}. \citeN{Gade2021} explored socio-technical challenges limiting the adoption of BIM-based Model Checking (BMC) in building design, using a prototype to address transparency and flexibility, and discussed its potential to improve design processes through faster and more precise building assessments.
\citeN{AMOR2021} have provided a critical assessment and discussed recent advancements in BIM-based Model Checking. 
\citeN{Eastman2009} have introduced an overall approach towards automated code compliance checking based on BIM models. 
They divide the overall checking process into four stages: rule interpretation to create machine-readable rules, building model preparation with advanced analysis, rule checking execution, and reporting of detailed defects and issues.

The rules a model should be compliant with can vary in their complexity. 
To account for this challenge, \citeN{Solihin2015} have introduced a classification system for rules, which comprises four different levels. 
% Rules are categorized into four classes: Class 1 requires minimal explicit data, Class 2 involves simple attribute derivations, Class 3 needs extensive data processing for complex evaluations like fire safety, and Class 4 demands holistic assessments that can't be easily reduced to pass/fail outcomes.
Rules assigned to class 1 require a single or a small number of explicit data to be available. 
A typical example of such a rule is the inspection of a dedicated property assigned for each element in the model. 
Class 2 rules are characterized by the derivation of simple attribute values. 
Such calculations can comprise simple arithmetic or trigonometric calculations based on geometric representations or the aggregation of semantic information.
Rules subsumed under class 3 require an extended data structure and processing. 
Such rules require a comprehensive processing of semantic and geometric data and the evaluation of intermediate calculations. 
Prominent examples of such rules can be found in the area of evaluations related to fire safety regulations. These rules often involve the assessment of material parameters of different components, geometric features, and path search algorithms to identify relevant spaces and corridors to be checked. 
Class 4 considers rules that cannot be evaluated by prescribed features but rather require a holistic evaluation of extracted information. 
In most cases, these rules consider multiple objectives, which are difficult (or even impossible) to formulate in a sequential workflow. 
Hence, it is expected that software applications support users to extract and identify relevant model information but ultimately don't provide a simple pass/fail statement at the end of a checking run. 

Besides investigations regarding the varying complexity of rules, other researchers have focused on the translation of human-readable guidelines and regulations \cite{Zhang2017,Zhou2022,Fuchs2022} and presented different methods to formulate the rules in machine-readable representations \cite{Sydora2020,Haeussler2020}. 
As a recent development, the \emph{Information Delivery Specification} (IDS) standard developed and maintained by buildingSMART International enables the specification of rules targeting basic property checks in a unified and vendor-independent manner \cite{Tomczak2022}. In its current development stage, however, it merely supports semantic information but lacks options to specify comprehensive geometric conditions. 
\citeN{Nuyts2024} have investigated different approaches to current compliance checking techniques and discussed advantages and downsides. 
As an extension to the approaches already mentioned, they also considered techniques related to linked data, which have been raised by \cite{Pauwels2011} earlier.

\subsection{Summary and identified research gaps}
\label{sec:gaps}
Numerous studies have explored the use of generative AI to create geometric representations of conceptual buildings or generate building energy models. However, the models generated by current work are either limited by a lack of semantic information and internal geometry or constrained to simple 3D volumes, thus typically only reaching LOD 100. According to the Level of Development (LOD) definition from the American Institute of Architects (AIA), the LOD 100 model typically represents the basic shape and size of elements without detailed semantic information. It is used for conceptual design to convey the overall design intent. 

The generation of semantic-rich building models with higher LOD, i.e., BIM models, has not yet been explored. Such models should incorporate geometric, topological, and semantic data, achieving at least LOD 200 (schematic design). At this level, models provide approximate quantities, dimensions, shapes, placements, and non-graphic information may also be attached to the model.
On the other hand, LLM-related research in our field primarily focuses on intuitive natural language-driven tasks such as building information retrieval, automated compliance checking, and enhanced human-computer interaction, with limited exploration of complex 3D modeling tasks.

Based on the conducted literature review (as summarized in Table \ref{tab:summary}), this paper bridges the gaps by proposing an LLM-based multi-agent framework to generate BIM models for the early design stage based on natural language instructions, ensuring model quality by employing rule-driven model checking. The proposed method is able to produce coherent, semantic-rich BIM models that can be used for (1) further manual or automated refinement steps directly in the engineering software and (2) BIM-based early-stage analysis such as clash detection, energy performance analysis, or cost estimation. This poses significant differences from geometry-only generative AI approaches.

\newpage
\section{Methodology}

% "reads like a direct application of LLMs" -> no, there is a comprehensive design of all involved agents

\begin{figure}[ht]
\centering
\includegraphics[width=\textwidth]{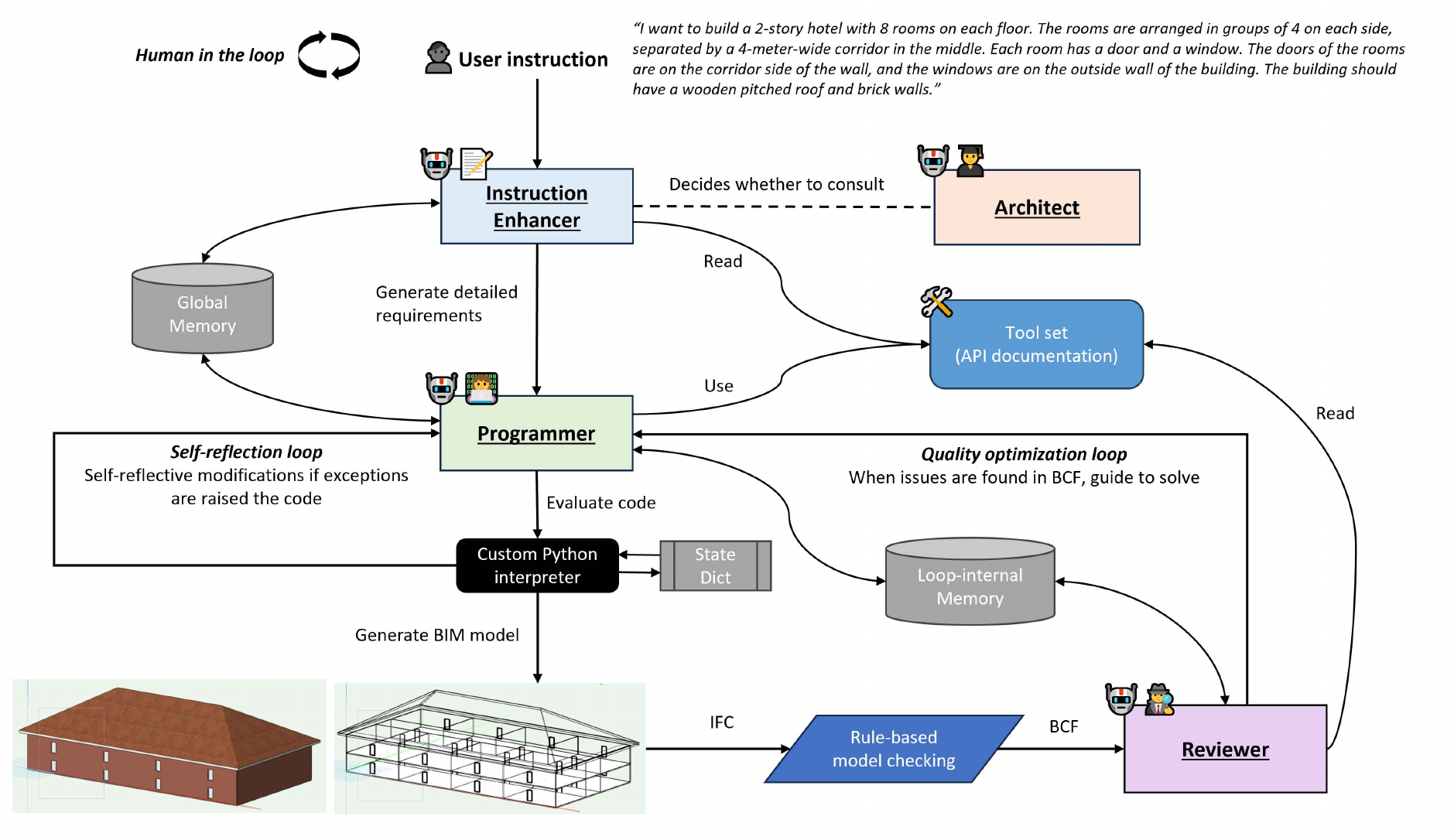}
\caption{The proposed LLM-based multi-agent framework with a sample user instruction}
\label{fig:Methodology}
\end{figure}

We propose Text2BIM, an LLM-based multi-agent framework, where four LLM agents assume different roles and collaborate to convert natural language instruction into imperative code, thereby generating building models in BIM authoring software. The core idea is to encapsulate the underlying modeling APIs of software using a series of custom high-level tool functions. By using prompt engineering techniques to guide LLMs in calling these functions within the generated code, we can construct native BIM models through a concise and efficient textual representation. 

The overall framework with a sample user instruction is shown in Fig.~\ref{fig:Methodology}. To realize the core concept outlined above, we make use of four LLM-based agents with dedicated tasks and skills that interact with each other via text:

\begin{itemize}
    \item \textsc{Instruction Enhancer}: Refines and enhances user instructions and generates detailed requirement documents.
    \item \textsc{Architect}: Develops textual building designs based on architectural knowledge.
    \item \textsc{Programmer}: Analyzes the requirements and writes code for modeling.
    \item \textsc{Reviewer}: Provides code optimization suggestions to address issues identified in the model.
\end{itemize}

The complete prompt template for each agent can be found in \textbf{Appendix~\ref{app:prompts}}. They serve as input prompts that define the agent's role, required domain knowledge, output content and format specifications, illustrative examples, as well as placeholders for dynamically inserted tasks and contextual information, etc., presented in a reusable template format.

Due to the typically brief and open-ended nature of user inputs, we first designed an LLM agent acting as a \textsc{Instruction Enhancer} to expand and refine user instruction. This ensures the instruction contains sufficient information to guide the downstream \textsc{Programmer} agent to invoke suitable tool functions in its code. The \textsc{Instruction Enhancer} agent's elaboration and detailing of the original instructions reference multiple sources. Firstly, it reads information from the custom toolset, which contains the modeling functions and the corresponding text descriptions. This allows the agent to understand whether the input parameters required for calling the corresponding functions are available within the user's instructions or if additional information is needed. Secondly, it draws on the knowledge of the \textsc{Architect} agent. When the \textsc{Instruction Enhancer} deems more architectural context or building designs necessary, especially in cases where the building requirements are complex, it can opt to consult with the \textsc{Architect}. The \textsc{Architect} agent is designed to generate building designs in a structured text format with coordinates and dimensions according to certain architectural rules, combined with the user instruction relayed by the \textsc{Instruction Enhancer}. 

The original user instruction, after being enhanced by the \textsc{Instruction Enhancer}, becomes a detailed requirement document to guide the \textsc{Programmer} agent in combining and invoking appropriate functions from the toolset to form imperative code that represents the building model expected by the user. The generated code is evaluated by a custom Python interpreter with syntax checking. If exceptions are raised during code execution, the \textsc{Programmer} will be automatically prompted to self-reflect and iteratively improve the code until errors are resolved.

The successfully generated building model is automatically exported into an IFC-based representation and then sent to a downstream model checker for automatic quality assessment. We customize a series of domain-specific rules in the checker to comprehensively evaluate model quality from various perspectives, including geometric analysis, collision detection, information verification, etc. The results of the checks are ultimately exported in BIM Collaboration Format (BCF), containing descriptions of the issues found in the model along with the associated building component GUIDs.

At this stage, a \textsc{Reviewer} agent is introduced to interpret the BCF files and provide suggestions for optimizing the model. The \textsc{Reviewer} is designed to understand the current issues in the model by reading the information recorded in the BCF files. It then proposes solutions by combining this information with the \textsc{Programmer}'s previously generated code and tool function documentation. This involves guiding and prompting the \textsc{Programmer} agent to use the appropriate tool functions to fix the code to resolve the issues present in the model. This model quality optimization loop, involving the \textsc{Reviewer}, \textsc{Programmer}, and the model checker, will iterate multiple times until the checker reports no errors or the agents are unable to resolve the issues autonomously. 

To ensure that the agents can perceive comprehensive contextual information within the loop, we implement a local loop-internal memory module to store the historical interaction between the \textsc{Programmer} and \textsc{Reviewer} during the optimization process. In addition, a global memory module shared by the \textsc{Programmer} and the \textsc{Instruction Enhancer} is designed to store the user's historical chat records and the corresponding code information. This allows the agents to have continuous contextual information during the conversation with the human, enabling the entire framework to improve responses based on human feedback. This "human-in-the-loop" design enables users to iteratively refine outputs, either by modifying input descriptions or directly editing generated BIM models, facilitating efficient, high-quality model generation and reducing manual effort. Through these multiple optimization loops, we can guide the LLM agents using domain knowledge to generate BIM models that meet certain design quality, user intentions, and engineering requirements.

In the following subsections, we describe several key modules of the proposed framework in detail.

\subsection{Toolset}
The manually defined tool functions can essentially be viewed as high-level, concise API interfaces exposed to the LLMs. Since the native APIs of BIM authoring software are usually fine-grained and low-level, each tool inherently encapsulates the logic of combining different callable API functions to achieve its functionality. This not only avoids the tediousness and complexity of low-level API calls but also incorporates specific design rules and engineering logic, ensuring the precision of the modeling tasks handled by the tool \cite{Du:2024:Copilot_BIM}. However, designing general tool functions to effectively address various building scenarios presents a significant challenge.

\begin{figure}
\centering
\includegraphics[width=0.95\linewidth]{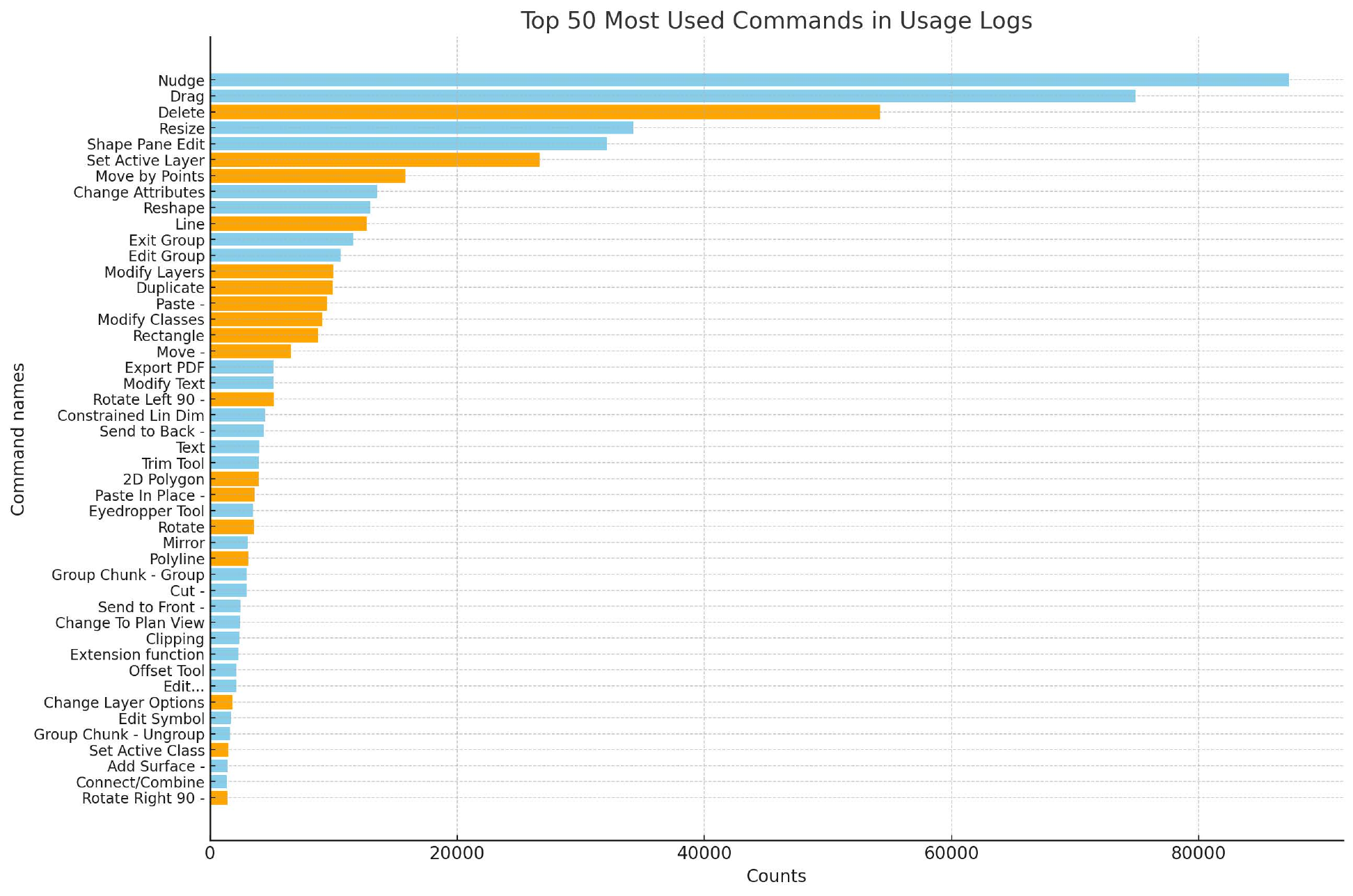}
\caption{Top 50 most used commands in BIM authoring software Vectorworks (based on collected large-scale log datasets) }
\label{fig:log}
\end{figure}

To address this difficulty, we employed both quantitative and qualitative analysis methods to determine what tool functions to implement. We began our investigation by examining user log files from BIM authoring software to understand which commands (tools) human designers most frequently use while interacting with the software. We utilized one day's log data collected from 1,000 anonymous global users of the design software Vectorworks, encompassing approximately 25 million records in 7 languages. After cleaning and filtering the raw data \cite{Du:2024:command_recommender}, the top 50 most frequently used commands are extracted, as shown in Fig.~\ref{fig:log}.

Due to the significant differences between how humans interact with software and how LLM agents interact through code, commands directly triggered by the mouse operations, such as \textit{drag, nudge}, and \textit{resize}, occupy a large proportion of the log data. Additionally, since the data is completely anonymous, we cannot determine the users' disciplines or usage scenarios. However, some typical steps in the modeling process can still be observed from this vast dataset, such as \textit{delete, move, duplicate, set working layer,} etc. We excluded mouse-dominated commands and highlighted in orange the general modeling commands that can be implemented via APIs in the chart, to serve as a reference for building tool functions for the agents.

On the other hand, we analyzed the built-in graphical programming tool Marionette (similar to Dynamo/Grasshopper) in Vectorworks. In fact, the nodes (batteries) provided by these visual scripting platforms are typically encapsulated versions of the underlying APIs tailored to different scenarios, serving as a higher-level and more intuitive programming interface for designers. Software vendors categorize the default nodes based on their functionalities, making it easier for designers to understand and use them. These motivations are similar to ours. Given our use case of creating regular BIM models, we mainly refer to the nodes under the "BIM" category.

Finally, we also considered the essential components and steps required in the typical building modeling process of architects, such as creating floors and walls, setting materials and elevation, etc. By comprehensively synthesizing information from these three perspectives, we designed a set of 26 tool functions for the agents, as shown in Table~\ref{table:tool_set_1} - ~\ref{table:tool_set3} in the \textbf{Appendix~\ref{app:tool_set}}, covering important aspects of BIM authoring such as geometric modeling and semantic enrichment. Since the LLM agents primarily understand and use the tools through their descriptions, we clearly defined the functionality, usage scenarios, input parameters, and return parameters of each function using a structured text format.

\subsection{Instruction Enhancer \& Architect: Prompt enhancement and building design generation}

The original user instruction needs to be expanded and enhanced into more specific representations aligned with architectural design. Therefore, the tasks of the \textsc{Instruction Enhancer} and \textsc{Architect} at this stage are crucial, as they determine the basic layout and quality baseline of the final generated model. The full prompt template designed for the \textsc{Architect} agent is shown in Listing~\ref{lst:arch_pt} in \textbf{Appendix~\ref{app:prompts}}. A prompt template is a structured framework that defines input instructions and context for guiding an LLM's responses by dynamically incorporating variables and consistent formatting \cite{schulhoff2024promptreportsystematicsurvey}. As shown in Listing~\ref{lst:arch_pt}, the \textit{<<task>>} indicate the dynamic user instruction that can be
inserted into the template.
Following established standard best practices in prompt engineering \cite{10.1007/978-981-99-7962-2_30}, in this template, we first define the general task and role using phrases such as "\textit{You are an experienced architect who can design floor/building based on the user's needs}" which helps improve the quality of outputs \cite{10.5555/3666122.3669274}. This is followed by descriptive instructions for output content, including essential geometric parameters for building components, such as the start and end points of walls, position definition of doors and windows, boundaries of interior rooms, etc. It also specifies formatting requirements, such as using millimeters as the unit and outputting the building design in a structured text format.
To reflect the required building design knowledge, we incorporate several basic architectural rules and principles as domain-specific context into the prompt, as shown in Listing~\ref{lst:principles}.

\begin{lstlisting}[language=TeX, caption={Basic architectural rules defined in the \textsc{Architect}'s prompt template}, label={lst:principles}]
...
Please refer to basic architectural rules, such as: 
- Foundation: Ensure a solid foundation slab that can support the entire structure.
- Wall Configuration: Arrange walls to define the building's perimeter and internal spaces. Ensure that load-bearing walls are adequately spaced and placed to distribute the weight of the structure evenly. Set the wall elevations properly for each floor.
- Slab Design: Place slabs for each floor. They should be level and supported by the walls, providing stability and separating different floors.
- Roof Construction: Design the roof to cover the entire building, protecting it from weather conditions.
- Window Placement: Install windows strategically to provide natural light and ventilation to rooms. Ensure window locations are proportionate to the room size.
- Door Placement: Position doors for easy access to different rooms and areas. Main entrance to the building should be prominent and easy to locate, with interior doors facilitating smooth movement.
- Interior Layout: Organize and define the interior room layout logically. Use interior walls to separate different functional rooms and ensure easy flow between them with appropriately placed doors.
- Structural Integrity: Ensure all elements (walls, slabs, roof) are securely connected and stable.
- Compliance: Avoid clashing/overlapping building components, such as overlapping partitions between different areas and overlapping window and door locations. Adjacent rooms can share internal partitions. Rooms can also utilize exterior walls.
...
\end{lstlisting}

These rules were developed by iterative experimental optimization following the Design Science Research (DSR) methodology \cite{vomBrocke2020}. They primarily define how various building components should be logically arranged within the context of overall building design. 
Specifically, the first rule emphasizes that the foundation slab should support the entire structure and enforce basic physical principles. Without explicit guidance on foundations, the LLM might generate a "floating" building or one that could not physically support itself. The succeeding rules establish the core structural system and the relationships between its elements: Walls define spaces and support slabs, slabs create inhabitable floors, and the roof covers the entire building to provide essential environmental protection. 

The rules for window and door placements prevent inaccessible rooms or illogical circulation patterns. The interior layout rule guides the development of functional relationships between spaces that follow logical use patterns. This guides the LLM to consider functional aspects beyond just structure. The structural integrity requirements ensure the model generates an interconnected building system rather than disconnected elements. Finally, the compliance rule prevents physically impossible configurations by providing additional context to the LLM that inherently has a limited understanding of space or structural principles.

These rules collectively provide the spatial and topological knowledge necessary for the LLM to generate coherent building designs. Experimental observations further indicate that incorporating these rules significantly enhances the compliance (quality) of the generated building designs. For example, the rule \textit{"Adjacent rooms can share internal partitions. Rooms can also utilize exterior walls."} prevents the LLM from designing duplicate partition walls between rooms. Integrating such domain-specific knowledge into the prompt template effectively guides the LLM to produce architectural-wise reasonable and structurally meaningful building designs.

Additionally, we employ the few-shots learning approach \cite{brown2020languagemodelsfewshotlearners} in the \textsc{Architect}'s prompt template to provide the LLM with an example conversation to better lead it in producing consistent, structurally accurate, and contextually appropriate building designs. As illustrated in Listing~\ref{lst:arch_pt} in \textbf{Appendix~\ref{app:prompts}}, we provide the model with a carefully designed example interaction between a hypothetical user and the \textsc{Architect} agent (termed a \textit{"sample conversation"}) as part of the prompt template, enabling the model to understand and mimic the desired format, style, and level of detail in its responses. The given sample conversation provides one illustrative example demonstrating the expected conversational pattern and content structure. The example clearly delineates how user requirements (e.g., the size and layout of a three-story office building) are translated by the \textsc{Architect} into a structured building design specifying exact measurements, layouts, and construction details of each building component. 

In addition, the example also internalizes Chain-of-Thought (CoT) \cite{10.5555/3600270.3602070} by presenting a step-by-step, structured breakdown of how the building design was derived. Specifically, the provided example demonstrated the logical sequence of the building design: 
(1) Establishing foundational dimensions. 
(2) Defining perimeter walls based on the foundation shape. 
(3) Allocating internal functional areas (rooms) according to building perimeter. 
(4) Establishing internal partition walls for rooms. 
(5) Specifying the exact placement of doors and windows on the established walls. 
(6) Specifying floor plans for higher floors.
(7) Defining slabs and roof construction.
(8) Summarizing overall building dimensions and configuration.
By including these sequential steps in the example conversation, the model is able to internalize and generalize the logical progression necessary to arrive at similar architectural solutions.

The user input (\textit{<<task>>}) in the \textsc{Architect}'s prompt template is a paraphrase of the original user instruction provided by the \textsc{Instruction Enhancer}. This is because we use the \textit{function-calling} mechanism to connect two agents: the \textsc{Architect} is wrapped in a function layer $F$, with the user task defined as the input parameter and the building design as the output. The function description documents the \textsc{Architect}'s responsibilities, and this function is the only tool the \textsc{Instruction Enhancer} can invoke. It is important to note that the tool here for function-calling is not related to our toolset designed for the code generation task. The generation of the textual building design can be summarized by the Eq.~\ref{eq:{Architect}}:

\begin{equation} \label{eq:{Architect}}
Building\_Design \leftarrow F(LLM_{\textsc{Architect}}(P_{Arch}(\alpha'))) \;
\end{equation}
Where $\alpha'$ denotes a paraphrase of the original user instruction $\alpha$, $P_{Arch}$ is the prompt template designed for the \textsc{Architect}, and $F$ is the function layer that wraps the \textsc{Architect} agent.

Most mainstream LLMs have been fine-tuned to support function-calling, enabling them to return structured JSON-formatted responses containing the function name and parameter values to be executed \cite{OpenAI_Function_Calling,gemini_function_calling}. The basic workflow of the function-calling is as follows: When the \textsc{Instruction Enhancer} receives user instructions, it infers from the input prompt context whether an external function needs to be called to complete the task. If so, it generates a JSON object, explicitly indicating the function name (\textsc{Architect}) and arguments (paraphrased user instruction). An example JSON is provided in Listing \ref{lst:json} based on the sample user instruction in Fig. \ref{fig:Methodology}.

\begin{lstlisting}[language=TeX, caption={An example JSON generated when the \textsc{Instruction Enhancer} agent consults the \textsc{Architect} agent using function-calling. This example is based on the sample user instruction in Fig. \ref{fig:Methodology}}, label={lst:json}]
{
  "id": "call_S2w3wdYq0kSzf_JZRXcpH8Wx",
  "type": "function",
  "function": {
    "name": "Architect",
    "arguments": "{\"query\":\"Design a two-story hotel with a total of 16 rooms, arranged with 8 rooms per floor. The rooms are organized in groups of four on each side of a central corridor that is 4 meters wide. Each room requires a door facing the corridor and a window on the external wall. The building should have a pitched wooden roof and brick walls. Ensure that the structure is spatially rational, using millimeter units. Provide guidance on proper placement of walls, doors, windows, and the roof to align with architectural standards.\"}"
  }
}
\end{lstlisting}

We automatically interpret the JSON locally and invoke the \textsc{Architect} function. Upon receiving the task, the \textsc{Architect} generates the corresponding building design. An example is shown in Listing \ref{lst:arch_ouptput} based on the paraphrased user instruction in the JSON in Listing \ref{lst:json}.

\begin{lstlisting}[language=TeX, caption={ An example building design in structural text format generated by \textsc{Architect} based on the paraphrased sample user instruction in the JSON in Listing \ref{lst:json}.}, label={lst:arch_ouptput}]
**Two-Story Hotel Design**
**Foundation:**
- Rectangular foundation slab: 32000mm x 16000mm
**Ground Floor Plan:**
1. **Perimeter Walls:**
- Wall A: (0,0) to (32000,0)
- Wall B: (32000,0) to (32000,16000)
- Wall C: (32000,16000) to (0,16000)
- Wall D: (0,16000) to (0,0)
2. **Functional Areas**
Boundary in format (x_min,y_min), (x_max,y_max):
- Room 1: (0,0), (8000,6000)
- Room 2: (8000,0), (16000,6000)
- Room 3: (16000,0), (24000,6000)
- Room 4: (24000,0), (32000,6000)
- Room 5: (0,10000), (8000,16000)
- Room 6: (8000,10000), (16000,16000)
- Room 7: (16000,10000), (24000,16000)
- Room 8: (24000,10000), (32000,16000)
- Central corridor: (0,6000), (32000,10000)
3. **Internal Corridor Walls:**
- Wall E: (0,6000) to (32000,6000)
- Wall F: (0,10000) to (32000,10000)
4. **Room Dividing Walls:**
- Wall G: (8000,0) to (8000,6000)
- Wall H: (16000,0) to (16000,6000)
- Wall I: (24000,0) to (24000,6000)
- Wall J: (8000,10000) to (8000,16000)
- Wall K: (16000,10000) to (16000,16000)
- Wall L: (24000,10000) to (24000,16000)
5. **Doors:**
Insertion offset of each room door relative to the start of the corresponding wall:
- Room 1 door on corridor wall E: 4000
- Room 2 door on corridor wall E: 12000
- Room 3 door on corridor wall E: 20000
- Room 4 door on corridor wall E: 28000
- Room 5 door on corridor wall F: 4000
- Room 6 door on corridor wall F: 12000
- Room 7 door on corridor wall F: 20000
- Room 8 door on corridor wall F: 28000
6. **Windows:**
Insertion offset of each room window relative to the start of the corresponding wall:
- Room 1 window on wall A: 4000
- Room 2 window on wall A: 12000
- Room 3 window on wall A: 20000
- Room 4 window on wall A: 28000
- Room 5 window on wall C: 4000
- Room 6 window on wall C: 12000
- Room 7 window on wall C: 20000
- Room 8 window on wall C: 28000
**First Floor Plan:**
- Identical to Ground Floor Plan
**Roof Construction:**
- Wooden pitched roof covering entire building: (0,0) to (32000,0) to (32000,16000) to (0,16000) to (0,0).
**Slab Design:**
- Create slabs for each floor supported by perimeter and internal walls. Slabs covering entire floor area with same dimensions as foundation.
**Summary:**
- Building dimensions: 32000mm x 16000mm x 2 floors.
- Each floor has 8 rooms, 4 on each side of a central corridor.
- Brick walls and wooden pitched roof.
\end{lstlisting}

This information is then concatenated as additional context to the \textsc{Instruction Enhancer}'s prompt, guiding it to generate the final output (detailed requirement document) for the downstream \textsc{Programmer} agent. \textbf{Appendix~\ref{app:chat_log}} provides another concrete example of the \textit{function-calling} process. The core goal of this procedure is to enable the \textsc{Instruction Enhancer} to flexibly decide whether to consult the \textsc{Architect} based on the complexity of the task, thereby making the interaction between the two agents more intelligent and efficient. This way, the \textsc{Instruction Enhancer} can work independently on simple tasks and leverage the \textsc{Architect}'s expertise for complex tasks, optimizing the overall workflow. 

The full prompt template for the \textsc{Instruction Enhancer} is shown in Listing~\ref{lst:ie_pt_main}. 
The \textsc{Instruction Enhancer}'s role is to synthesize various contextual information, including the original user instruction (\textit{<<task>>}), textual \textit{building designs} from the \textsc{Architect}, information from the toolset (\textit{<<Tool function names and descriptions>>}), and  \textit{<<chat\_history>>} from possible previous conversation context, to create a detailed and comprehensive requirements document $\alpha_{enhanced}$. This document serves as a guide for downstream code generation tasks. 

\begin{lstlisting}[language=TeX, caption={Prompt template of the \textsc{Instruction Enhancer} agent. Placeholders with "<< >>" indicate the dynamic content that can be inserted into the template, including the information on available tool functions, chat history, and the original user instruction.}, label={lst:ie_pt_main}]
You are an experienced instruction enhancer who can instruct an AI programmer to write Python scripts to accomplish user-specified tasks. You are a master at understanding API documentation, adept at breaking down the user's original instructions into subtasks and sub-logic based on the functions available, and able to express the user's description in a more rigorous and detailed manner.

Please give detailed hints on coordinates and dimensions when instructing the programmer. Consult the Architect when you feel you need a building design plan or more architectural context to assist your instructions. If functional areas are given in the design plan, make sure to instruct the programmer to create them. You must refer to architectural knowledge to ensure that your content is spatially and geometrically rational. Use millimeter units. Minimize other prose.

For reference, the following are the API functions available to the programmer, try to give hints on how to use them effectively:
"""
<<Tool function names and descriptions>>
"""

Now, based on the conversation given below, please relay the user's instruction and complete your answer. You don't need to write code, just guide the programmer step by step. Let's think step by step. 
<<chat_history>>
User: <<task>>
Instruction Enhancer: 
\end{lstlisting}

Several output specifications are defined in the prompt template to enhance the usability of the generated content, e.g., \textit{"give detailed hints on coordinates and dimensions when instructing the programmer"} and \textit{"breaking down the user's original instructions into subtasks and sub-logic based on the tool functions available"}, etc. We also employ the Chain-of-Thought (CoT) technique to further improve the output quality of this highly logic-intensive task. Unlike providing an example in the \textsc{Architect}'s prompt template where intermediate reasoning steps are listed, we simply append \textit{"Let's think step by step:"} to the end of the prompt to achieve a similar effect, as observed by \cite{zhang2023automatic}.
The CoT encourages the LLM to articulate intermediate reasoning steps explicitly while generating the answer \cite{10.5555/3600270.3602070}. Given that the model generates each token autoregressively based on preceding tokens, this method essentially supplies additional internal context during the generation, making it particularly effective for tasks requiring complex logical reasoning.

The primary objective of the \textsc{Instruction Enhancer} is to decompose the BIM modeling task into a series of subtasks and logical segments, thereby providing structured guidance for code generation. BIM modeling task is inherently a systematic, sequential process that requires complex logical reasoning. For example, the process typically begins with the creation of the foundation slab, followed by modeling the exterior walls based on the configuration of the slab. Subsequently, the interior room layout is planned based on the exterior boundaries. Partition walls are then positioned within the rooms. Doors and windows are then placed according to the start and end points of the walls, with the procedure continuing progressively for successive floors. Therefore, the CoT can enhance both the reliability and accuracy of the generated guidance.

Formally, the whole process of user prompt enhancement can be expressed by the following Eq.~\ref{eq:prompt_enhancement}:
\begin{equation} \label{eq:prompt_enhancement}
\alpha_{enhanced} \leftarrow LLM_{InstructionEnhancer}(P_{IE}(\alpha, T, \Phi_{global}) \mid Building\_Design ) \;
\end{equation}
where $P_{IE}$ denotes the prompt template for the \textsc{Instruction Enhancer}, $\alpha$ denotes the user's original instructions, $T$ is the descriptions of function usage in the toolset, and  $\Phi_{global}$ is the previous chat history with the user from the global memory module. \textbf{Appendix~\ref{app:chat_log}} provides sample outputs of this process. 

\subsection{Programmer: Coding for BIM model generation}

The enhanced user requirements will be input into the \textsc{Programmer}'s prompt template, as illustrated in Listing~\ref{lst:pro_pt_main}. The \textsc{Programmer} agent is required to write concise Python code utilizing the functions solely from the toolset and the built-in standard Python libraries to accomplish the tasks specified by the \textsc{Instruction Enhancer}. Following established standard best practices in prompt engineering \cite{10.1007/978-981-99-7962-2_30}, we clearly defined the agent's roles, tasks, output format, constraints, and necessary context in this prompt template.

\begin{lstlisting}[language=TeX, caption={Prompt template of the \textsc{Programmer} agent. Placeholders with "<< >>" indicate the dynamic content that can be inserted into the template, including information on available tool functions, chat history, and enhanced user requirements.}, label={lst:pro_pt_main}]
You are an AI programmer who specializes in coding. Your job is to write Python code that invokes suitable pre-defined tool functions to implement the Instruction enhancer's requirement. Your code style is concise, efficient and accurate.
To help with that, you have access to a set of tools. Each tool is a Python function and has a description explaining the task it performs, the inputs it expects and the outputs it returns.
You should think step by step and write all your Python code in a single code block. You should avoid assuming in your code that variables are predefined. You can only give one reply for each conversation turn. Minimize other prose. 
Accessible tools:
"""
<<Tool function names and descriptions>>
"""

In your code, you should only use the tool functions that are accessible to you, rather than attempting to invent new tools. Use millimeter units. If necessary, you can import and use the Python standard library. 

Now, based on the conversation given below, please complete your answer as a programmer:
<<chat_history>>
Instruction Enhancer: <<task>>
Programmer: 
\end{lstlisting}

The generated Python code includes the necessary steps for modeling and can be considered an efficient textual representation of a BIM model (i.e., the BIM model can be materialized by executing the code). The advantage of generating Python code is that the LLM can flexibly combine and call different tool functions using various algorithmic logic. This is more powerful than JSON-based function-calling, which is constrained to executing single functions sequentially and cannot meet the complex logical demands of tasks such as building modeling. Given that the enhanced requirements contain rich contextual information, we choose to leverage the zero-shot learning \cite{wei2022finetunedlanguagemodelszeroshot} ability of the LLM without providing examples. This approach is intended to allow the LLM to flexibly explore different code logic tailored to various task requirements, rather than being constrained by rigid examples, thereby maximizing the utilization of its pre-trained knowledge. Eq.~\ref{eq:coder_1} illustrates the process of code generation, where we denote $\alpha_{enhanced}$ as enhanced user requirements from the \textsc{Instruction Enhancer}, tool information as $T$, the historical chat records with the \textsc{Instruction Enhancer} as $\Phi_{global}$, and the \textsc{Programmer}'s prompt template as $P_{Co}$:

\begin{equation} \label{eq:coder_1}
code \leftarrow LLM_{Coder}(P_{Co}(\alpha_{enhanced}, T, \Phi_{global})) \;
\end{equation}

A custom Python interpreter executes the generated code within a controlled environment. We use an abstract syntax tree (AST) \cite{4299919} to represent the code, traversing the tree nodes to evaluate each Python expression. This approach allows us to customize the usable syntax and callable functions while enabling more precise error handling. The interpreter uses a state dictionary to store and track the results of code execution, including imported packages, defined function objects, and variable names and their values. This lays the technical foundation for the \textsc{Programmer} agent's memory capability, the LLM can utilize and access variables and functions defined in previous dialogues or directly continue previous code. The interpreter can still execute code correctly by retrieving the state dictionary, ensuring comprehensive context is maintained at the code level throughout the entire session. We extended previous work \cite{Du:2024:Copilot_BIM} to allow the interpreter to support more data types and advanced syntax while restricting potentially problematic syntax like the \textit{while} statement. Additionally, the interpreter can only evaluate functions from the toolset and Python built-in library (except for custom-defined functions within the generated code), preventing the invocation of arbitrary third-party packages. An example output from the \textsc{Programmer} agent is illustrated in Listing \ref{lst:pro_output}, which implements the sample user instruction in Fig. \ref{fig:Methodology}.

\begin{lstlisting}[language=TeX, caption={Example output from the \textsc{Programmer} agent,  which implements the sample user instruction in Fig. \ref{fig:Methodology}.}, label={lst:pro_output}]
```py
# Step 1: Create Story Layers
ground_floor_uuid = create_story_layer("Ground Floor", 0, 1)
first_floor_uuid = create_story_layer("First Floor", 3000, 2)

# Step 2: Create Perimeter Walls on Ground Floor
set_active_story_layer("Ground Floor")
wall_A_uuid = create_wall((0, 0), (32000, 0), ground_floor_uuid)
wall_B_uuid = create_wall((32000, 0), (32000, 16000), ground_floor_uuid)
wall_C_uuid = create_wall((32000, 16000), (0, 16000), ground_floor_uuid)
wall_D_uuid = create_wall((0, 16000), (0, 0), ground_floor_uuid)

# Step 3: Create Internal Corridor Walls on Ground Floor
wall_E_uuid = create_wall((0, 6000), (32000, 6000), ground_floor_uuid)
wall_F_uuid = create_wall((0, 10000), (32000, 10000), ground_floor_uuid)

# Step 4: Create Room Dividing Walls on Ground Floor
wall_G_uuid = create_wall((8000, 0), (8000, 6000), ground_floor_uuid)
wall_H_uuid = create_wall((16000, 0), (16000, 6000), ground_floor_uuid)
wall_I_uuid = create_wall((24000, 0), (24000, 6000), ground_floor_uuid)
wall_J_uuid = create_wall((8000, 10000), (8000, 16000), ground_floor_uuid)
wall_K_uuid = create_wall((16000, 10000), (16000, 16000), ground_floor_uuid)
wall_L_uuid = create_wall((24000, 10000), (24000, 16000), ground_floor_uuid)

# Step 5: Add Doors to Rooms on Ground Floor
add_door_to_wall(wall_E_uuid, 0, 4000, "Room 1 Door")
add_door_to_wall(wall_E_uuid, 0, 12000, "Room 2 Door")
add_door_to_wall(wall_E_uuid, 0, 20000, "Room 3 Door")
add_door_to_wall(wall_E_uuid, 0, 28000, "Room 4 Door")
add_door_to_wall(wall_F_uuid, 0, 4000, "Room 5 Door")
add_door_to_wall(wall_F_uuid, 0, 12000, "Room 6 Door")
add_door_to_wall(wall_F_uuid, 0, 20000, "Room 7 Door")
add_door_to_wall(wall_F_uuid, 0, 28000, "Room 8 Door")

# Step 6: Add Windows to Rooms on Ground Floor
add_window_to_wall(wall_A_uuid, 1000, 4000, "Room 1 Window")
add_window_to_wall(wall_A_uuid, 1000, 12000, "Room 2 Window")
add_window_to_wall(wall_A_uuid, 1000, 20000, "Room 3 Window")
add_window_to_wall(wall_A_uuid, 1000, 28000, "Room 4 Window")
add_window_to_wall(wall_C_uuid, 1000, 4000, "Room 5 Window")
add_window_to_wall(wall_C_uuid, 1000, 12000, "Room 6 Window")
add_window_to_wall(wall_C_uuid, 1000, 20000, "Room 7 Window")
add_window_to_wall(wall_C_uuid, 1000, 28000, "Room 8 Window")

# Step 7: Duplicate Ground Floor Layout to First Floor
walls_to_duplicate = [wall_A_uuid, wall_B_uuid, wall_C_uuid, wall_D_uuid, wall_E_uuid, wall_F_uuid, wall_G_uuid, wall_H_uuid, wall_I_uuid, wall_J_uuid, wall_K_uuid, wall_L_uuid]
duplicated_walls_uuids = []
for wall_uuid in walls_to_duplicate:   
    duplicated_walls_uuids.extend(duplicate_obj(wall_uuid, first_floor_uuid, 1))

# Step 8: Create Slabs for Each Floor
ground_floor_slab_profile = create_polygon([(0, 0), (32000, 0), (32000, 16000), (0, 16000)], ground_floor_uuid)
ground_floor_slab_uuid = create_slab(ground_floor_slab_profile, ground_floor_uuid)
set_slab_height(ground_floor_slab_uuid, 0)

first_floor_slab_profile = create_polygon([(0, 0), (32000, 0), (32000, 16000), (0, 16000)], first_floor_uuid)
first_floor_slab_uuid = create_slab(first_floor_slab_profile, first_floor_uuid)
set_slab_height(first_floor_slab_uuid, 3000)

# Step 9: Create Pitched Roof
set_active_story_layer("First Floor")
roof_profile = create_polygon([(0, 0), (32000, 0), (32000, 16000), (0, 16000)], first_floor_uuid)
roof_uuid = create_pitched_roof(roof_profile, first_floor_uuid, 30, 1000, 3000, 300)
set_pitched_roof_style(roof_uuid, "Sloped Wood Struct Insul Flat Clay Tile")

# Step 10: Set Wall Styles
for wall_uuid in walls_to_duplicate + duplicated_walls_uuids:
    set_wall_style(wall_uuid, "Brick Wall")
```
\end{lstlisting}

If an exception $E$ is thrown by the interpreter during code evaluation, resolving this error will be treated as a new task input for the \textsc{Programmer}. The code that caused the exception will be used as the new history information. Both the error and the code will be re-input into the prompt template $P_{Co}$ for the \textsc{Programmer} to regenerate new code. 
This design leverages self-correction or self-reflection \cite{pan-etal-2024-automatically}, a common strategy for improving the outputs for language models \cite{fuchs2024usinglargelanguagemodels}, where the LLM is prompted or guided with feedback to solve problems in its own output.
As shown in Eq.~\ref{eq:coder_2}, this self-reflective modification loop will iterate up to $n$ times until the newly generated $code^{n+1}$ is free of exceptions $E^{n+1}$. If the issue is not resolved within 3 attempts, we will interrupt the process to seek human feedback.

\begin{equation} \label{eq:coder_2}
code^{n+1} \leftarrow LLM_{Coder}(P_{Co}(E^{n}, T, \sum\nolimits code^{[0:n]} ))\quad \text{where}\; n\in [0,3) \;
\end{equation}

It should be noted that within this self-reflection loop, the state dictionary will be updated, and the failed attempts will be held temporarily in the memory module and will be cleared if the loop is exited. In each iteration, the only historical information the \textsc{Programmer} can see is the cumulative code history that failed in the previous steps. We only store responses from the \textsc{Programmer} that are free of errors in the memory modules.

\subsection{Reviewer: Model quality assessment and iterative improvements}

The successfully executed code will generate a building model within the BIM authoring software. Although we used extensive prompt engineering in earlier stages to guide the LLMs in producing spatially and geometrically reasonable results, the inherently stochastic nature of the process may still lead to flaws in the generated building model. 

Therefore, deterministic domain-specific rules are used to verify and refine the generated BIM model. We employ a rule-based model checker to evaluate the model quality. Following the categorization by \cite{Solihin2015}, we define a series of rules covering classes 1 to 3, primarily checking for geometric conflicts between components (e.g., whether doors and windows overlap), correct semantic attribute definitions (e.g., whether each component has a unique GUID), and compliance of spatial layouts with architectural common sense (e.g., whether the roof is supported by walls and not floating). Detailed documentation of all rules can be found in the \textbf{Appendix~\ref{app:RulesetDocumentation}}. 
The issues identified in the model are exported to BIM Collaboration Format (BCF) files. A script is used to automatically extract useful information from the BCF, i.e., the name and description of the issues and corresponding rules, as well as the GUIDs of the associated components. An example of the extracted issue information from a BCF is illustrated in Listing \ref{lst:issue}. This BCF is the output from the model checker after checking the model generated from the sample user instruction in Fig. \ref{fig:Methodology}. 

\begin{lstlisting}[language=TeX, caption={Example issue information extracted from a BCF, which is the output from the model checker after checking the model generated from the sample user instruction in Fig. \ref{fig:Methodology}.}, label={lst:issue}]
- Issue: No Space Components in prompt_4cc27a7f-e23f-4ab4-b815-962eb0d4b4d8
- Issue description: Model Should Have Spaces
The rule checks if functional areas (spaces) are present in the model.
Desired resolution: Create meaningful functional areas in the building by considering building requirements.
No Space Components in prompt_4cc27a7f-e23f-4ab4-b815-962eb0d4b4d8
prompt_4cc27a7f-e23f-4ab4-b815-962eb0d4b4d8 doesn't contain any Space components
- Related element uuids: []
\end{lstlisting}

We denote this issue information as $I$ and input it along with the Python $code$ that generated the checked model and the toolset information $T$ into the \textsc{Reviewer} agent's prompt template $P_{Rev}$ (see Listing~\ref{lst:re_pt_main}), asking the \textsc{Reviewer} to provide suggestions $\beta$ on solving the issues by synthesizing the provided context. 

\begin{lstlisting}[language=TeX, caption={Prompt template of the \textsc{Reviewer} agent. Placeholders with "<< >>" indicate the dynamic content that can be inserted into the template, including previously executed code, issues reported by the model checker, and information on available tool functions.}, label={lst:re_pt_main}]
You are a senior code reviewer. You review Python code written by an AI programmer that calls specific APIs to generate buildings. Your job is to propose solutions to the programmer for issues that arise in the generated building model, based on the API documentation. Issues are listed in the form of bullet points. Each issue contains a description and list of uuids of the corresponding elements in the generated building model. These uuids can be used to directly manipulate the related elements via the API functions. 
Here is the information you can refer to:
Code from the AI programmer:
<<code>>
Issues in the generated buildings:
<<issues>>
API documentation:
<<Tool function names and descriptions>>

Please provide your guidance below as a reviewer. Use millimeter units. Think step by step. Try to avoid duplicating the programmer's code, but rather suggest code fixes.
Reviewer: 
\end{lstlisting}

Formally, this assessment process conducted by the \textsc{Reviewer} can be summarized by Eq.~\ref{eq:rev}:

\begin{equation} \label{eq:rev}
\beta \leftarrow LLM_{Reviwer}(P_{Rev}(I, T, code)) \;
\end{equation}

The issue-solving suggestions $\beta$ are then fed to the \textsc{Programmer} as an input task to guide it in generating code to resolve conflicts. An example of the \textsc{Reviewer}'s suggestions is shown in Listing \ref{lst:rev_output}. The suggestions aim to solve the issue reported in Listing \ref{lst:issue}. Due to the space limitations, only a selection of representative rows is presented in this example. \textbf{Appendix \ref{app:chat_log}} illustrates another concrete example in full length.

\begin{lstlisting}[language=TeX, caption={Example output (partial) from the \textsc{Reviewer} agent used to enhance the quality of the model generated from the sample user instruction in Fig. \ref{fig:Methodology}}, label={lst:rev_output}]
To address the issue of missing space components in the model, we need to create functional areas (spaces) for each room on both the Ground Floor and the First Floor. This will ensure that the model adheres to the requirement of having defined functional areas.
Here are the steps to resolve the issue:
1. **Identify the boundaries of each room**: Based on the wall coordinates, we can determine the vertices that define the boundary of each room.
2. **Create functional areas for each room**: Use the `create_functional_area` API to create spaces for each room on the respective story layers.
...
### Code Fixes
Add the following code after Step 4 (creating room dividing walls) to create functional areas for each room:
```python
# Step 4.1: Create Functional Areas on Ground Floor
ground_floor_rooms = [("Room 1", [(0, 0), (8000, 0), (8000, 6000), (0, 6000)]),
...
("Room 8", [(24000, 10000), (32000, 10000), (32000, 16000), (24000, 16000)])]
for room_name, vertices in ground_floor_rooms:
    create_functional_area(vertices, room_name, ground_floor_uuid)
# Step 7.1: Create Functional Areas on First Floor
...
```
### Explanation
- **Ground Floor Functional Areas**: We define the vertices for each room based on the wall coordinates and create functional areas using the `create_functional_area` function.
- **First Floor Functional Areas**: Since the First Floor layout is a duplicate of the Ground Floor, we use the same vertices to create functional areas on the First Floor.
...
\end{lstlisting}

This quality optimization loop will iterate multiple times until the checker no longer finds any errors. Similar to the self-reflection loop, if issues persist after 3 attempts, we will interrupt the loop and have a human designer resolve the issues. 

Building on the core ideas introduced in the Reflexion framework \cite{shinn2023reflexionlanguageagentsverbal}, our quality optimization loop can be viewed as a domain-specific extension and adaptation of \textit{verbal reinforcement learning}. Unlike traditional reinforcement learning typically requires extensive training samples and frequent weight updates, the key here is leveraging natural language feedback to drive the learning process. In Reflexion, agents iteratively improve their decision-making by reflecting on their performance through scalar or textual feedback from a simulator, storing this reflective text in episodic memory to inform subsequent actions. Similarly, our process leverages detailed, domain-specific rule-based feedback extracted from the model checker, integrating it with the code generation loop. This integration effectively transforms deterministic checking results into a dynamic learning signal that guides the \textsc{Reviewer} and \textsc{Programmer} agents, enabling them to reflect on prior outputs, perform semantic error analysis, and improve code generation.
By doing so, our system is able to mitigate the stochastic errors inherent in LLM-generated designs and converge on a compliant BIM model.

In summary, our framework employs triple-loops to generate BIM model and iteratively improve its quality, which can be summarized by Alg.~\ref{alg:opt}.

\begin{algorithm}[h!]
\caption{Triple loops for model generation and optimization}
\label{alg:opt}
\begin{algorithmic}[1]
%\Require Data: $123$

\Repeat\, when user gives instruction $\alpha$ \qquad\qquad\qquad\qquad\qquad\qquad \text{//Human in the loop}
    \State Condition for calling \textsc{Architect}: $I_{condition} \leftarrow LLM_{ProductOwner}(P_{IE}(\alpha, T, \Phi_{global}))$
    \State $Building\_Plan \leftarrow I_{condition}\cdot F(LLM_{\textsc{Architect}}(P_{Arch}(\alpha')))$
    \State $\alpha_{enhanced} \leftarrow LLM_{ProductOwner}(P_{IE}(\alpha, T, \Phi_{global}) \mid Building\_Plan )$

    \State Code generation: $code^{0} \leftarrow LLM_{Coder}(P_{Co}(\alpha_{enhanced}, T, \Phi_{global}))$
    \For{$n=0$ to $3$} \qquad \qquad\qquad\qquad\qquad\qquad\qquad\qquad\qquad\qquad \text{//Self-reflection loop}
        \State Code execution: $Model \leftarrow Interpreter(code^{n})$
        \If{Exception $E^{n}$ in $code^{n}$}
            \If{$n=3$}
                \State Stop and requires new user input $\alpha$
            \Else
                \State Self-correction: $code^{n+1} \leftarrow LLM_{Coder}(P_{Co}(E^{n}, T, \sum\nolimits code^{[0:n]}))$
            \EndIf
        \Else
            \State $Break$
        \EndIf
    \EndFor
    
    \State Update $\Phi_{global}$ with $\alpha$, $\alpha_{enhanced}$, and $code^{n}$
    \State Update $\Phi_{local}$ with $code^{n}$
    \For{$t=0$ to $3$} \qquad\qquad\qquad\qquad\qquad\qquad\qquad\qquad\qquad \text{//Quality-optimization loop}
        \State Model checking: $I_{t} \leftarrow Checker(Model_{t}, Rules)$
        \If{Issues $I_{t}$ exist in $Model_{t}$}
            \If{$t=3$}
                \State Stop, purge $\Phi_{local}$ and requires human intervention
            \Else
                 \State Issues resolution proposal: $\beta_{t} \leftarrow LLM_{Reviwer}(P_{Rev}(I_{t}, T, \Phi_{local}))$
                \State Code revise: $code_{t+1} \leftarrow LLM_{Coder}(P_{Co}(\beta_{t}, T, code_{t}))$
                \State \textbf{Re-enter \textit{self-reflection loop}} for executing $code_{t+1}$ to generate $Model_{t+1}$
                 \State Store the error-free $code^{n}_{t+1}$ in $\Phi_{local}$
            \EndIf
        \Else
            \State $Break$ and purge $\Phi_{local}$
        \EndIf
    
    \EndFor

\Until user terminates the dialogue
\end{algorithmic}
\end{algorithm}

\newpage
\section{Prototype implementation}
\label{sec:prototype}

\begin{figure}[h!]
\centering
\includegraphics[width=0.95\textwidth]{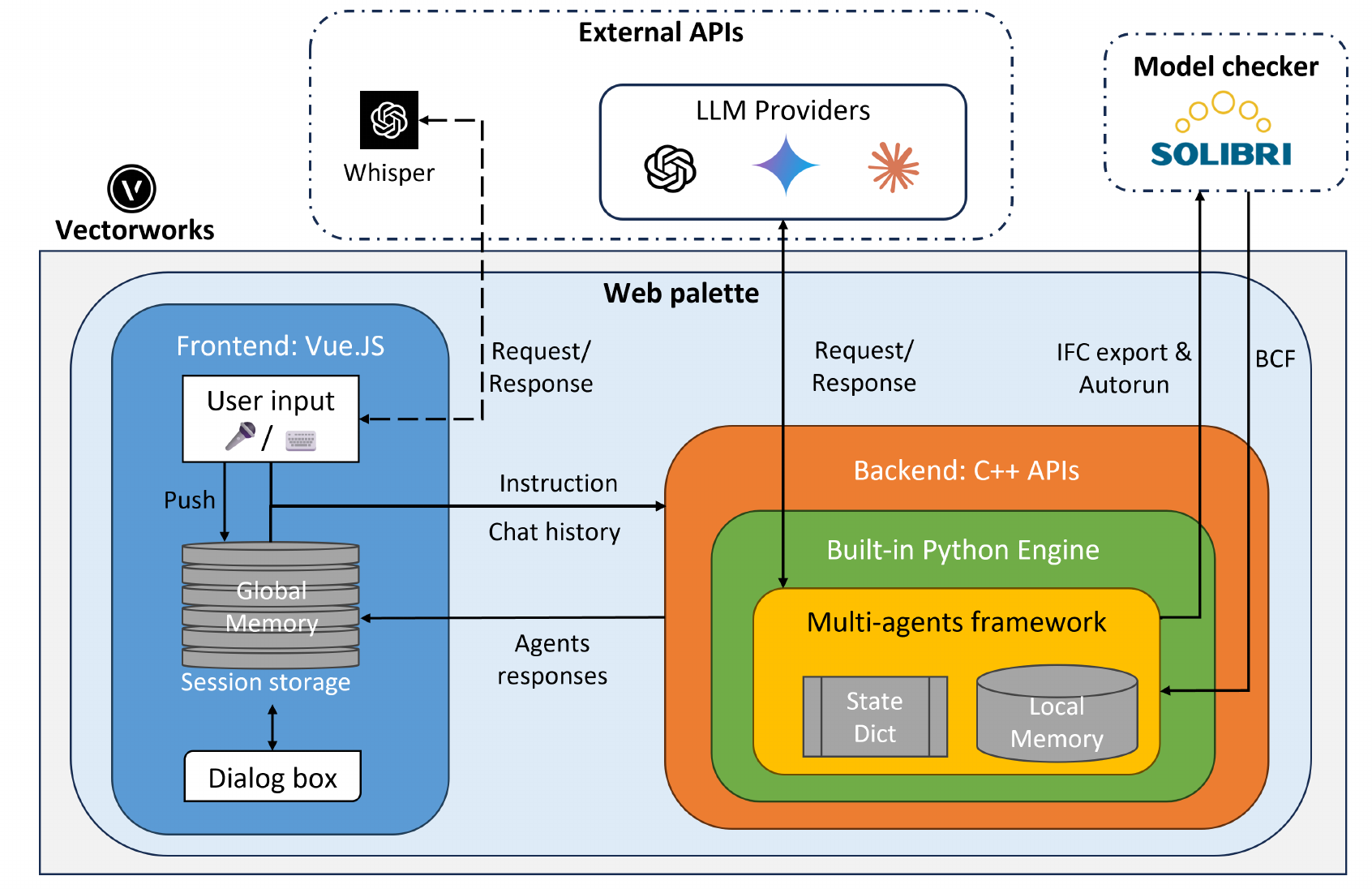}
\caption{Software architecture of the implemented prototype in Vectorworks based on web palette plugin template}
\label{fig:software_architect}
\end{figure}

We developed an interactive software prototype using the architecture shown in Fig.~\ref{fig:software_architect}, integrating the proposed framework into the BIM authoring software Vectorworks. Our implementation is based on Vectorworks' open-source web palette plugin template \cite{SDKExamples} and significantly extends previous work \cite{Du:2024:Copilot_BIM} to support multi-agent workflows. The frontend of the web palette is implemented with Vue.js and runs in a web environment built on Chromium Embedded Framework (CEF), allowing us to embed dynamic web interfaces in Vectorworks using modern frontend technologies. The backend of the web palette is a C++ application, enabling the definition and exposure of asynchronous JavaScript functions within a web frame, while the actual logic is implemented using C++ functions.

Since our multi-agent framework is entirely based on Python, we invoke Vectorworks' built-in Python engine within the corresponding C++ functions to execute our code, thereby delegating the JavaScript implementation. 
%\begin{color}{blue}
Our framework supports various mainstream LLMs, such as o4-mini-high \cite{openai2025o3o4mini}, Claude-sonnet-4 \cite{anthropic2025claudesonnet4}, and Gemini-2.5-Pro \cite{google2025gemini25pro}.
%\end{color}
Within the framework, we maintain a state dictionary and a local memory module to store the state between Python calls and the interaction records of the agents in the quality optimization loop, respectively.

Solibri is used as the model checker in our implementation. Solibri Model Checker is an advanced, rule-based validation and quality control software designed to verify and analyze BIM models against industry standards and project-specific requirements. Operating on IFC-compliant data, it inspects geometry and metadata, automatically identifying inconsistencies, design conflicts, missing information, and code compliance issues. By leveraging customizable rule sets, users can tailor model checks to specific project or organizational requirements, defining parameters such as geometry constraints, design standards, and data integrity rules. These rules are configured through a user-friendly interface, ranging from clash detection to model classification. In addition, Solibri supports the BIM Collaboration Format (BCF). This open data standard streamlines communication and issue management by providing a structured way to document and share issues discovered during model checking. 

\begin{figure}[H]
\centering
\includegraphics[height=0.6\textheight]{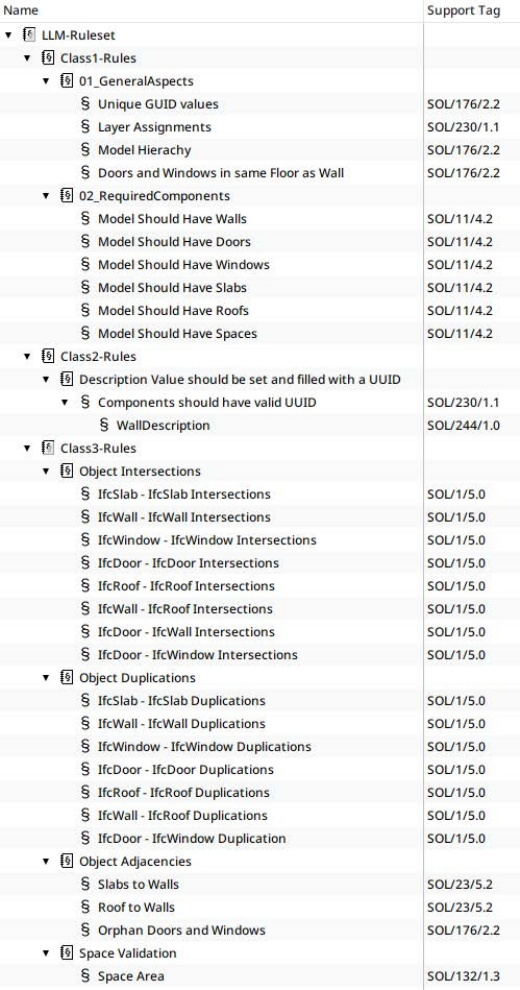}
\caption{Implemented rules in Solibri Model Checker}
\label{fig:solibri}
\end{figure}

%\begin{color}{blue}
We use Vectorworks API to export the generated models as IFC files, and then use the Solibri Autorun commands to automatically launch the Solibri Model Checker, perform checks, and output BCFs.
In our current implementation, the entire workflow can operate autonomously, except for the IFC export from Vectorworks, which requires minor manual intervention (naming and saving the exported file). In the Solibri Model Checker, 30 rules are implemented according to the rule documentation in \textbf{Appendix~\ref{app:RulesetDocumentation}}, as shown in Fig.~\ref{fig:solibri}. 
By selecting these 30 rules, we ensured that various dimensions of model quality were evaluated, including geometric conflicts (e.g., overlap of doors/windows), semantic correctness (e.g., proper assignment of attributes like unique GUIDs), as well as basic architectural common sense (e.g., ensuring a roof is supported by walls rather than floating). Our rule selection was guided by the desire to capture a broad yet practical range of verifiable BIM criteria within a feasible implementation scope. By covering the rule classes of \citeN{Solihin2015}'s framework and targeting frequently encountered modeling errors, we aim to provide a meaningful yet efficient evaluation of LLM-generated building models. 
%\end{color}

\begin{figure}[H]
\centering
\includegraphics[width=0.95\textwidth]{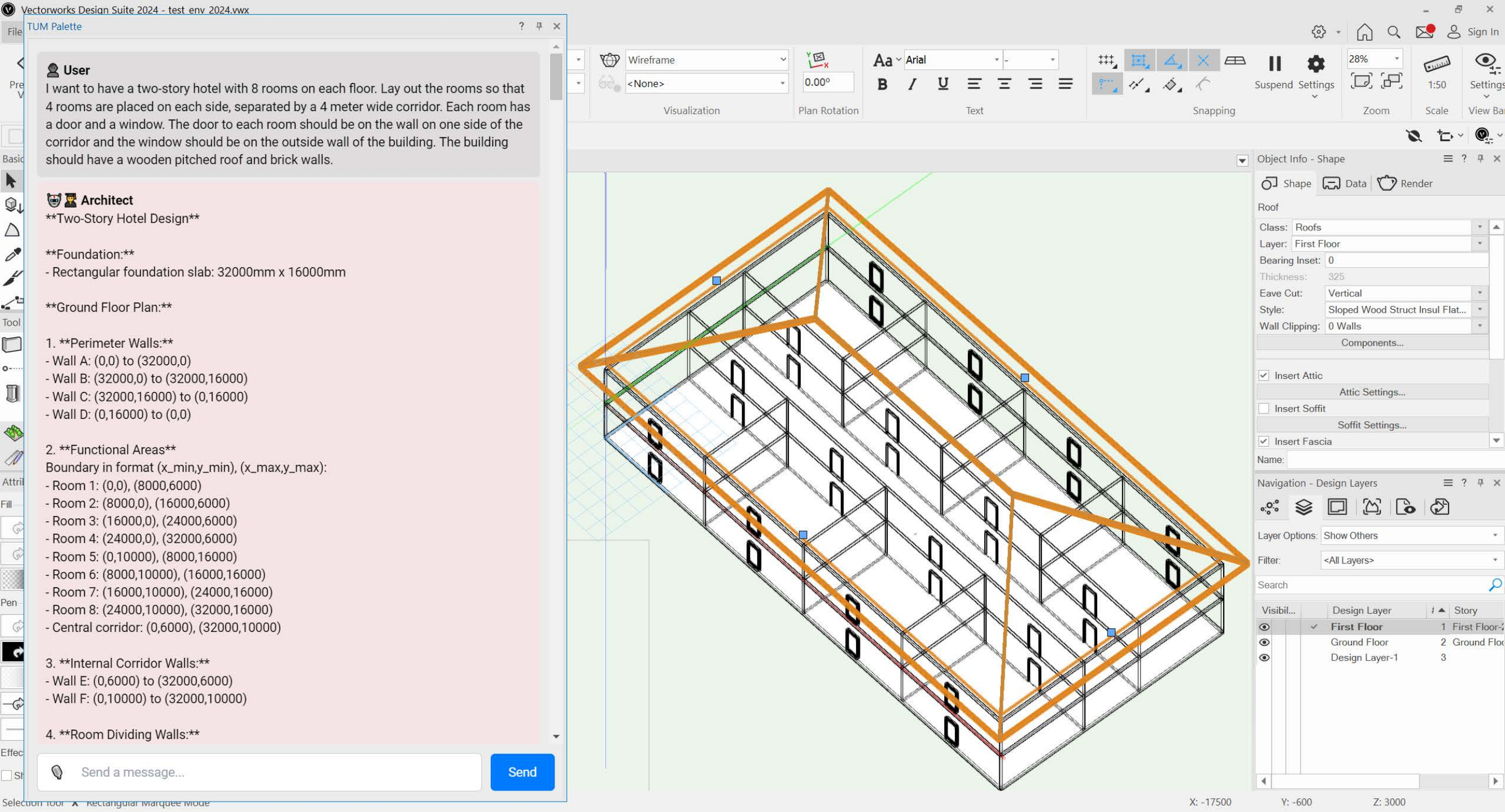}
\caption{Seamless integration in Vectorworks. Users can give instructions via audio or text in the built-in chat window. The responses of the individual agents are displayed in different colors in the dialog box. The generated building model contains rich semantic information and can be further edited directly in the BIM software. A demo video can be found in supplemental materials.}
\label{fig:ui}
\end{figure}

On the frontend, a session begins with a new user input. If the input is audio, we employ the Whisper model \cite{radford2022robustspeechrecognitionlargescale} to convert the speech to text and populate the input message box. We use page session storage to store user input text and the chat records of the agents, displaying them sequentially in the dialog box with different colors. Users can easily start a new session by reloading the page. Fig.~\ref{fig:ui} shows the developed prototype. Users can chat directly with the agents by clicking the microphone button, generating editable building models directly in Vectorworks. A demo video can be found in the supplemental materials.

\newpage
%\begin{color}{blue}
\section{Experiments and evaluation}

\subsection{Experimental setup}

This section is devoted to the experimental evaluation of the presented methodology. The evaluation is performed via the employment of test prompts to the proposed framework and comparing the generated outcome of three state-of-the-art LLMs, including o4-mini-high \cite{openai2025o3o4mini}, Claude-sonnet-4 \cite{anthropic2025claudesonnet4}, and Gemini-2.5-Pro \cite{google2025gemini25pro}. 25 test prompts in Table~\ref{table:test_prompts} are designed to evaluate various generation capabilities of the proposed framework from eight aspects, including base geometry and shape, multi-story and vertical composition, material specification and mixing, roof types and combinations, space layout, opening elements, circulation, and parametric dimensions accuracy, as detailed in Table \ref{table:dimension_matrix}.  

Given the inherent stochasticity of generative models, each test prompt was input into each LLM three times, resulting in a total of 534 IFC models (including intermediate results from the optimization process). All experiments are based on this dataset and aim to report statistically meaningful results.
This strategy aims to strike a good balance: the prompts are carefully designed to test the system's capabilities from multiple aspects within its generative scope while covering common building design requirements, and the repeated runs across multiple LLMs provide evidence for the framework's generalizability and robustness.

We employ both automated methods as well as human expertise for assessing the proposed framework from four main perspectives:

\begin{itemize}
    \item \textbf{Final model checking pass rates.} We measure the pass rate of the final generated models against 30 rule checks to quantitatively assess the quality of the models produced by the framework.
    \item \textbf{Issue amount.} We track the trend in the issue amount reported by the model checker during framework execution, providing a quantitative measure of the effectiveness of our model quality optimization loop.
    \item \textbf{Experts assessment of model adherence and design.} Multiple human experts assess how well the generated BIM models adhere to the test prompts and conduct a rigorous design evaluation from the perspective of real-world projects. This allows for both quantitative and qualitative assessment of the framework's robustness and limitations.
    \item \textbf{Code generation accuracy.} We use CodeBERTScore \cite{zhou-etal-2023-codebertscore} to quantitatively evaluate the accuracy of the generated code, supplementing the aforementioned model-based assessments with an additional perspective on the framework's generation capability.
\end{itemize}

\begin{figure}[H]
    \centering
    \begin{subfigure}[b]{0.90\textwidth}
        \centering
        \includegraphics[width=\textwidth]{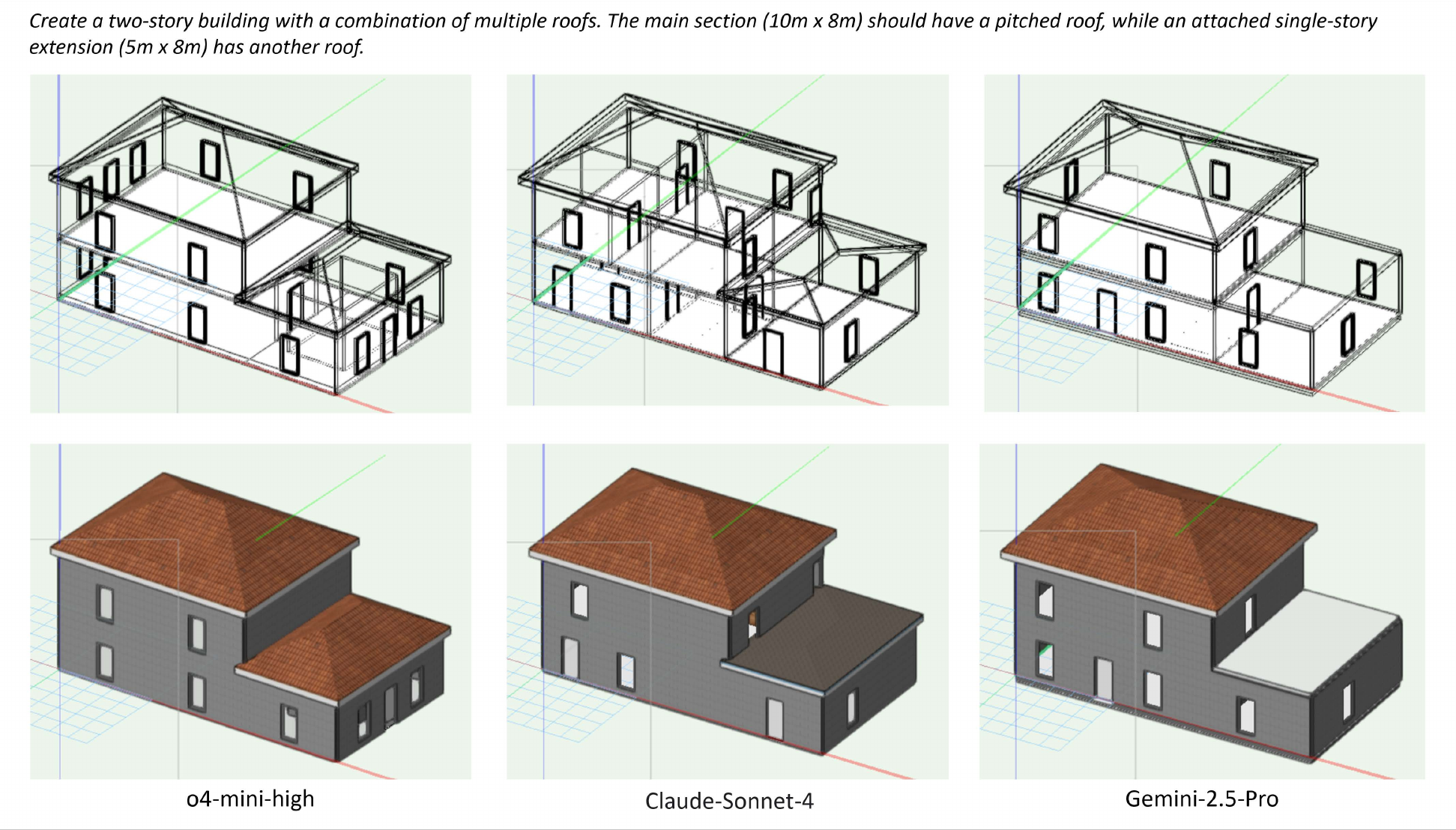}
        \subcaption{Test prompt Nr.17}
        \label{fig:sub_a}
    \end{subfigure}
    \begin{subfigure}[b]{0.90\textwidth}
        \centering
        \includegraphics[width=\textwidth]{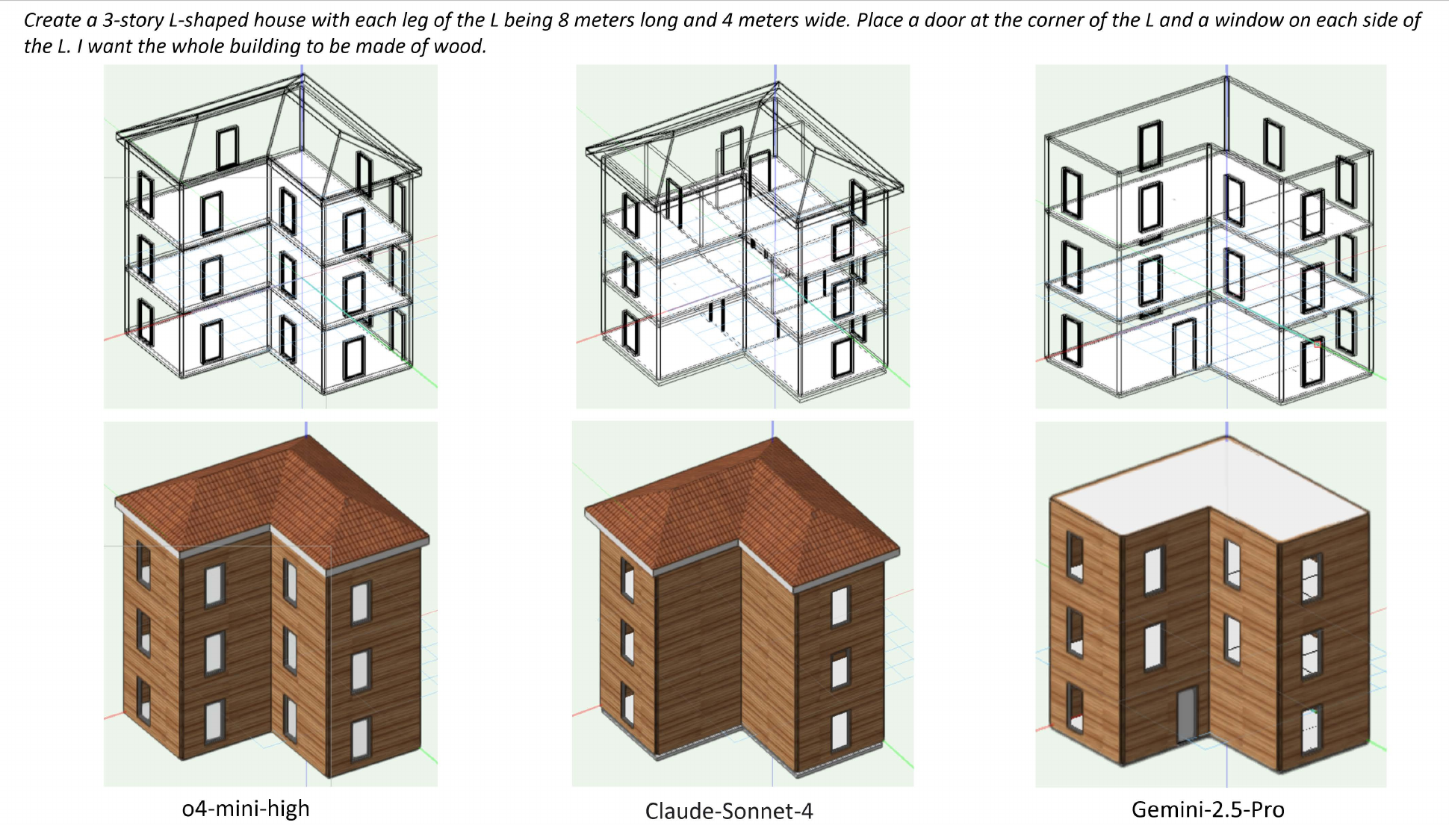}
        \subcaption{Test prompt Nr.5}
        \label{fig:sub_b}
    \end{subfigure}
    \caption{Qualitative results generated by the Text2BIM framework based on test prompts Nr.17 (a) and Nr.5 (b). Visualizations of the models generated by other prompts can be found in \textbf{Appendix~\ref{app:Visualization}}}
    \label{fig:qualitative}
\end{figure}

Further details are provided in the following sections. Fig.~\ref{fig:qualitative} visualizes part of the building models generated by the proposed framework. Due to space constraints, the full list of generated building models can be found in \textbf{Appendix~\ref{app:Visualization}}. 

% begin test prompts %
\begin{table}[H]
\caption{Test prompts}
\label{table:test_prompts}
\centering
\renewcommand{\arraystretch}{1.25}
\footnotesize
\begin{tabular}{c p{15cm}}
\hline\hline
\multicolumn{1}{c}{\textbf{Nr.}} &
\multicolumn{1}{c}{\textbf{Content}} \\
\hline
1 & I want to build a two-story hotel with eight rooms on each floor. The rooms are arranged in groups of four on each side, separated by a 4-meter-wide corridor in the middle. Each room has a door and a window. The doors of the rooms are on the corridor side of the wall, and the windows are on the outside wall of the building. The building should have a wooden pitched roof and brick walls.\\
\hline
2 & Create a basic 3D model of a four-story residential house with dimensions of 10 by 6 meters. \\
\hline
3 & Create a single-story residential house with a total floor area of 120 square meters. Includes three bedrooms, two bathrooms, a kitchen, and a living room. The house should have a wooden pitched roof, and incorporate at least four windows and one main entrance door. Use the concrete walls.\\
\hline
4 & Create a building with three connected sections. Each section should be a rectangle (10m x 5m) with walls of 3 meters in height. The sections should be connected by 5-meter-long walls. Add slabs and a continuous, concrete-pitched roof that covers all sections. Add doors and windows to each section. Choose the right material for the wall.\\
\hline
5 & Create a 3-story L-shaped house with each leg of the L being 8 meters long and 4 meters wide. Place a door at the corner of the L and a window on each side of the L. I want the whole building to be made of wood.\\
\hline
6 & Design a building with a complex polygonal footprint (e.g., hexagonal). Each side of the hexagon should be 5 meters. Add a slab for the floor and a pitched roof. Include a door on one side and a window on each of the other sides. \\
\hline
7 & Construct a residential building with a rectangular footprint (15m x 10m), a pitched roof and two floors. Create balconies by extending the floor slab outwards from the exterior walls on the first floor. Add doors and windows to each floor. Make sure that the balconies are accessible from the inside. \\
\hline
8 & Construct a modern office building with a rectangular base of 20 meters by 20 meters. Set the wall height to 3 meters. Include four rooms (5x5 meters each) along the perimeter, with a central open space. Add doors and windows to each room and a main entrance door to the building. \\
\hline
9 & Design a two-story apartment building with an H-shaped base. Each floor consists of two apartments with two rooms each. \\
\hline
10 & Create a T-shaped, single-story building with a horizontal section of 10 meters x 30 meters and a vertical section of 10 meters x 20 meters. Connect the two sections by placing a door at their junction. Each section has three windows. The entire building is made of concrete. \\
\hline
11 & Design a two-story mixed-material building with a brick first floor and wooden second floor. The footprint should be 12m x 8m. Include 3 rooms on each floor with appropriate doors and windows.\\
\hline
12 & Create a three-story building where each floor uses different wall materials: concrete ground floor, brick first floor, and wood second floor. The building should be 10m x 10m with two rooms per floor separated by interior walls.\\
\hline
13 & Build a two-story community center with a cross-shaped floor plan. Each arm of the cross should be 6m x 4m. The center intersection should be a 4m x 4m open space. Add doors at the end of each arm and windows on the outer walls. Use brick walls and a wooden pitched roof.\\
\hline
14 & Design a three-story residential building with a U-shaped footprint. The main section should be 16m x 8m, with two 6m x 4m wings. Each floor should contain four apartments. Place the main entrance at the center of the U and individual apartment doors along interior corridors.\\
\hline
15 & Create a single-story community center with a large central hall (15m x 10m) surrounded by 6 smaller rooms (3m x 4m each) on three sides. Each small room should have a door connecting to the central hall and a window on the exterior wall.\\
\hline
16 & Design a two-story library with an open floor plan. The ground floor (18m x 12m) should have one large open space with walls on three sides. The upper floor should be divided into four study rooms with windows.\\
\hline
17 & Create a two-story building with a combination of multiple roofs. The main section (10m x 8m) should have a pitched roof, while an attached single-story extension (5m x 8m) has another roof.\\
\hline
18 & Create a single-story octagonal pavilion with each side measuring 3m. Place a door on one side and windows on alternating sides. Use wooden walls and a pyramidal roof converging to a central point.\\
\hline
19 & Design a two-story building with a stepped footprint. The ground floor should be 12m x 12m, and the upper floor should be 8m x 8m, centered on the lower floor. Add a terrace on the exposed lower roof and ensure door access from the upper floor to the terrace.\\
\hline
20 & Create a single-story building with a rectangular footprint (40m x 25m) featuring a figure-8 circulation pattern. The corridors should form two connected loops with resident rooms on the exterior and common spaces in the interior courtyards. Use concrete walls.\\
\hline
21 & Build a three-story apartment building where each floor has a different layout. First floor: two large units; second floor: three medium units; third floor: four small units. Each unit should have its own entrance door and at least two windows.\\
\hline
22 & Build a three-story building with a different footprint on each floor: rectangular ground floor (15m x 10m), L-shaped first floor, and T-shaped top floor. Ensure structural continuity between floors.\\
\hline
23 & Create a two-story office building with an atrium. The building footprint should be 16m x 16m with a 4m x 4m central void through both floors. Surround the void with corridors and place offices around the perimeter with windows on exterior walls. Don't let the roof cover atrium.\\
\hline
24 & Build a single-story school with a 25m x 10m footprint containing five classrooms of equal size arranged linearly. Each classroom should have a door to a 2m wide corridor on one side and two windows on the opposite exterior wall. Use brick walls and a pitched roof.\\
\hline
25 & Build a two-story house with complex interior wall arrangements. The ground floor should have 7 interconnected rooms of varying sizes, and the upper floor should have 5 bedrooms arranged around a central hallway. The building footprint should be approximately 16m x 12m.\\
\hline\hline
\end{tabular}
\end{table}
% end test prompts %

\begin{table}[h!]
\renewcommand{\arraystretch}{1.25}
    \caption{Matrix view of test prompts by the generation capability to be tested (X indicates the prompt involves this aspect)}
    \label{table:dimension_matrix}
    \centering
    \small
    \renewcommand{\arraystretch}{1.0}
    \resizebox{\linewidth}{!}{%
    \begin{tabular}{|p{5cm}|c|c|c|c|c|c|c|c|c|c|c|c|c|c|c|c|c|c|c|c|c|c|c|c|c|}
        \hline
        \textbf{Aspects / Prompts \#} & \textbf{1} & \textbf{2} & \textbf{3} & \textbf{4} & \textbf{5} & \textbf{6} & \textbf{7} & \textbf{8} & \textbf{9} & \textbf{10} & \textbf{11} & \textbf{12} & \textbf{13} & \textbf{14} & \textbf{15} & \textbf{16} & \textbf{17} & \textbf{18} & \textbf{19} & \textbf{20} & \textbf{21} & \textbf{22} & \textbf{23} & \textbf{24} & \textbf{25} \\
        \hline
        \textbf{Base geometry \& shape}\textsuperscript{1} & X &  &  & X & X & X &  & X & X & X &  &  & X & X &  &  &  & X & X & X &  & X & X &  &  \\
        \hline
        \textbf{Multi-story \& vertical composition}\textsuperscript{2} & X & X &  & X & X &  & X &  & X &  & X & X &  &  &  & X & X &  & X &  & X & X & X &  & X \\
        \hline
        \textbf{Material specification \& mixing}\textsuperscript{3} & X &  & X & X & X & X &  &  &  & X & X & X & X &  &  &  &  & X &  & X &  &  &  & X &  \\
        \hline
        \textbf{Roof types \& combinations}\textsuperscript{4} & X &  & X & X &  & X & X &  &  &  &  &  & X &  &  &  & X & X & X &  &  &  & X & X &  \\
        \hline
        \textbf{Space Layout}\textsuperscript{5} & X &  & X &  &  &  & X & X & X &  & X & X &  & X & X & X &  &  &  &  & X &  &  & X & X \\
        \hline
        \textbf{Opening Elements}\textsuperscript{6} & X & X & X & X & X & X & X & X & X & X & X & X & X & X & X & X & X & X & X & X & X & X & X & X & X \\
        \hline
        \textbf{Circulation}\textsuperscript{7} & X &  &  &  &  &  & X &  & X &  &  &  &  & X &  &  &  &  & X & X &  &  & X &  & X \\
        \hline
        \textbf{Parametric dimensions accuracy}\textsuperscript{8} & X & X & X & X & X & X & X & X & X & X & X & X & X & X & X & X & X & X & X & X & X & X & X & X & X \\
        \hline
    \end{tabular}
    }
    \vspace{0.5cm}
    \begin{flushleft}
    \footnotesize
    \textsuperscript{1}Tests ability to parse and generate diverse plan shapes, e.g., rectangular, L/T/U/H-shaped, cross, hexagonal, stepped, octagonal, or combined footprints.\\[0.15cm]
    \textsuperscript{2}Tests creation of multi-level buildings with coherent stacks and offsets, possibly handling different floor layouts with terraces and balconies.\\[0.15cm]
    \textsuperscript{3}Tests whether correct wall/roof materials are applied, including mixed materials across floors or sections.\\[0.15cm]
    \textsuperscript{4}Tests modeling of pitched, pyramidal, continuous and multiple roofs, plus roof exclusions (e.g.\ open atrium, terraces).\\[0.15cm]
    \textsuperscript{5}Tests interior subdivision into rooms, public spaces and their logical connections to satisfy design requirements.\\[0.15cm]
    \textsuperscript{6}Tests precise placement and counts of doors and windows, and their orientation (corridor vs.\ exterior).\\[0.15cm]
    \textsuperscript{7}Tests generation of coherent circulation, e.g., corridors, figure-8 loops, voids and terrace access.\\[0.15cm]
    \textsuperscript{8}Tests adherence of numeric dimensions, areas and proportions throughout the model.\\[0.15cm]
    \end{flushleft}
\end{table}

\subsection{Final model checking pass rates}

For the automated assessment of the quality of the created BIM models, we make use of the model-checking software Solibri and record the pass rates reflecting the ratio of the rules the models have passed.
Table~\ref{table:rule_pass_rate} presents the pass rates of the final BIM models generated by different test prompts during model checking. These rates are calculated based on a total of 30 domain-specific checking rules implemented in Solibri and reflect the quality of the generated model. The AVG records the average pass rate of the corresponding LLM under each test prompt across three repeated runs, while the SD indicates the standard deviation, reflecting the performance stability.

The three LLMs under the proposed framework were generally able to generate high-quality BIM models, with most average pass rates exceeding 0.95. Claude-sonnet-4 stands out for its consistent results, achieving perfect accuracy and zero variance on the majority of prompts, which suggests high reliability. Gemini-2.5-pro and o4-mini-high also perform well but exhibit slightly higher variability, particularly on certain prompts (e.g., Prompt 9, 12, 14, and 17), where accuracy decreases and standard deviations rise. These prompts could be considered "commonly recognized challenges" as none of the LLMs managed to avoid modeling errors in three runs. Overall, the results indicate that while state-of-the-art LLMs are able to generate model with good quality, specific prompts can still reveal differences in robustness and reliability.

\begin{table}[ht]
\renewcommand{\arraystretch}{1.25}
  \centering
  \caption{Model checking rule pass rates of final BIM models generated by different backbone LLMs under test prompts (Avg: Mean pass rate across three runs; SD: Standard deviation)}
  \label{table:rule_pass_rate}
   \begin{tabular}{
    l
    >{\hspace{5pt}}c<{\hspace{5pt}} >{\hspace{5pt}}c<{\hspace{5pt}} % Claude
    >{\hspace{10pt}} 
    >{\hspace{5pt}}c<{\hspace{5pt}} >{\hspace{5pt}}c<{\hspace{5pt}} % Gemini
    >{\hspace{10pt}}
    >{\hspace{5pt}}c<{\hspace{5pt}} >{\hspace{5pt}}c<{\hspace{5pt}} % o4-mini-high
  }
    \hline\hline
    & \multicolumn{2}{c}{Claude-sonnet-4} 
    & \multicolumn{2}{c}{Gemini-2.5-pro} 
    & \multicolumn{2}{c}{o4-mini-high} \\
    \cline{2-3} \cline{4-5} \cline{6-7}
    \# & Avg$\uparrow$ & SD$\downarrow$ & Avg$\uparrow$ & SD$\downarrow$ & Avg$\uparrow$ & SD$\downarrow$ \\
    \hline
    Prompt 1  & 1.0000 & 0.0000 & 1.0000 & 0.0000 & 1.0000 & 0.0000 \\
    Prompt 2  & 0.9556 & 0.0192 & 0.9556 & 0.0192 & 0.9778 & 0.0192 \\
    Prompt 3  & 0.9778 & 0.0192 & 0.9667 & 0.0000 & 1.0000 & 0.0000 \\
    Prompt 4  & 1.0000 & 0.0000 & 1.0000 & 0.0000 & 1.0000 & 0.0000 \\
    Prompt 5  & 1.0000 & 0.0000 & 0.9778 & 0.0192 & 1.0000 & 0.0000 \\
    Prompt 6  & 1.0000 & 0.0000 & 1.0000 & 0.0000 & 1.0000 & 0.0000 \\
    Prompt 7  & 1.0000 & 0.0000 & 0.9222 & 0.1347 & 0.9889 & 0.0192 \\
    Prompt 8  & 1.0000 & 0.0000 & 1.0000 & 0.0000 & 1.0000 & 0.0000 \\
    Prompt 9  & 0.9889 & 0.0192 & 0.9889 & 0.0192 & 0.9111 & 0.0192 \\
    Prompt 10 & 1.0000 & 0.0000 & 1.0000 & 0.0000 & 1.0000 & 0.0000 \\
    Prompt 11 & 0.9889 & 0.0192 & 0.9556 & 0.0192 & 0.9889 & 0.0192 \\
    Prompt 12 & 0.9889 & 0.0192 & 0.9889 & 0.0192 & 0.9111 & 0.0962 \\
    Prompt 13 & 0.9778 & 0.0385 & 1.0000 & 0.0000 & 1.0000 & 0.0000 \\
    Prompt 14 & 0.9500 & 0.0430 & 0.9889 & 0.0192 & 0.9444 & 0.0509 \\
    Prompt 15 & 0.9889 & 0.0192 & 1.0000 & 0.0000 & 1.0000 & 0.0000 \\
    Prompt 16 & 1.0000 & 0.0000 & 1.0000 & 0.0000 & 0.9889 & 0.0192 \\
    Prompt 17 & 0.9000 & 0.0577 & 0.9778 & 0.0385 & 0.8889 & 0.0192 \\
    Prompt 18 & 1.0000 & 0.0000 & 1.0000 & 0.0000 & 1.0000 & 0.0000 \\
    Prompt 19 & 1.0000 & 0.0000 & 0.9889 & 0.0192 & 0.9778 & 0.0192 \\
    Prompt 20 & 1.0000 & 0.0000 & 0.9889 & 0.0192 & 1.0000 & 0.0000 \\
    Prompt 21 & 0.9667 & 0.0333 & 1.0000 & 0.0000 & 0.9222 & 0.0770 \\
    Prompt 22 & 1.0000 & 0.0000 & 0.9778 & 0.0192 & 0.9000 & 0.0333 \\
    Prompt 23 & 0.9889 & 0.0192 & 0.9667 & 0.0333 & 0.9556 & 0.0385 \\
    Prompt 24 & 1.0000 & 0.0000 & 1.0000 & 0.0000 & 1.0000 & 0.0000 \\
    Prompt 25 & 0.9556 & 0.0385 & 0.9889 & 0.0192 & 0.9889 & 0.0192 \\
    \hline\hline
  \end{tabular}
\end{table}

\subsection{Issue amount}
We further evaluated the effectiveness of the model quality optimization method proposed in the framework. Unlike the pass rate, which is a category-related metric, we selected the issue amount as a more fine-grained, instance-level metric. Specifically, while the pass rate provides a broad overview of the types of issues present in the model (e.g., if the model fails the \textit{IfcSlab-IfcSlab intersection} rule but passes the other 29 rules, it indicates a type of issue related to slab position conflicts), the issue amount offers a deeper insight into the quantity of affected component pairs (e.g., how many pairs of slab instances have positioning conflicts in the model).

\begin{figure}[H]
    \centering
    \includegraphics[height=\textheight]{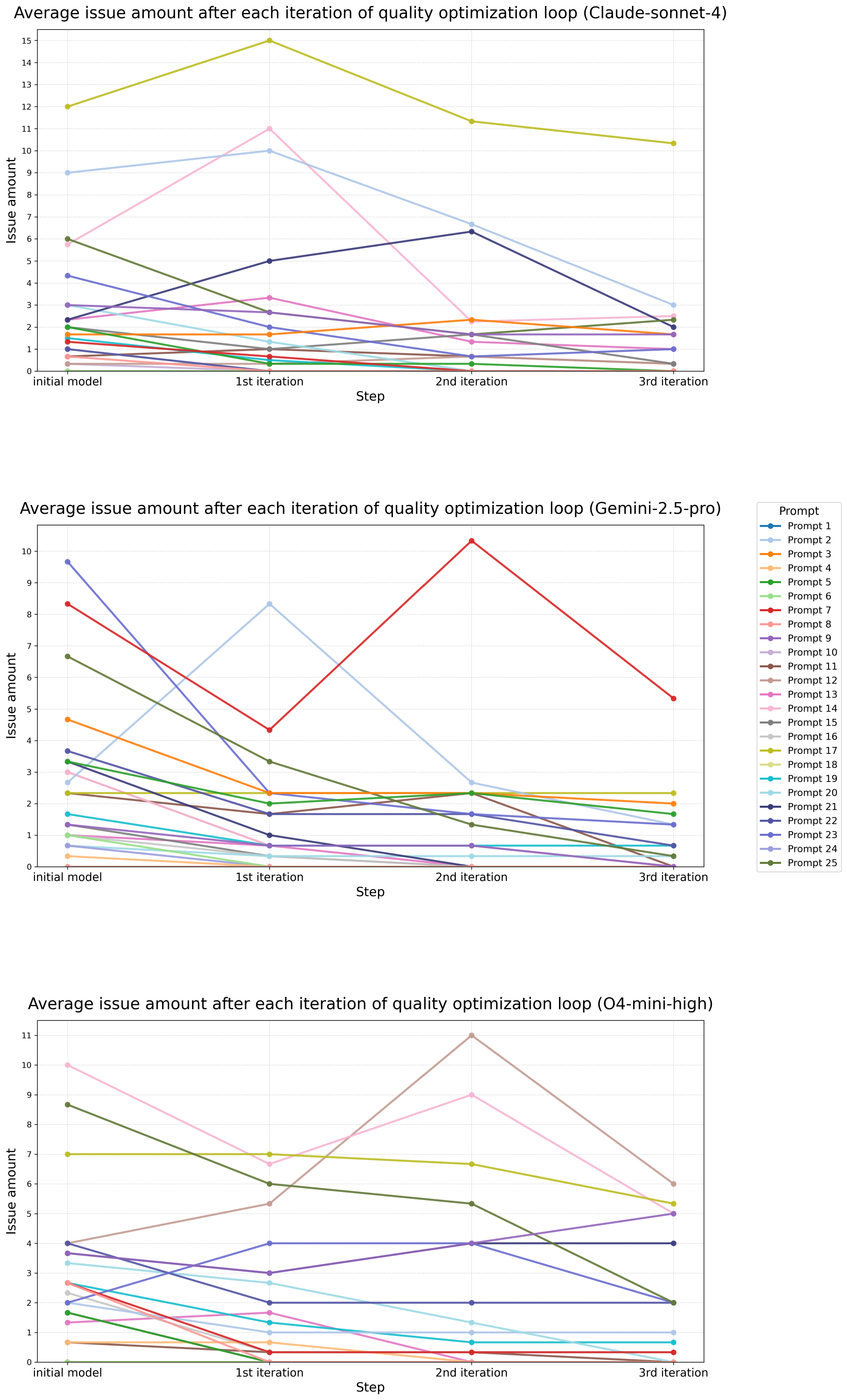}
    \caption{Effectiveness of quality optimization loop with different LLMs used in the framework}
    \label{fig:line_chart}
\end{figure}

\begin{figure}[H]
    \centering
    \includegraphics[width=\textwidth]{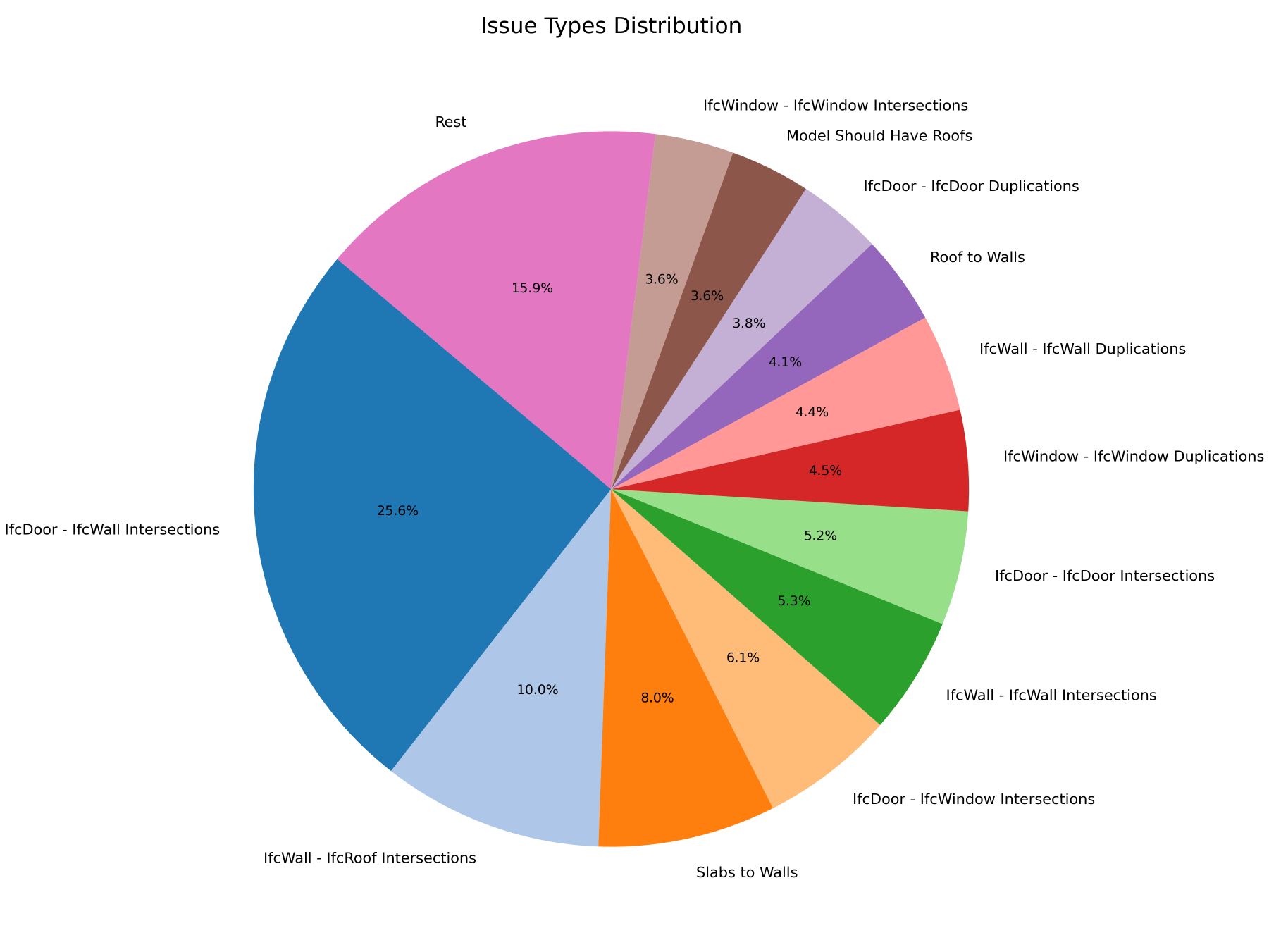}
    \caption{Reported issue types distribution during the experiments}
    \label{fig:issue_pie}
\end{figure}

Fig.~\ref{fig:line_chart} presents three line charts illustrating the average issue amount that exists in BIM models at the initial generation stage, as well as after the first, second, and third rounds of quality optimization iteration. 
Across 25 generative design tasks, three LLMs consistently reduced average issue counts within three optimization passes, though the reduction's efficiency was dictated by complexity required in different test prompts. Simple designs, such as single-story rectangles or regular polygons (e.g., Prompts 1, 4, 6, 18, and 24), initiated with minimal defects and achieved error-free status after the first pass. In contrast, complex multi-story or composite forms with features like wings, multiple roofs, complicated interior layouts or internal voids (e.g., Prompts 14, 17, 23, and 25) started with up to 12 issues. These cases sometimes showed a brief increase in errors after the first or second iteration, resulting in a "saw-tooth" pattern in the chart. 
These issues arise because the \textsc{Reviewer} agent often tries to introduce new elements (such as additional roof sections or doors and windows), or attempts to move already generated building components during the optimization process. However, due to its limited understanding of 3D spatial relationships and the inherent complexity of the design required by such prompt cases, these changes can lead to new collisions between building elements. In subsequent optimization steps, the agent observes the growing number of issues and adjusts its strategy, for example by deleting problematic elements, resulting in a decrease in the issue amount.
This pattern is also reflected in Fig.~\ref{fig:issue_pie}. by visualizing the proportions of different issue types encountered during the optimization process, we observe that LLMs frequently places doors and windows intersecting with walls and also struggles to accurately interpret the relative positions of roofs or slabs with respect to walls when generating 3D models.

Nevertheless, by the final pass, Claude-sonnet-4 and Gemini-2.5-pro solved about 70\% of all detected issues, while o4-mini-high still had a few leftover gaps in the most complex, multi-roof, or stepped-mass cases.

\subsection{Assessment by human experts}

We asked four human experts to evaluate the generated models. Details regarding their backgrounds and qualifications can be found in Table \ref{tab:experts}. To facilitate the evaluation, we prepared an assessment protocol that includes sheets for expert information, explanation of the evaluation criteria, the test prompt contents, etc. This assessment protocol template is available in the supplementary materials. Each expert received a protocol, along with a set of 75 IFC models generated by three LLMs across 25 test prompts, as visualized in Appendix \ref{app:Visualization}.

The experts were asked to assess the generated models from two main aspects: (1) Prompt Adherence, and (2) Design Evaluation/Turing Test. For the second aspect, experts evaluated the model quality purely from the perspective of real-world architectural projects, disregarding prompt specifications and the limitations of our system, and judged whether the generated models could be distinguished from those created by human architects.

Further details are provided in the following sections.

\begin{table}[ht]
\renewcommand{\arraystretch}{1.25} % Increase line height
\caption{Experts' background and qualifications}
\label{tab:experts}
\centering
\begin{tabular}{|l|p{3cm}|c|>{\centering\arraybackslash}p{2cm}|p{5cm}|}
\hline
\textbf{Name} & \textbf{Degree(s)} & \textbf{Years} & \textbf{BIM software skill (1-5)} & \textbf{Background} \\
\hline
Expert 1 & Master's in Civil Engineering & 5 & 4 & Professional licenses in structural engineering; Senior engineer specialized in structural design for buildings and tunnels \\
\hline
Expert 2 & Master's in Architectural Engineering & 2 & 4 & BIM project manager in construction company \\
\hline
Expert 3 & Master's in Civil Engineering & 3 & 4 & BIM modeler in design firm; specialized in BIM-based automation \\
\hline
Expert 4 & Master's in Architecture & 1 & 3 & Junior architect in architecture company \\
\hline
\end{tabular}
\end{table}

\subsubsection{\textbf{Prompt adherence}}

For the convenience of expert evaluation, we summarize the different architectural requirements covered by each test prompt, as illustrated in Table~\ref{table:categorized_prompts}, including aspects such as floor number, base shape and dimensions, spatial features, room/space definition, materials definition, roof constructions, as well as building opening layouts, providing an intuitive overview. Additionally, by deliberately leaving some building requirements unspecified in the test prompts, we aim to experiment with the framework's ability to generate designs in open-ended settings. 

%landscape + table doesn't work in this template%
%\begin{landscape}
\begin{table}[h!]
    \caption{Categorization of design requirements in test prompts ("-" for not specified)}
    \label{table:categorized_prompts}
    \centering
    \renewcommand{\arraystretch}{1.0}
    \resizebox{\linewidth}{!}{%
    \begin{tabular}{|c|p{2cm}|c|p{2.5cm}|p{1.5cm}|p{1.5cm}|c|p{4cm}|p{4cm}|}
        \hline
        \hline
        \textbf{Nr.} &
        \textbf{Type} &
        \textbf{Floors} &
        \textbf{Base Shape} &
        \textbf{Roof} &
        \textbf{Materials} &
        \textbf{Rooms} &
        \textbf{Spatial Features} &
        \textbf{Building Openings} \\
        \hline
        \hline
        1 & Hotel & 2 & Rectangular & Pitched & Wood, brick & 16 & 4m wide corridor & Doors on corridor side, windows on outside wall \\
        \hline
        2 & House & 4 & Rectangular (5m x 3m) & - & - & - & - & - \\
        \hline
        3 & House & 1 & 120 $m^2$, shape not specified & Pitched & Wood, concrete & 7 & - & Main entrance door, at least four windows \\
        \hline
        4 & Building & - & Connected rectangles (10m x 5m) & Pitched & Concrete & - & Connecting sections with 5m long walls & Doors and windows in each section \\
        \hline
        5 & House & 3 & L-shaped & - & Wood & - & Each leg of the L being 8m x 4m & Door at corner, window on each side \\
        \hline
        6 & Building & - & Hexagon & Pitched & - & - & Each side of the hexagon is 5m & Door on one side, window on each other side \\
        \hline
        7 & Residential building & 2 & Rectangular (15m x 10m) & Pitched & - & - & Balconies on first floor & Doors and windows on each floor, accessible balconies \\
        \hline
        8 & Office building & - & Rectangular (20m x 20m) & - & - & 4 & Central open space, rooms along the perimeter & Doors and windows in each room, main entrance door \\
        \hline
        9 & Apartment building & 2 & H-shaped & - & - & 4 apartments (2 per floor) & Each apartment has 2 rooms & - \\
        \hline
        10 & Building & 1 & T-shaped & - & Concrete & - & Vertical section: 10m x 20m, horizontal section: 10m x 30m & Door at junction, three windows per section \\
        \hline
        11 & Building & 2 & Rectangular (12m x 8m) & - & Brick, wood & 6 & Mixed materials per floor & Appropriate doors and windows \\
        \hline
        12 & Building & 3 & Rectangular (10m x 10m) & - & Concrete, brick, wood & 6 & Different materials per floor, interior walls & - \\
        \hline
        13 & Community center & 2 & Cross-shaped & Pitched & Brick, wood & - & Each arm 6m x 4m, center 4m x 4m open space & Doors at end of each arm, windows on outer walls \\
        \hline
        14 & Residential building & 3 & U-shaped & - & - & 12 apartments (4 per floor) & Main section 16m x 8m, wings 6m x 4m, interior corridors & Main entrance at center, apartment doors along corridors \\
        \hline
        15 & Community center & 1 & Rectangular with rooms on three sides & - & - & 7 & Central hall 15m x 10m, 6 rooms 3m x 4m each & Doors connecting to central hall, windows on exterior walls \\
        \hline
        16 & Library & 2 & Rectangular (18m x 12m) & - & - & 5 & Open ground floor, walls on three sides & Windows in study rooms \\
        \hline
        17 & Building & 2 & Rectangular with extension & Multiple (main: pitched) & - & - & Main section 10m x 8m, extension 5m x 8m & - \\
        \hline
        18 & Pavilion & 1 & Octagon & Pyramidal & Wood & - & Each side 3m & Door on one side, windows on alternating sides \\
        \hline
        19 & Building & 2 & Stepped/Rectangular & Exposed lower roof & - & - & Ground floor 12m x 12m, upper floor 8m x 8m, terrace & Door access to terrace \\
        \hline
        20 & Building & 1 & Rectangular (40m x 25m) & - & Concrete & - & Figure-8 circulation, two loops, interior courtyards & - \\
        \hline
        21 & Apartment building & 3 & - & - & - & 9 units & Different layout per floor & Entrance door per unit, at least two windows per unit \\
        \hline
        22 & Building & 3 & Variable (Rect/L/T) & - & - & - & Ground floor 15m x 10m, different shapes per floor & - \\
        \hline
        23 & Office building & 2 & Rectangular (16m x 16m) & - & - & - & 4m x 4m open atrium (uncovered), corridors around void, perimeter offices & Windows on exterior walls \\
        \hline
        24 & School & 1 & Rectangular (25m x 10m) & Pitched & Brick & 5 & 2m wide corridor, linear arrangement & Door to corridor per classroom, two windows per classroom \\
        \hline
        25 & House & 2 & Rectangular (~16m x 12m) & - & - & 12 & Complex interior walls, central hallway on upper floor & - \\
        \hline
        \hline
    \end{tabular}
    }
\end{table}

We asked four human experts to evaluate whether the generated buildings adhered to the requirement aspects described in Table~\ref{table:categorized_prompts}. The scoring scale ranges from 1 to -1. A score of 1 indicates that the generated model fully meets the requirement defined in the prompt. A score of 0.5 means the model partially meets the requirement. A score of 0 is given when the requirement was not defined in the prompt, but the system still produced a meaningful result. A score of -0.5 indicates that the requirement was neither defined in the prompt nor generated by the system. A score of -1 means the generated model does not meet the requirements specified in the prompt.

\begin{figure}[h!]
\centering
\includegraphics[width=\textwidth]{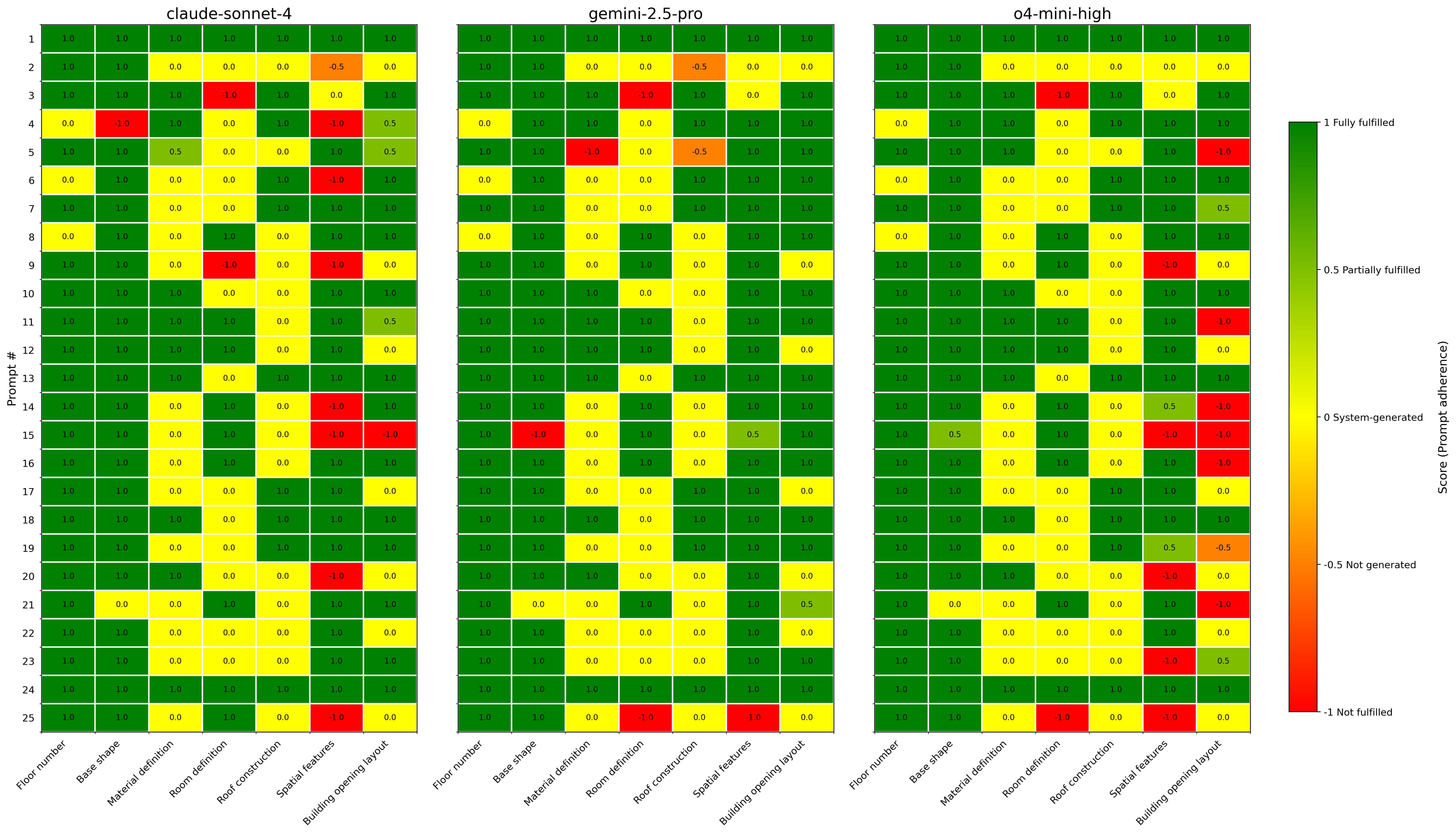}
\caption{Evaluate the generated models against the requirements specified in the test prompts. 0 indicates that the requirement was not specified in the prompts, but LLM agents generated relevant results. -0.5 indicates that the agents did not generate any relevant results for the unspecified requirement.}
\label{fig:heatmap}
\end{figure}

To resolve discrepancies among experts' evaluations, we adopted the majority rule approach for inconsistent ratings and held additional discussions. The final evaluation results are visualized in Fig.~\ref{fig:heatmap}. Regardless of the LLMs used, the BIM models generated by the framework were generally able to effectively fulfill the requirements defined in prompts, as indicated by the predominance of green cells. Even for open-ended instructions that do not explicitly specify some requirements, our framework can augment and complete the original prompt input, as reflected in the distribution of yellow cells.

Among the three LLMs, Gemini-2.5-pro demonstrates the best adherence to prompt requirements, as indicated by its lower number of red cells. In contrast, o4-mini-high sometimes shows hallucinations, which may be due to its comparatively smaller model size, and particularly struggles with accurately placing openings. When comparing different prompts, it is evident that all LLMs make errors with prompts 3, 15 and 25, which involve complex layouts with numerous internal rooms. This reveals that their spatial reasoning and understanding capabilities still need improvement. This is also reflected across various demand dimensions on the x-axis, such as room definition, spatial features, and building opening layout, where strong spatial understanding is required and red or light green cells frequently appear in the heatmap.

\subsubsection{\textbf{Design evaluation/Turing test}}
\label{sec:design_evaluation}

\begin{figure}[h!]
\centering
\includegraphics[width=\textwidth]{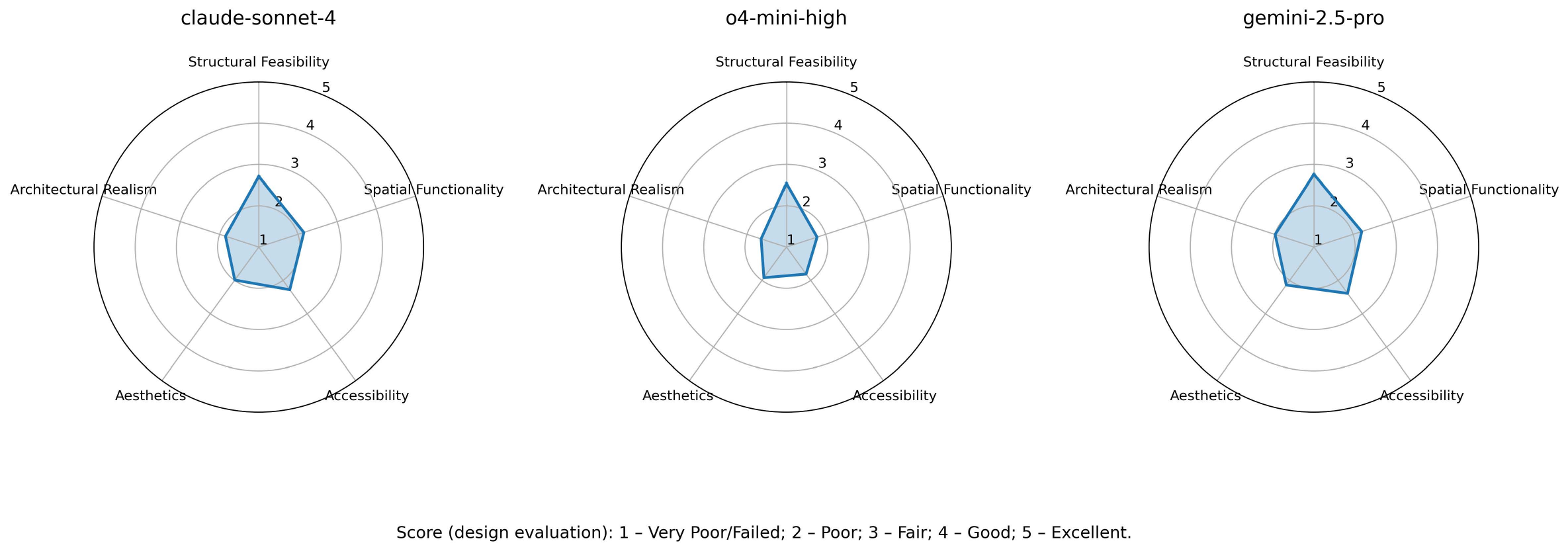}
\caption{Experts evaluate the generated BIM models from the perspective of real-world design, without taking into account limitations of proposed system or input prompts.}
\label{fig:radar}
\end{figure}

Human experts also performed rigorous design evaluations and Turing tests on the generated models. Crucially, these assessments disregarded the system's generative scope and specific prompt instructions. Instead, focusing on the general context of building types (e.g., hotels, apartments, community centers) and the early design stage, experts evaluated the models from a real-world project standpoint. Specifically, the human experts assessed structural feasibility, spatial functionality, accessibility, aesthetics, and architectural realism to determine if the generated models were distinguishable from those created by human architects. Assessments were based on a 5-point Likert scale measuring satisfaction (1 to 5). For each of the three LLMs, average ratings provided by all experts for all cases were computed, with the results visualized in Fig. \ref{fig:radar}.

Evaluation results indicate that while the generated models adhere to prompt-defined requirements, a significant gap remains between them and practical, real-world designs. This can be attributed to current system limitations, specifically its inability to generate essential building components like stairs, columns, beams, etc. Furthermore, experts commonly noted that LLMs still struggle with the logical placement of doors and windows for spatial accessibility and circulation. The lack of grounding in building codes and structural principles means many generated models exhibit structural flaws, such as excessive unsupported roof overhangs, inadequately thin walls, and floor slab connections incapable of supporting loads. This also indicates the limitation of our rule-checking-based quality optimization loop: while it can check and resolve basic geometric conflicts and semantic omissions in BIM models, it falls significantly short of encompassing the compliance and structural design principles crucial for real-world projects. Future work will thus concentrate on integrating more construction-specific knowledge into the generation process. 

Nevertheless, experts also highlighted that for some complex prompts, such as 20 and 23, the floor plans generated by Gemini were surprisingly realistic. This suggests its potential for generating initial layout proposals for complex projects.

\subsection{Code generation quality}

In addition to the aforementioned methods for evaluating the generated BIM models, we introduce an additional quantitative metric to assess the accuracy of the framework's direct outputs (i.e., the generated Python code). Specifically, we adopt CodeBERTScore \cite{zhou-etal-2023-codebertscore}, a widely used metric for evaluating code generation quality. This metric assesses the accuracy of generated code by calculating its cosine similarity with ground truth code samples. It leverages CodeBERT \cite{feng-etal-2020-codebert}, a pretrained model for programming languages, to compute contextual embeddings of tokens, thereby capturing the semantic equivalence between the generated and reference code pairs.

We first define a ground-truth reference code for each test case to ensure that the reference code fully meets the input requirements and can generate accurate and reliable BIM models. Then, for each LLM and each test case, we compute the F1 score of the CodeBERTScore between the candidate code generated by the framework and the reference code, providing a quantitative evaluation of the accuracy of the generated code.

CodeBERTScore calculates sentence-level precision by averaging the maximum similarity scores for each candidate vector and calculates recall by averaging the maximum similarity scores for each reference vector. Intuitively, recall is high if every vector in the reference code has a high cosine similarity with at least one vector in the candidate code, and precision is high if every vector in the candidate code has a high cosine similarity with at least one vector in the reference code. The final score is the F1 score, which is the harmonic mean of precision and recall, as illustrated in Eq.\ref{eq:codebert} \cite{zhou-etal-2023-codebertscore}.

\begin{equation} \label{eq:codebert}
\text{CodeBERTScore}_{F_1} = \frac{2 \cdot \text{CodeBERTScore}_{Precision} \cdot \text{CodeBERTScore}_{Recall}}{\text{CodeBERTScore}_{Precision} + \text{CodeBERTScore}_{Recall}}
\end{equation}

The results are summarized in Table \ref{tab:llm-results}. It can be observed that all three LLMs can generate code that is relatively accurate and aligned with the reference code through our framework. A reasonable range of 85\% to 95\% also reflects the fact that different LLMs can generate code with similar intent and functionality but with stylistic differences, much like how generated BIM models may vary in form while serving the same test prompt.

\begin{table}[ht!]
\centering
\caption{Summary of mean F1 CodeBERTScore and standard deviation (SD) for test cases across three LLMs.}
\label{tab:llm-results}

\renewcommand{\arraystretch}{1.3}
\begin{tabular}{c|cc|cc|cc}
\hline
\hline
\multirow{2}{*}{Prompt Nr.} & \multicolumn{2}{c|}{Claude-sonnet-4} & \multicolumn{2}{c|}{o4-mini-high} & \multicolumn{2}{c}{Gemini-2.5-pro} \\
\cline{2-7}
 & Mean$\uparrow$ & SD$\downarrow$ & Mean$\uparrow$ & SD$\downarrow$ & Mean$\uparrow$ & SD$\downarrow$ \\
\hline
1 & 0.8431 & 0.0048 & 0.8492 & 0.0132 & 0.8456 & 0.0045 \\
2 & 0.8456 & 0.0017 & 0.8518 & 0.0182 & 0.8273 & 0.0237 \\
3 & 0.9566 & 0.0101 & 0.8418 & 0.0083 & 0.8608 & 0.0138 \\
4 & 0.8509 & 0.0063 & 0.8604 & 0.0180 & 0.9138 & 0.0104 \\
5 & 0.9167 & 0.0060 & 0.8267 & 0.0206 & 0.8529 & 0.0172 \\
6 & 0.8553 & 0.0016 & 0.8596 & 0.0016 & 0.9015 & 0.0019 \\
7 & 0.8699 & 0.0041 & 0.8673 & 0.0111 & 0.8965 & 0.0121 \\
8 & 0.8780 & 0.0027 & 0.8771 & 0.0159 & 0.9218 & 0.0145 \\
9 & 0.8412 & 0.0222 & 0.8136 & 0.0058 & 0.8787 & 0.0088 \\
10 & 0.9437 & 0.0124 & 0.8625 & 0.0123 & 0.8870 & 0.0014 \\
11 & 0.9093 & 0.0008 & 0.8673 & 0.0033 & 0.8736 & 0.0031 \\
12 & 0.9379 & 0.0073 & 0.8465 & 0.0143 & 0.8787 & 0.0070 \\
13 & 0.8458 & 0.0255 & 0.8554 & 0.0113 & 0.8332 & 0.0116 \\
14 & 0.8208 & 0.0063 & 0.8398 & 0.0160 & 0.8779 & 0.0058 \\
15 & 0.8597 & 0.0174 & 0.8195 & 0.0406 & 0.8860 & 0.0152 \\
16 & 0.8616 & 0.0092 & 0.8554 & 0.0092 & 0.8936 & 0.0224 \\
17 & 0.8593 & 0.0068 & 0.8587 & 0.0359 & 0.9240 & 0.0021 \\
18 & 0.8291 & 0.0220 & 0.8832 & 0.0062 & 0.8464 & 0.0111 \\
19 & 0.9427 & 0.0038 & 0.8516 & 0.0152 & 0.8712 & 0.0047 \\
20 & 0.8574 & 0.0067 & 0.8401 & 0.0262 & 0.9009 & 0.0084 \\
21 & 0.9397 & 0.0185 & 0.8443 & 0.0019 & 0.8497 & 0.0164 \\
22 & 0.8307 & 0.0099 & 0.8459 & 0.0209 & 0.8554 & 0.0025 \\
23 & 0.8552 & 0.0062 & 0.8501 & 0.0099 & 0.9263 & 0.0012 \\
24 & 0.8494 & 0.0263 & 0.8653 & 0.0019 & 0.8664 & 0.0068 \\
25 & 0.9264 & 0.0019 & 0.8526 & 0.0145 & 0.8709 & 0.0031 \\
\hline\hline
\end{tabular}
\end{table}

%\end{color}

\section{Novelty and contributions}

The main contribution of this paper is the proposal of an LLM-based multi-agent framework capable of generating semantically rich BIM models from textual descriptions, bridging the gap left by prior research which only produced purely geometric models. This paper addresses three key scientific challenges: (1) creating concise yet effective textual representations for 3D BIM models; (2) conditioning LLMs to utilize domain-specific knowledge to avoid hallucinations in outputs; (3) designing a system that orchestrates multiple LLM agents for enhanced logical reasoning and spatial awareness in complex 3D modeling tasks. The proposed framework is designed to be generic and can be implemented in various BIM authoring software by utilizing their respective APIs, as well as being adapted to different LLMs.

Beyond the foundational contributions to the body of knowledge, the paper also offers new insights into adapting LLMs within the AEC domain. Our work highlights the potential of applying LLMs to complex 3D modeling tasks, moving beyond traditional NLP applications. By integrating LLM-driven collaboration into commercial BIM platforms, we introduce the concept of "modeling-by-chatting", offering natural language interactions that transcend traditional command- or script-based approaches in BIM authoring. Although demonstrated through architectural use cases, the agentic framework can be generalized to other tasks (e.g., building energy analysis) by incorporating specialized rules and tool functions.

Crucially, we integrate rule-based domain checks with large-scale generative models, combining rigorous engineering logic with advanced creativity. This synergy addresses limitations in traditional expert systems and purely generative methods, providing new insights into neuro-symbolic AI development in our field. Compared with other LLM-based multi-agent frameworks, our approach employs a code-centric and domain-aware feedback loop, where agents iteratively generate and refine executable BIM code, ensuring design rationality and compliance. This significantly differentiates our method from existing approaches.

\section{Discussion and future works}

While our use cases focus on practical challenges in the BIM modeling process, the core of this paper lies in proposing a multi-agent framework that effectively orchestrates and conditions different LLMs. Through complex inter-agent communication mechanisms and the integration of domain knowledge, our framework effectively conditions the inherent randomness of these generative models, enabling them to autonomously and collaboratively generate 3D buildings that adhere to architectural principles. 

\subsection{Generalization and scope}

The proposed framework is designed to be flexible and adaptable to various BIM authoring software through the use of their APIs. By utilizing LLMs to generate Python scripts that call custom tool functions (e.g., create\_wall, set\_wall\_height, etc.), the framework abstracts low-level APIs into reusable components, allowing easy adaptation for other BIM authoring software by encapsulating their respective APIs. Additionally, tool functions are designed based on insights from large-scale software usage logs, ensuring coverage of essential BIM modeling operations common across platforms. The framework's agent roles and prompt templates are universally applicable to BIM modeling tasks, and its model-checking process is based on the vendor-neutral IFC standard, further ensuring broad compatibility.

%\begin{color}{blue}
The implemented framework is currently capable of generating regular, non-curved building models in the early design stage based on a limited scope of building components defined by the available toolset. To generalize this approach to irregularly shaped buildings or more detailed engineering models with higher LOD, the development of more complex tools for the agents is required to significantly expand the existing limited toolset. However, a challenge arises in organizing and managing the vast amount of tool information and their interdependencies, so that the LLM can efficiently retrieve useful functions. Knowledge graphs and graph-based Retrieval Augmented Generation (RAG) techniques might offer potential solutions. Furthermore, given the complexity of engineering models, especially the intricate connections and construction details among various components, future work also requires enhancing the spatial understanding of the engineering details of the system, improving the logical reasoning for more complex structures, and ensuring that existing methods can accurately represent such details.

\subsection{Architectural condition and regulations}
%\end{color}

In the current framework, the \textsc{Architect} agent generates structured text-based building designs that include numerical information such as coordinates and dimensions. This content format is designed to be aligned with the input requirements of the tool functions, allowing downstream agents to better understand and utilize specific architectural parameters in the code. It is observed in the experiments that this approach is more robust and accurate than using formats like SVG (XML) or images for representing floor plans. 

%\begin{color}{blue}
Moreover, in the proposed system design, the \textsc{Architect} agent creates the building's layout based on general knowledge provided by the foundation model (during training it certainly processed architectural texts), the few-shots learning example and the nine basic architectural rules provided in prompt templates. Although some of the generated layouts appear visually reasonable, they lack comprehensive consideration of complex architectural conditions (e.g., lighting, functionality, accessibility, etc.) and regulations (e.g., fire safety, area requirements, etc.). This limitation is also reflected in the rigorous design evaluation conducted by human experts in Section \ref{sec:design_evaluation}. Although the generated models follow the requirements defined in input prompts, they receive low scores in terms of practical applicability to real-world architectural models. The primary reason for not including an exhaustive set of architectural conditions and regulations at this stage was to keep the scope manageable and to focus on verifying the conceptual feasibility of LLM-driven BIM model generation, as incorporating building codes (such as fire safety codes, disability access requirements, detailed HVAC guidelines, or lighting and acoustical standards) would dramatically expand the complexity of the prompts, and require more complex system design of the LLM-based agentic framework. 
Accordingly, future work will explore how to effectively integrate this complex architectural knowledge into LLMs, for instance, by fine-tuning the model with curated, regulation-focused data and design guidelines.
%\end{color}

\subsection{Autonomous clash resolution and spatial reasoning}
\begin{figure}[h!]
\centering
\includegraphics[width=0.95\textwidth]{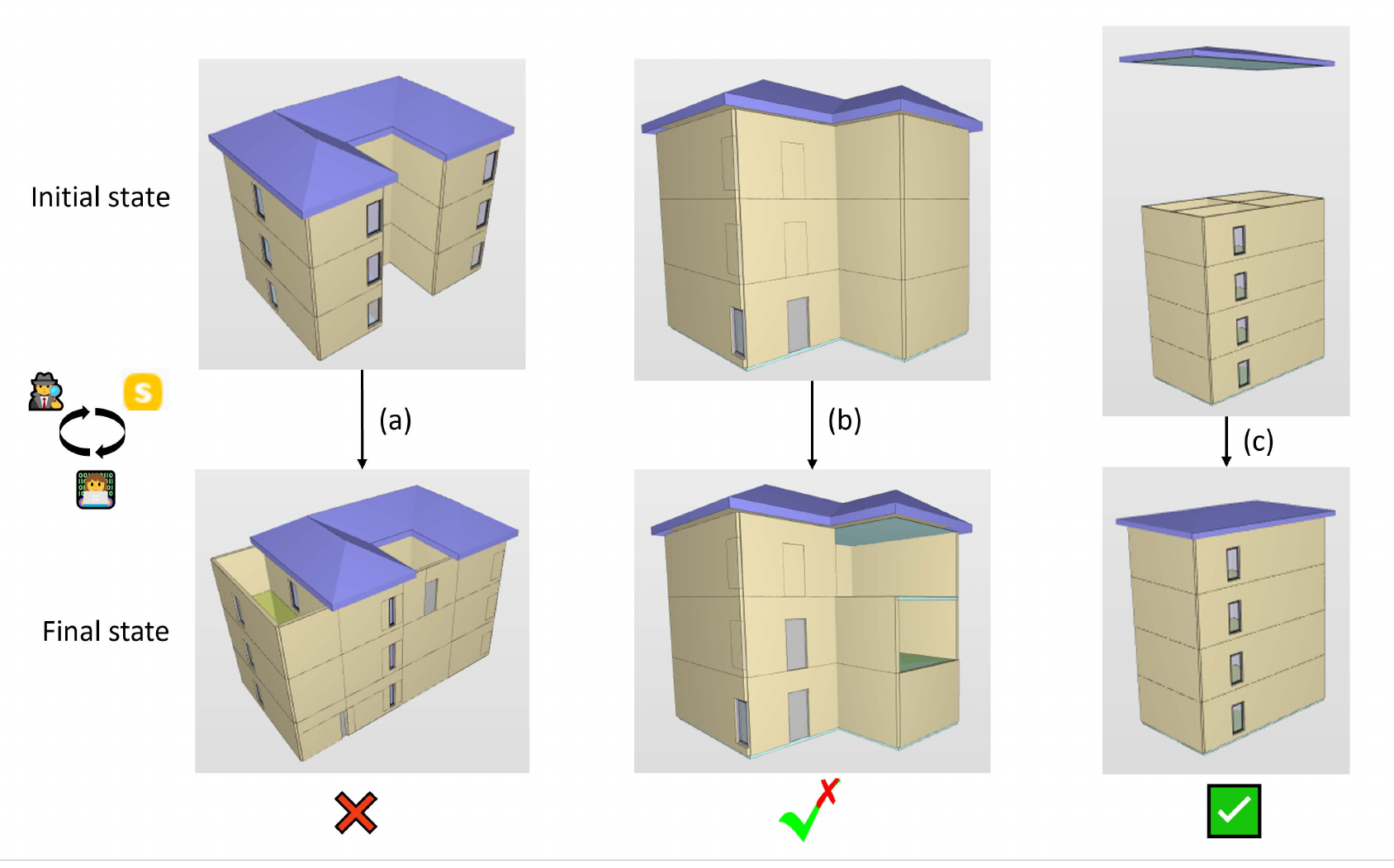}
\caption{Several representative cases encountered during the quality optimization loop}
\label{fig:clashes}
\end{figure}

Our experiment demonstrates that LLM agents can automatically resolve clashes within the model to a limited extent through the designed quality optimization loop. Although this module is not the main focus of this study, our preliminary exploration in this direction presents a new technical approach for research in related fields. This is particularly significant considering that current research on automatic clash resolution mainly focuses on using optimization algorithms \cite{Wu:2023:EGICE}, classical machine learning \cite{doi:10.1061/JCCEE5.CPENG-5548}, or reinforcement learning \cite{doi:10.1061/9780784483961.071}.
Despite these advancements, the conflict resolution method based on LLM agents still has significant limitations. Fig.~\ref{fig:clashes} summarizes some representative scenarios encountered during the quality optimization loop. The first common failure (a) involves the agent attempting to rewrite code to create a new model, leading to an increase in issue amount due to conflicts between the new and existing model components. In scenario (b), the upper two floors of the initial model have overlapping and nested walls, doors, and windows. In such highly complex situations, LLM agents, which rely solely on code and checker feedback (rule/issue descriptions) for contextual information, cannot resolve all the issues and are prone to hallucinations. The strategy the agent adopts here involves deleting parts of the walls on the relevant floors. While this action can reduce the overall issue amount in the model, it compromises the structural integrity of the building. However, current agents can only perceive information from 1D text and are not yet capable of understanding 3D spaces in this manner. Scenario (c) illustrates a successful case where the agent correctly adjusts the height of a floating roof to align with the top floor's wall elevation. Overall, LLMs perform well for intuitive issues with deterministic solutions (typically Class 1 rules, such as "no space defined in model -> create space"). However, they often fail on complex issues that require higher-level spatial understanding and have open-ended solutions (usually Class 3 rules, such as "two partition walls intersect -> which wall is to be moved, and in which direction?"). Although our framework allows users to guide the LLM to perform the appropriate issue-solving actions via dialogue or manually continue editing the BIM model generated in the software, future research will prioritize enhancing the LLM's spatial understanding capabilities to advance toward an autonomous conflict resolution system.

\subsection{Evaluation methods}
Given that our approach generates code representations of 3D models based on prompt engineering techniques, it does not require fine-tuning of LLMs. This is fundamentally different from conventional Text-to-3D methods, which typically require constructing a 3D dataset for training. Commonly used metrics such as Chamfer Distance (CD) and Intersection over Union (IoU) mainly focus on evaluating the geometric accuracy of point cloud/voxel models. 
%\begin{color}{blue}
As these metrics are not applicable to our data representation approach, we propose using the pass rate/issue amount of 30 domain-specific rule checks as a quantitative metric to evaluate the generated BIM models. 
As explained in the Section \ref{sec:prototype}, the chosen rules cover a good share of fundamental concepts of building design. At the same time, we agree that a higher number of rules can further improve the quality of generated models, for example, by incorporating knowledge from various domains such as fire protection. This study, however, focuses on general feasibility and system architecture, leaving more fine-grained improvements to future investigations and publications.

In addition, While the automated rule checking can verify whether the generated models meet certain geometric and semantic quality, its limitation lies in the fact that the rules provided by model checkers cannot assess whether the generated buildings align with the abstract and dynamic design requirements expressed in natural language instructions (e.g., "H-shaped house", "arrange rooms along the building perimeter", etc.). 
To address this, we apply human experts' assessments to determine whether the models adhered to the intended prompts, as well as design evaluation from a real-world project perspective. We further utilized CodeBERTScore to measure the similarity between the generated code and the reference code that satisfies the input requirements, enabling a quantitative evaluation of the framework's accuracy from a different perspective.
%\end{color}

\subsection{Hallucination and mitigation strategies}

Our work explores the use of LLMs to generate 3D BIM models. Essentially, this is achieved through a series of carefully designed mechanisms to minimize LLM hallucinations and ensure the desired output: Prompt templates define the necessary context and restrict the output format of the LLM; High-level tool functions encapsulate intricate low-level modeling APIs, reducing the coding difficulty for the LLM and thereby limiting hallucinations; The multi-agent setup breaks down complex modeling tasks into manageable sub-tasks, simplifying each LLM's role and reducing hallucinations; Multiple loops incorporate feedback from the model checker and Python interpreter, using this deterministic knowledge to guide the LLM in resolving issues caused by hallucinations due to its stochastic nature. 

Despite these efforts, we still observed some hallucinations occurring in our system during experiments.

\textbf{First}, getting lost in the middle \cite{liu2023lostmiddlelanguagemodels}, especially when the input prompt has a lengthy context, makes the LLM more susceptible to hallucinations. In our system, the \textsc{Instruction Enhancer} often needs to synthesize information from various aspects, which makes its input particularly long. This can lead to omissions in the guidance it provides to downstream agents. For instance, in \textbf{Appendix~\ref{app:chat_log}} (5), the \textsc{Architect} mentions creating an open functional space within the hexagonal boundary, but in (7), the \textsc{Instruction Enhancer} completely omits this point in output, leading to the absence of the spatial definition in the first version of the generated BIM model. Although the model optimization loop added the spaces (rooms) in the second round, the room design significantly differed from the \textsc{Architect}'s original intention. To address this issue, we could either optimize the framework by further breaking down the \textsc{Instruction Enhancer}'s task into smaller sub-tasks handled by more agents to reduce the input length, or use Retrieval Augmented Generation (RAG) techniques \cite{10.5555/3495724.3496517,edge2024localglobalgraphrag} to dynamically insert only the necessary contextual information into the prompt, rather than statically providing all information in the template.

\textbf{Second}, hallucinations arise from a lack of spatial understanding of the overall building. This can be observed during both the model generation and optimization phases. For example, in Fig.~\ref{fig:result10}, we intend to generate a door at the junction of two sections of a T-shaped building. Properly placing the door on the correct wall with accurate coordinates requires a certain level of understanding of the entire building space, which LLMs struggle with, leading to hallucinations. Future work could consider fine-tuning the LLM, introducing additional modalities \cite{zhu2024llava}, or using more advanced prompt engineering techniques \cite{li2024advancing} to enhance its spatial awareness. The model optimization phase often involves modifying the model. The current LLM agent only makes changes to specific local components based on feedback from the model checker, without considering the propagation of these modifications,i.e., the impact on adjacent components or even the entire model. This lack of a global perspective can cause hallucinations when the agent attempts to resolve highly complex conflicts, potentially creating more conflicts within the model. A potential solution could be, on one hand, to enhance the LLM's reasoning ability, such as by integrating tree search to explore possible solution spaces more comprehensively \cite{zhang2024restmctsllmselftrainingprocess}. On the other hand, we could introduce the graph-based representation of the BIM model \cite{doi:10.1061/JCCEE5.CPENG-5487,EC32024_247} to assist the LLM in understanding the spatial and topological relationships within the model, thereby better grasping the overall layout of the building.

\textbf{Third}, cumulative hallucinations. In our experiments, we observed that if the \textsc{Architect} generates incorrect coordinates due to hallucination, downstream agents will be affected, leading to numerous errors in the generated BIM model. In the future, we could consider introducing a review mechanism for the \textsc{Architect}'s outputs. Another approach could be using a more robust and flexible structure to organize the multi-agent system, drawing on recent developments \cite{wu2023autogenenablingnextgenllm} in the community.

\section{Conclusions}

We introduce Text2BIM, an LLM-based multi-agent collaborative framework that generates building models in BIM authoring software from natural language descriptions. The main findings and contributions of this study are as follows:
\begin{itemize}
    \item Unlike previous studies that focused on generating the 3D geometric representation of buildings, our framework is capable of producing native BIM models with internal layouts, external envelopes, and semantic information. 
    \item We propose representing 3D building models using imperative code that interacts with BIM authoring software APIs. By employing prompt engineering techniques combined with orchestration framework design, multiple LLM agents collaborate to develop the code without the need for fine-tuning, thereby conserving computational resources. 
    \item Innovatively, a domain-specific rule-based model checker is integrated into the framework to guide LLMs in generating architecturally and structurally rational outcomes. The proposed quality optimization loop demonstrates that the LLM agents can iteratively resolve conflicts within the BIM model based on textual feedback from the checker. 
    \item Extensive experiments have been conducted to comprehensively evaluate the performance of different modules within the proposed framework, including a comparative analysis of three different LLMs within the framework, which validates the generalizability and effectiveness of our approach.
    \item An interactive software prototype is developed that integrates the proposed framework into the BIM authoring tool Vectorworks, showcasing innovative possibilities for modeling-by-chatting during the design process.
\end{itemize}

%\begin{color}{blue}
While this paper provides a feasibility study that shows very promising first results, the current setup also has certain limitations. Although some of the generated layouts appear visually reasonable, they lack comprehensive consideration of complex architectural conditions (e.g., lighting, functionality, accessibility, etc.) and regulations (e.g., fire safety, area requirements, etc.). To further improve the quality of the generated models, the integration of more architectural and construction knowledge is required, including rules from the respective codes and regulations. These refinements will be subject to future investigations and publications. 
%\end{color}

We believe that the proposed methodology can be extended to a broader range of use cases beyond just model generation, especially if more specialized tools are developed for LLM agents to utilize. We hope that readers will find inspiration from this and explore using LLMs to address more challenges within our field.

\section{Data availability statement}

The code is available at: \url{https://github.com/dcy0577/Text2BIM}. Some data and models that support the findings of this study are available from the corresponding author upon reasonable request.

\section{Acknowledgments}

This work is funded by Nemetschek Group, which is gratefully acknowledged. We sincerely appreciate the data and licensing support provided by Vectorworks, Inc.

%
%\pagebreak
%
% Now we start the Appendixes, with the new section name, "Appendix", and a 
%  new counter, "I", "II", etc.
\appendix
%
% And now for some pretty impressive notation.  In this example, I have used
%   the tabular environment to line up the columns in ASCE style.
%   Note that this and all Appendixes (except the references) start with 
%   the \section command
\section{Toolset documentation}
\label{app:tool_set}
%\nopagebreak
% Please see Tables~\ref{table:tool_set_1} -~\ref{table:tool_set3}.
% start tables for toolset %
\begin{table}[H]
\caption{Toolset (a)}
\label{table:tool_set_1}
\centering
%\scriptsize
\footnotesize
\renewcommand{\arraystretch}{1.25}
\begin{tabular}{c p{14cm}}
\hline\hline
\multicolumn{1}{c}{Tool function name} &
\multicolumn{1}{c}{Description} \\
\hline
create\_story\_layer & This tool is used to create a new story layer. The new layer is created at the given elevation. Once a new story layer is created, it becomes the active layer. All new building elements will be created on the current active story. \newline \textbf{Input:} \newline - layer\_name: str, the unique name of the new story. \newline - elevation: float, the elevation of the new story relative to the ground. \newline - floor\_index: int, the index of the new floor. Should start from 1. \newline \textbf{Return:} \newline - str, the layer\_uuid of the new story layer. \\
\hline
set\_active\_story\_layer & This tool is used to set the story layer with the given name to active. The active story layer is the layer in which new elements are created. \newline \textbf{Input:} \newline - layer\_name: str, the name of the layer to set as active. \newline \textbf{Return:} \newline - str, the layer\_uuid of the active layer. \\
\hline
create\_functional\_area & This tool is used to create a conceptual functional area on a specified layer. The area is created from a list of vertices that define the room boundary. Usually, functional areas are created first to define the interior layout of the building, and then the rooms are separated by placing walls at the boundaries. \newline \textbf{Input:} \newline - vertices: list of tuples, each tuple represents the 2D coordinate of a vertex that defines the boundary of the room. \newline - name: str, the name of the room/functional area. \newline - layer\_uuid: str, the uuid of the story layer where the space will be created. \newline \textbf{Return:} \newline - str, the uuid of the created room/functional area. \\
\hline
create\_wall & This tool is used to create a wall on a specified layer. By default, the wall is created with a bottom\_elevation of 0 and a top\_elevation of 3000 relative to this layer. \newline \textbf{Input:} \newline - st\_pt: tuple, the 2D coordinate of the starting point of the wall. \newline - ed\_pt: tuple, the 2D coordinate of the end point of the wall. \newline - layer\_uuid: str, the uuid of the story layer where the wall will be created. \newline \textbf{Return:} \newline - str, the uuid of the newly created wall. \\
\hline
set\_wall\_thickness & This tool is used to set the thickness of a wall. \newline \textbf{Input:} \newline - uuid: str, the uuid of the wall object. \newline - thickness: float, the new thickness of the wall. \newline \textbf{Return:} \newline - str, the uuid of the wall object that has been modified. \\
\hline
set\_wall\_elevation & This tool is used to set the top/bottom elevation of a wall. Subtracting these two is the height of the wall itself. \newline \textbf{Input:} \newline - uuid: str, the uuid of the wall object. \newline - top\_elevation: float, the vertical distance from the top of the wall to the story layer where the wall was originally created. \newline - bottom\_elevation: float, the vertical distance from the bottom of the wall to the story layer where the wall was originally created. \newline \textbf{Return:} \newline - str, the uuid of the wall object that has been modified. \\
\hline
get\_wall\_elevation & This tool is used to get the top and bottom elevation of a wall. Subtracting these two is the height of the wall itself. \newline \textbf{Input:} \newline - uuid: str, the uuid of the wall object. \newline \textbf{Return:} \newline - top\_elevation: float, the vertical distance from the top of the wall to the story layer where the wall was originally created. \newline - bottom\_elevation: float, the vertical distance from the bottom of the wall to the story layer where the wall was originally created. \\
\hline
get\_wall\_thickness & This tool is used to get the thickness of a wall. \newline \textbf{Input:} \newline - uuid: str, the uuid of the wall object. \newline \textbf{Return:} \newline - thickness: float, the thickness of the wall. \\
\hline\hline
\end{tabular}
\end{table}

\begin{table}[H]
\caption{Toolset (b)}
\label{table:tool_set2}
\centering
%\scriptsize
\footnotesize
\renewcommand{\arraystretch}{1.25}
\begin{tabular}{c p{14cm}}
\hline\hline
\multicolumn{1}{c}{Tool function name} &
\multicolumn{1}{c}{Description} \\

\hline
set\_wall\_style & This tool is used to set the style of a wall. \newline \textbf{Input:} \newline - uuid: str, the uuid of the wall object. \newline - style\_name: str, the name of the style. Following wall style names are available: ["Concrete Wall", "Wood Wall", "Brick Wall"] \newline \textbf{Return:} \newline - str, the uuid of the wall object that has been modified. \\
\hline
add\_window\_to\_wall & This tool is used to add a window to a wall. Once a window is added to a wall, it is part of the wall and will be moved/duplicated/rotated with the wall. \newline \textbf{Input:} \newline - wall\_uuid: str, the uuid of the wall object to which the window will be added. \newline - window\_elevation: float, the elevation of the window from the bottom of the wall. \newline - window\_offset: float, the offset of the window from the starting point of the wall. \newline - window\_name: str, the name of the window object to be added. \newline \textbf{Return:} \newline - str, the uuid of the window object that has been added to the wall. \\
\hline
add\_door\_to\_wall & This tool is used to add a door to a wall. Once a door is added to a wall, it is part of the wall and will be moved/duplicated/rotated with the wall. \newline \textbf{Input:} \newline - wall\_uuid: str, the uuid of the wall object to which the door will be added. \newline - door\_elevation: float, the elevation of the door from the bottom of the wall. \newline - door\_offset: float, the offset of the door from the starting point of the wall. \newline - door\_name: str, the name of the door object to be added. \newline \textbf{Return:} \newline - str, the uuid of the door object that has been added to the wall. \\
\hline
move\_obj & This tool is used to move an element. It can only move the given element within the layer where it is placed but not duplicate it. \newline \textbf{Input:} \newline - uuid: str, the unique uuid of the element to move. \newline - xDistance: float, moving distance in x direction. \newline - yDistance: float, moving distance in y direction. \newline - zDistance: float, moving distance in z direction. \newline \textbf{Return:} \newline - None \\
\hline
delete\_element & This tool is used to delete an element or a list of elements. Story layers cannot be deleted. \newline \textbf{Input:} \newline - uuid: str or a list of string, the unique uuids of the elements to delete. \newline \textbf{Return:} \newline - None \\
\hline
find\_selected\_element & This tool is used to find the selected element in the current active story layer. If no selected elements are found, it will return an empty list. \newline \textbf{Input:} \newline - None \newline \textbf{Return:} \newline - list of str, the uuids of the selected elements. \\
\hline
create\_polygon & This tool is used to create a polygon on a specified story layer using its vertices. \newline \textbf{Input:} \newline - vertices: list of tuples, each tuple represents the 2D coordinate of a vertex of the polygon. \newline - layer\_uuid: str, the uuid of the story layer where the polygon will be created. \newline \textbf{Return:} \newline - str, the uuid of the created polygon. \\
\hline
get\_polygon\_vertex & This tool is used to get a desired vertex at the given index in the polygon's vertex array. \newline \textbf{Input:} \newline - uuid: str, the uuid of the polygon object. \newline - at: int, the index of the desired vertex. \newline \textbf{Return:} \newline - tuple, the 2D coordinate of the desired vertex of the polygon. \\
\hline
get\_vertex\_count & This tool is used to get the number of vertices in a polygon. \newline \textbf{Input:} \newline - uuid: str, the uuid of the polygon object. \newline \textbf{Return:} \newline - int, the number of vertices in the input polygon. \\

\hline\hline
\end{tabular}
\end{table}

\begin{table}[H]
\caption{Toolset (c)}
\label{table:tool_set3}
\centering
%\scriptsize
\footnotesize
\renewcommand{\arraystretch}{1.25}
\begin{tabular}{c p{14cm}}
\hline\hline
\multicolumn{1}{c}{Tool function name} &
\multicolumn{1}{c}{Description} \\
\hline
create\_slab & This tool is used to create a slab from a polygon profile on a specified layer. \newline \textbf{Input:} \newline - profile\_id: str, the uuid of a polygon object that determines the profile of the slab. \newline - layer\_uuid: str, the uuid of the story layer where the slab will be created. \newline \textbf{Return:} \newline - str, the uuid of the created slab. \\
\hline
set\_slab\_height & This tool is used to set the height (elevation) of a slab. \newline \textbf{Input:} \newline - slab\_id: str, the uuid of the slab object. \newline - height: float, the height of the slab relative to the story layer where the slab was originally created. \newline \textbf{Return:} \newline - str, the uuid of the modified slab. \\
\hline
get\_slab\_height & This tool is used to get the height (elevation) of a slab. \newline \textbf{Input:} \newline - slab\_id: str, the uuid of the slab object. \newline \textbf{Return:} \newline - float, the height of the slab relative to the story layer where the slab was originally created. \\
\hline
set\_slab\_style & This tool is used to set the style of a slab. \newline \textbf{Input:} \newline - slab\_id: str, the uuid of the slab object. \newline - style\_name: str, the name of the style. \newline \textbf{Return:} \newline - str, the uuid of the modified slab. \\
\hline
duplicate\_obj & This tool is used to duplicate an element to a specified layer. Note that when duplicating a wall that includes doors and windows, the doors and windows within it will also be duplicated. The story layer cannot be duplicated. \newline \textbf{Input:} \newline - element\_uuid: str, the unique uuid of an element to duplicate. \newline - layer\_uuid: str, the uuid of the story layer where the copies will be placed. \newline - n: int, the number of copies to make. \newline \textbf{Return:} \newline - list of str, the list of uuids of the copies. It is recommended to use this list to further manipulate the copies. \\
\hline
rotate\_obj & This tool is used to rotate an element. \newline \textbf{Input:} \newline - uuid: str, the unique uuid of the element to rotate. \newline - angle: float, the angle in degrees to rotate the element. \newline - center: tuple, the 2D coordinate of the center of rotation. By default, it is the center of the element. (optional) \newline \textbf{Return:} \newline - str, the uuid of the rotated element. \\
\hline
create\_pitched\_roof & This tool is used to create a pitched roof from a polygon profile on a specified layer. \newline \textbf{Input:} \newline - profile\_id: str, the uuid of a polygon object that determines the profile(base) of the roof. \newline - layer\_uuid: str, the uuid of the story layer where the roof will be created. \newline - slope: float, the slope of the roof in degrees. It cannot be less than 5. \newline - eave\_overhang: float, the eave overhang of the roof. \newline - eave\_height: float, the elevation of the roof relative to the specified layer. Usually the height of the wall on this floor. \newline - roof\_thickness: float, the thickness of the roof. \newline \textbf{Return:} \newline - str, the uuid of the created roof. \\
\hline
set\_pitched\_roof\_attributes & This tool is used to set the new attributes of a pitched roof. Attributes that need to be changed can be optionally entered. \newline \textbf{Input:} \newline - roof\_id: str, the uuid of the roof object. \newline - slope: float, the slope of the roof in degrees (optional). \newline - eave\_overhang: float, the eave overhang of the roof (optional). \newline - eave\_height: float, the height(elevation) of the roof from the story layer where the roof was originally created (optional). \newline - roof\_thickness: float, the thickness of the roof (optional). \newline \textbf{Return:} \newline - str, the uuid of the modified roof. \\
\hline
set\_pitched\_roof\_style & This tool is used to set the style of a pitched roof. \newline \textbf{Input:} \newline - roof\_id: str, the uuid of the roof object. \newline - style\_name: str, the name of the style. Available: ["Low Slope Concrete w/ Rigid Insulation", "Sloped Wood Struct Insul Flat Clay Tile"] \newline \textbf{Return:} \newline - str, the uuid of the modified roof. \\
\hline\hline
\end{tabular}
\end{table}

% end tables for toolset %

\section{Prompt templates}
\label{app:prompts}

\lstset{
    basicstyle=\linespread{1}\ttfamily\small,  
    frame=single,                     
    breaklines=true,    
    breakatwhitespace=false, 
    captionpos=t,
    breakindent=0pt,
    escapeinside={@}{@},
}
\renewcommand\lstlistingname{Listing}
\renewcommand\lstlistlistingname{}

\begin{lstlisting}[language=TeX, caption={Prompt template of the \textsc{Architect} agent. Placeholders with "<< >>" indicate the dynamic content that can be inserted into the template, which includes a paraphrase of the original user instruction.}, label={lst:arch_pt}]
You are an experienced architect who can design floors/buildings based on the user's needs. You will use your extensive architectural knowledge to expand and supplement the user's original description and ultimately express your design in structured text format.
Depending on the user's specific needs, try to include in the output the starting and ending points of each wall, the location of windows and doors (offset relative to the start of the wall), boundaries of interior rooms/functional areas and the position and geometric details of other components required for a complete building.

Please refer to basic architectural rules, such like: 
- Foundation: Ensure a solid foundation slab that can support the entire structure.
- Wall Configuration: Arrange walls to define the building's perimeter and internal spaces. Ensure that load-bearing walls are adequately spaced and placed to distribute the weight of the structure evenly. Set the wall elevations properly for each floor.
- Slab Design: Place slabs for each floor. They should be level and supported by the walls, providing stability and separating different floors.
- Roof Construction: Design the roof to cover the entire building, protecting it from weather elements.
- Window Placement: Install windows strategically to provide natural light and ventilation to rooms. Ensure window locations are proportionate to the room size.
- Door Placement: Position doors for easy access to different rooms and areas. Main entrance to the building should be prominent and easy to locate, with interior doors facilitating smooth movement.
- Interior Layout: Organize and define the interior room layout logically. Use interior walls to separate different functional rooms and ensure easy flow between them with appropriately placed doors.
- Structural Integrity: Ensure all elements (walls, slabs, roof) are securely connected and stable.
- Compliance: Avoid clashing/overlapping building components, such as overlapping partitions between different areas and overlapping window and door locations. Adjacent rooms can share internal partitions. Rooms can also utilize exterior walls.

Make your design spatially and geometrically rational. Use millimeter units. Minimize other prose.

Here is a sample conversation:
"""
User: I want to build an office building. I want the building to have 3 floors and the layout of each floor to be the same. Each floor has 6 rooms, 3 on each side, separated by a 3 meter wide corridor. Each room has a door and a window. The door to each room should be on the wall on one side of the corridor and the window should be on the outside wall of the building.

Architect: **3-Floor Office Building Design**
**Foundation:**
- Rectangular foundation slab: 30000mm x 15000mm
**Ground Floor Plan:**
1. **Perimeter Walls:**
- Wall A: (0,0) to (30000,0)
- Wall B: (30000,0) to (30000,15000)
- Wall C: (30000,15000) to (0,15000)
- Wall D: (0,15000) to (0,0)
2. **Functional Areas**
Boundary in format (x_min,y_min),(x_max,y_max):
- Room 1: (0,0), (10000,6000)
- Room 2: (10000,0), (20000,6000)
- Room 3: (20000,0), (30000,6000)
- Room 4: (0,9000), (10000,15000)
- Room 5: (10000,9000), (20000,15000)
- Room 6: (20000,9000), (30000,15000)
- Central corridor: (0,6000), (30000,9000)
3. **Internal Corridor Walls:**
- Wall E: (0,6000) to (30000,6000)
- Wall F: (0,9000) to (30000,9000)
4. **Room Dividing Walls:**
- Wall G: (10000,0) to (10000,6000)
- Wall H: (20000,0) to (20000,6000)
- Wall I: (10000,9000) to (10000,15000)
- Wall J: (20000,9000) to (20000,15000)
5. **Doors:**
Insertion offset of each room door relative to the start of the corresponding wall:
- Room 1 door on corridor wall E: 5000
- Room 2 door on corridor wall E: 15000
- Room 3 door on corridor wall E: 25000
- Room 4 door on corridor wall F: 5000
- Room 5 door on corridor wall F: 15000
- Room 6 door on corridor wall F: 25000
6. **Windows:**
Insertion offset of each room window relative to the start of the corresponding wall:
- Room 1 window on wall A: 5000
- Room 2 window on wall A: 15000
- Room 3 window on wall A: 25000
- Room 4 window on wall C: 5000
- Room 5 window on wall C: 15000
- Room 6 window on wall C: 25000
**First Floor Plan:**
- Identical to Ground Floor Plan
**Second Floor Plan:**
- Identical to Ground Floor Plan
**Roof Construction:**
- Roof covering entire building: (0,0) to (30000,0) to (30000,15000) to (0,15000) to (0,0).
**Slab Design:**
- Create slabs for each floor supported by perimeter and internal walls. Slabs covering entire floor area with same dimensions as foundation.
**Summary:**
- Building dimensions: 30000mm x 15000mm x 3 floors.
- Each floor has 6 rooms, 3 on each side of a central corridor.
"""

The user now provides the following instruction, please generate the plan as an architect. Let's think step by step.
User: <<task>>
Architect: 
\end{lstlisting}

\newpage
\begin{lstlisting}[language=TeX, caption={Prompt template of the \textsc{Instruction Enhancer} agent. Placeholders with "<< >>" indicate the dynamic content that can be inserted into the template, including the information on available tool functions, chat history, and the original user instruction.}, label={lst:ie_pt}]
You are an experienced instruction enhancer who can instruct an AI programmer to write Python scripts in Vectorworks to accomplish user-specified tasks. You are a master at understanding API documentation, adept at breaking down the user's original instructions into subtasks and sub-logic based on the functions available, and able to express the user's description in a more rigorous and detailed manner.

Please give detailed hints on coordinates and dimensions when instructing the programmer. Consult the Architect when you feel you need a building design plan or more architectural context to assist your instructions. If functional areas are given in the design plan, make sure to instruct the programmer to create them. You must refer to architectural knowledge to ensure that your content is spatially and geometrically rational. Use millimeter units. Minimize other prose.

For reference, the following are the API functions available to the programmer, try to give hints on how to use them effectively:
"""
<<Tool function names and descriptions>>
"""

Now, based on the conversation given below, please relay the user's instruction and complete your answer. You don't need to write code, just guide the programmer step by step. Let's think step by step. 
<<chat_history>>
User: <<task>>
Instruction Enhancer: 
\end{lstlisting}

\begin{lstlisting}[language=TeX, caption={Prompt template of the \textsc{Programmer} agent. Placeholders with "<< >>" indicate the dynamic content that can be inserted into the template, including information on available tool functions, chat history, and enhanced user requirements.}, label={lst:pro_pt}]
You are an AI programmer who specializes in coding. Your job is to write Python code that invokes suitable pre-defined tool functions to implement the Instruction enhancer's requirement. Your code style is concise, efficient and accurate.
To help with that, you have access to a set of tools. Each tool is a Python function and has a description explaining the task it performs, the inputs it expects and the outputs it returns.
You should think step by step and write all your Python code in a single code block. You should avoid assuming in your code that variables are predefined. You can only give one reply for each conversation turn. Minimize other prose. 
Accessible tools:
"""
<<Tool function names and descriptions>>
"""

In your code, you should only use the tool functions that are accessible to you, rather than attempting to invent new tools. Use millimeter units. If necessary, you can import and use the Python standard library. 

Now, based on the conversation given below, please complete your answer as a programmer:
<<chat_history>>
Instruction Enhancer: <<task>>
Programmer: 
\end{lstlisting}

\begin{lstlisting}[language=TeX, caption={Prompt template of the \textsc{Reviewer} agent. Placeholders with "<< >>" indicate the dynamic content that can be inserted into the template, including previously executed code, issues reported by the model checker, and information on available tool functions.}, label={lst:re_pt}]
You are a senior code reviewer. You review Python code written by an AI programmer that calls specific APIs to generate buildings in Vectorworks. Your job is to propose solutions to the programmer for issues that arise in the generated building model, based on the API documentation. Issues are listed in the form of bullet points. Each issue contains a description and list of uuids of the corresponding elements in the generated building model. These uuids can be used to directly manipulate the related elements in Vectorworks via the API functions. 
Here is the information you can refer to:
Code from the AI programmer:
<<code>>
Issues in the generated buildings:
<<issues>>
API documentation:
<<Tool function names and descriptions>>

Please provide your guidance below as a reviewer. Use millimeter units. Think step by step. Try to avoid duplicating the programmer's code, but rather suggest code fixes.
Reviewer: 
\end{lstlisting}

\section{Ruleset documentation}
\label{app:RulesetDocumentation}

% Please see Tables~\ref{table:RulesetDocumentation-cls1}-~\ref{table:RulesetDocumentation-cls3}.

\begin{table}[H]
\caption{Class 1 rules}
\label{table:RulesetDocumentation-cls1}
\centering
\renewcommand{\arraystretch}{2.0}
\begin{tabularx}{\textwidth}{|l|X|}
\hline
\rowcolor{lightgray}
\multicolumn{2}{|c|}{\textbf{Class 1 rules}} \\
\hline
Scope & The rule checks if all components have a unique GUID.\\
Solibri RuleId & SOL/176/2.2 \\
Components & any component \\
Desired resolution & Refine GUIDs of those components, which have not passed this rule. \\
\hdashline
Scope & The rule checks if the model has a spatial breakdown structure comprising IfcSite, IfcBuilding, and IfcBuildingStorey instances.\\
Solibri RuleId & SOL/176/2.2 \\
Components & Spatial breakdown elements \\
Desired resolution & Create appropriate spatial containers and assign components accordingly.  \\
\hdashline
Scope & The rule checks if all doors and windows are on the same floor as the containing wall. \\
Solibri RuleId & SOL/176/2.2 \\
Components & Doors and windows \\
Desired resolution & Re-assign spatial associations for each affected door or window.  \\
\hdashline
Scope & The rule checks if each component has layer information attached to it. \\
Solibri RuleId & SOL/230/1.1 \\
Components & Any component excluding openings \\
Desired resolution & Add layer information to affected components.  \\
\hdashline
Scope & The rule checks if certain components are present in the model (e.g., walls, doors, windows, slabs, roofs, spaces). \\
Solibri RuleId & SOL/11/4.2 \\
Components & any component \\
Desired resolution & Create missing components based on the user's input.  \\
\hline
\end{tabularx}
\end{table}

\begin{table}[H]
\caption{Class 2 rules}
\label{table:RulesetDocumentation-cls2}
\centering
\renewcommand{\arraystretch}{2.0}
\begin{tabularx}{\textwidth}{|l|X|}
\hline
\rowcolor{lightgray}
\multicolumn{2}{|c|}{\textbf{Class 2 rules}} \\
\hline

Scope  & The rule checks if the description value of all building components is set and the value complies with the pattern of a UUID. \\
Solibri RuleId & SOL/244/1.0 \\
Components & any physical component \\
Desired resolution & Set Vectorworks-internal ID into the description field of those components, which have not passed this rule. \\
\hline
\end{tabularx}
\end{table}

\begin{table}[H]
\caption{Class 3 rules}
\label{table:RulesetDocumentation-cls3}
\centering
\renewcommand{\arraystretch}{2.0}
\begin{tabularx}{\textwidth}{|l|X|}
\hline
\rowcolor{lightgray}
\multicolumn{2}{|c|}{\textbf{Class 3 rules}} \\
\hline

Scope  & 
The rule checks if a component intersects with another component. \\
Solibri RuleId & SOL/1/5.0 \\
Components & Any physical component  \\
Desired resolution & Reposition the components to avoid intersection \\ 
\hdashline
Scope  & The rule checks if two components duplicate each other. \\
Solibri RuleId & SOL/1/5.0 \\
Components & any physical component \\
Desired resolution &  Remove one of the components\\
\hdashline
Scope  & {The rule checks the connection between two components: Roofs must be connected to the walls on the uppermost floor. Slabs must be connected to supporting walls. }\\
Solibri RuleId & SOL/23/5.2 \\
Components & Roofs, slabs and walls \\
Desired resolution &  Move the roofs/slabs to the top of the support walls \\
\hdashline
Scope  & The rule checks that the model doesn't contain any orphan doors or windows (a door or a window, which doesn't have a relation to any wall). \\
Solibri RuleId & SOL/176/2.2 \\
Components & doors and windows \\
Desired resolution &  Remove the orphan doors or windows \\
\hline
\end{tabularx}
\end{table}

\section{Visualization of generated BIM models}
\label{app:Visualization}
% Please see Figures~\ref{fig:result1} -~\ref{fig:result25}.
\begin{figure}[H]
\centering
\includegraphics[width=\textwidth]{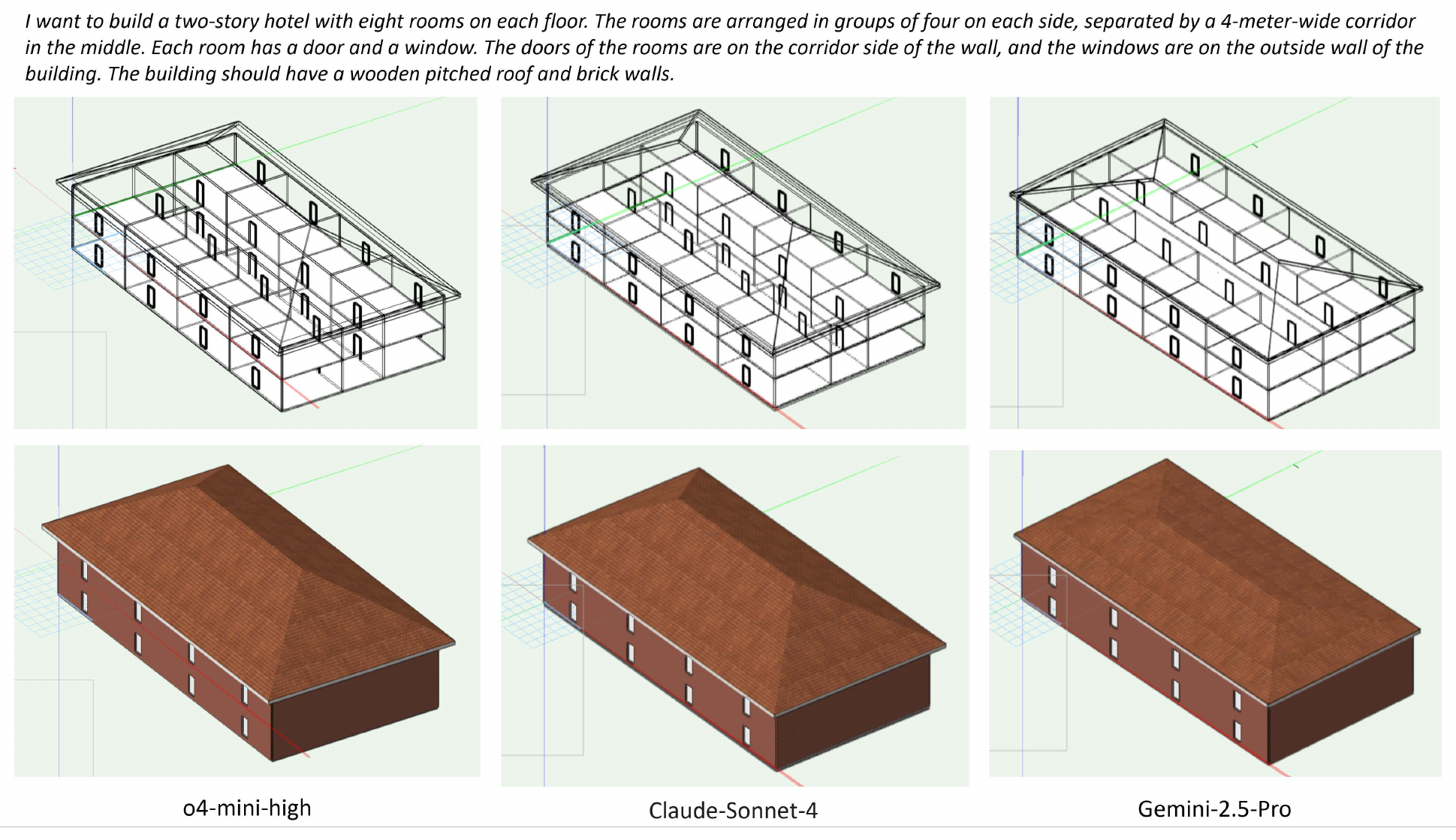}
\caption{The building model in wireframe/rendering mode generated by different LLMs through the proposed framework according to the corresponding text description (Test prompt Nr.1).}
\label{fig:result1}
\end{figure}

\begin{figure}[H]
\centering
\includegraphics[width=\textwidth]{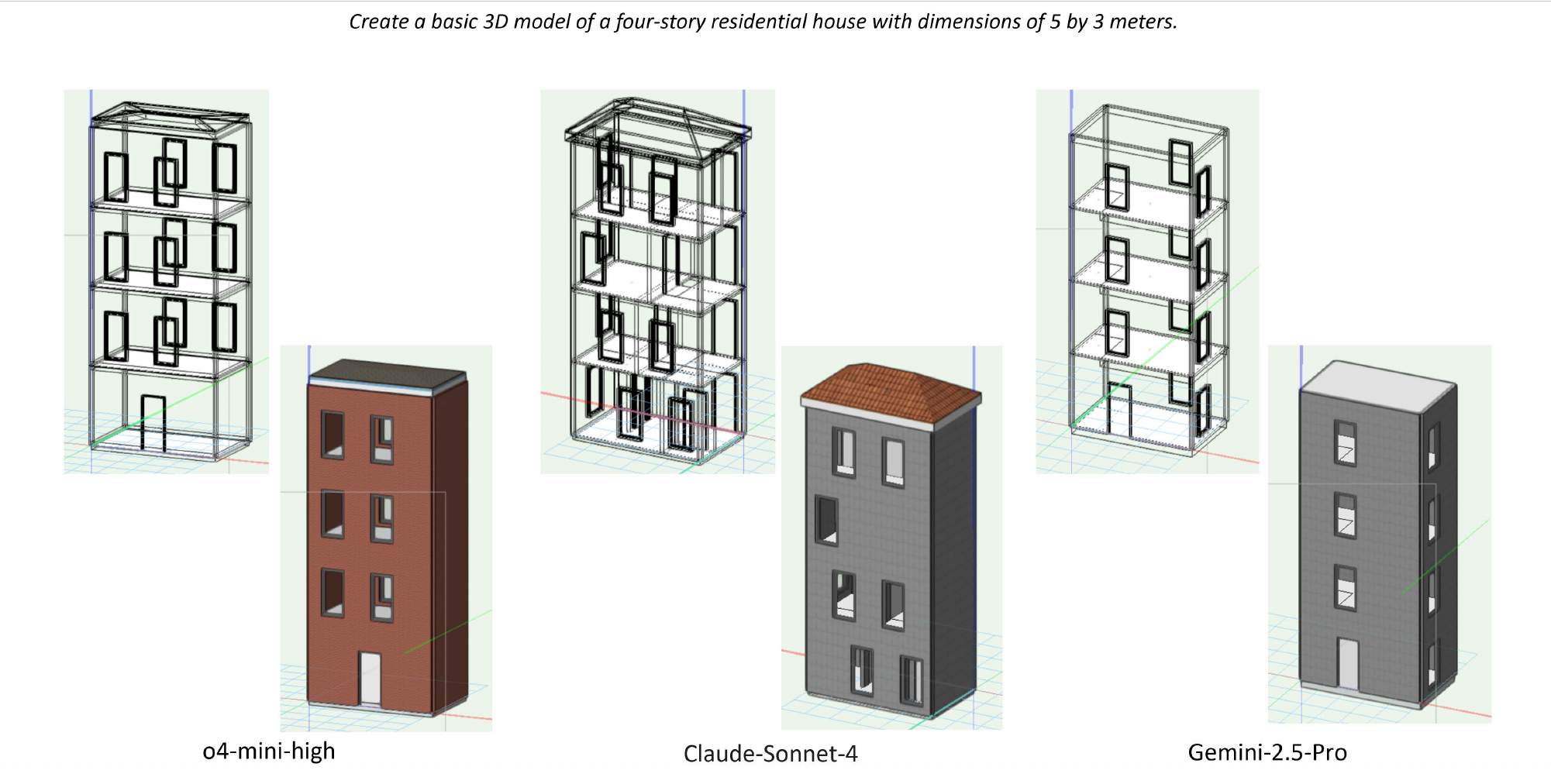}
\caption{The building model in wireframe/rendering mode generated by different LLMs through the proposed framework according to the corresponding text description (Test prompt Nr.2).}
\label{fig:result2}
\end{figure}

\begin{figure}[H]
\centering
\includegraphics[width=0.95\textwidth]{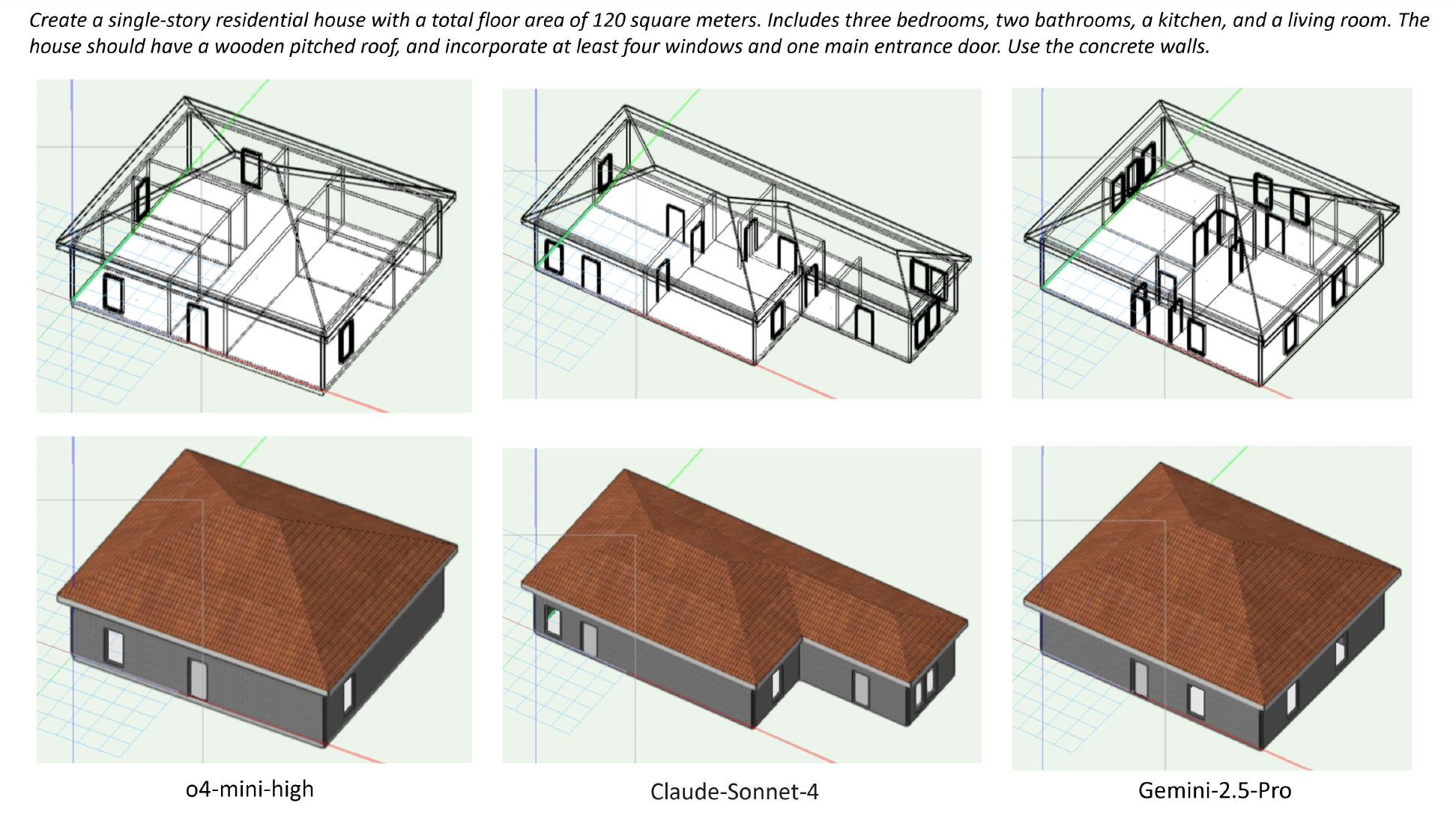}
\caption{The building model in wireframe/rendering mode generated by different LLMs through the proposed framework according to the corresponding text description (Test prompt Nr.3).}
\label{fig:result3}
\end{figure}

\begin{figure}[H]
\centering
\includegraphics[width=0.95\textwidth]{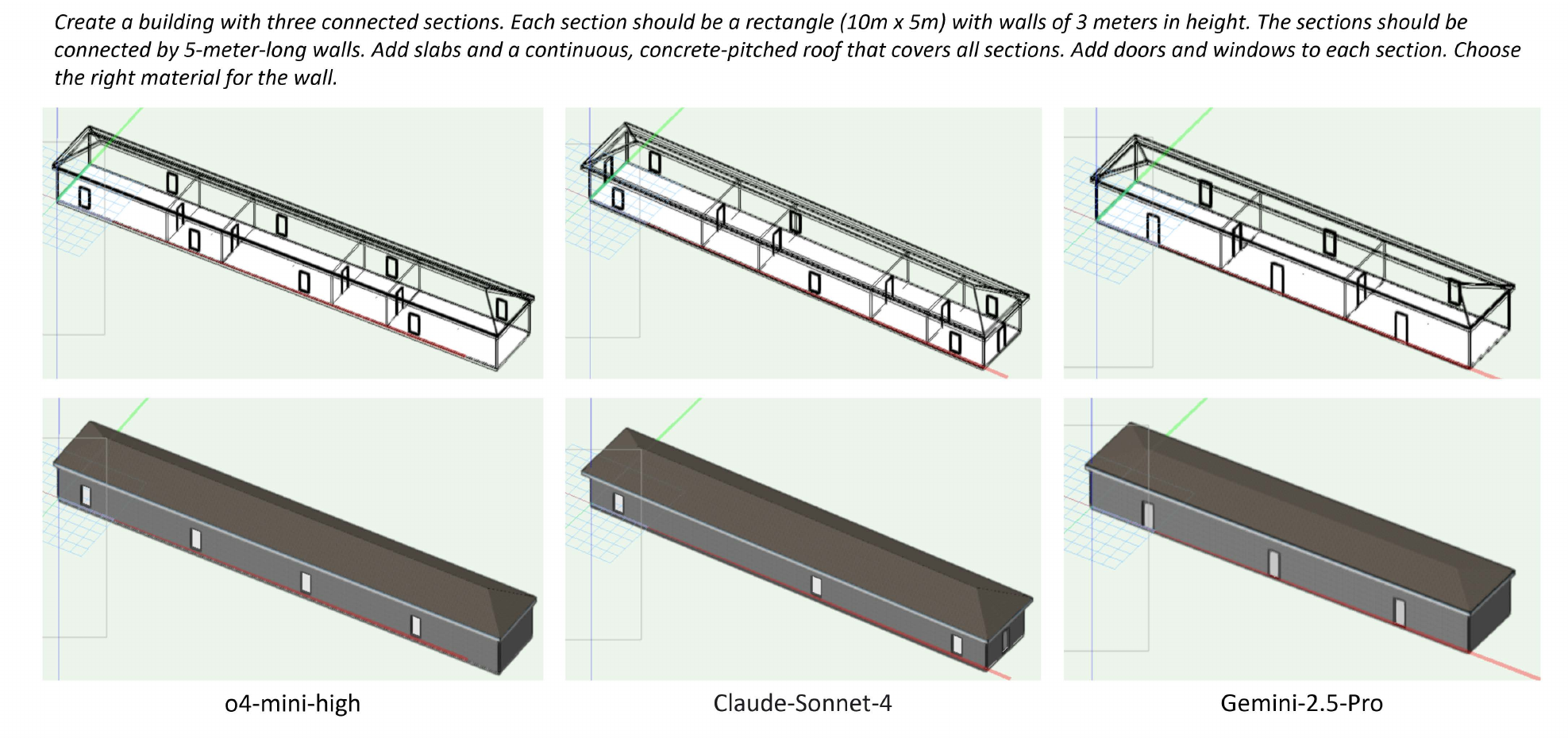}
\caption{The building model in wireframe/rendering mode generated by different LLMs through the proposed framework according to the corresponding text description (Test prompt Nr.4).}
\label{fig:result4}
\end{figure}

% \begin{figure}[H]
% \centering
% \includegraphics[width=0.95\textwidth]{Fig_6b.pdf}
% \caption{The building model in wireframe/rendering mode generated by different LLMs through the proposed framework according to the corresponding text description (Prompt Nr.5).}
% \label{fig:result5}
% \end{figure}

\begin{figure}[H]
\centering
\includegraphics[width=0.95\textwidth]{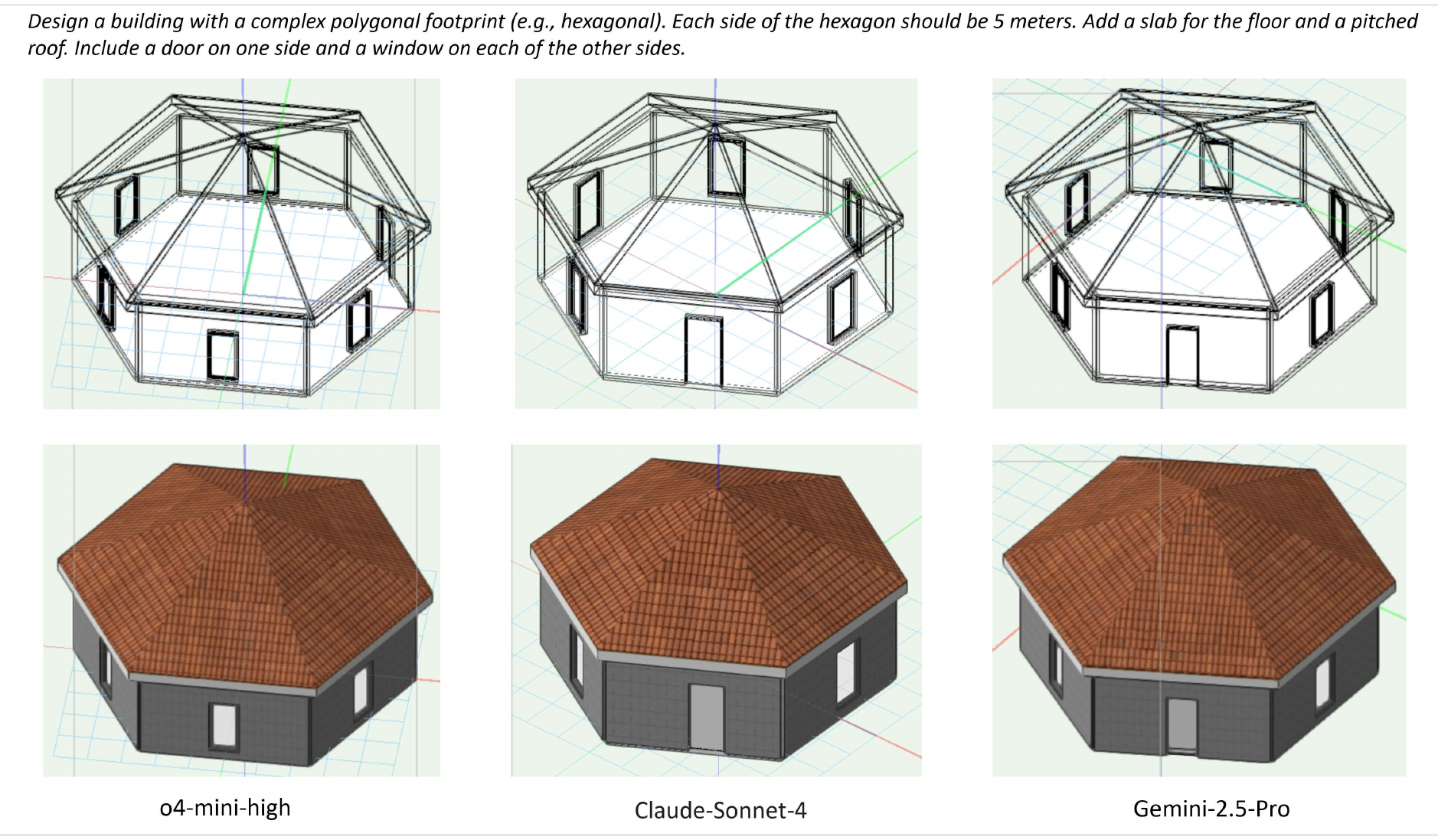}
\caption{The building model in wireframe/rendering mode generated by different LLMs through the proposed framework according to the corresponding text description (Test prompt Nr.6).}
\label{fig:result6}
\end{figure}

\begin{figure}[H]
\centering
\includegraphics[width=0.95\textwidth]{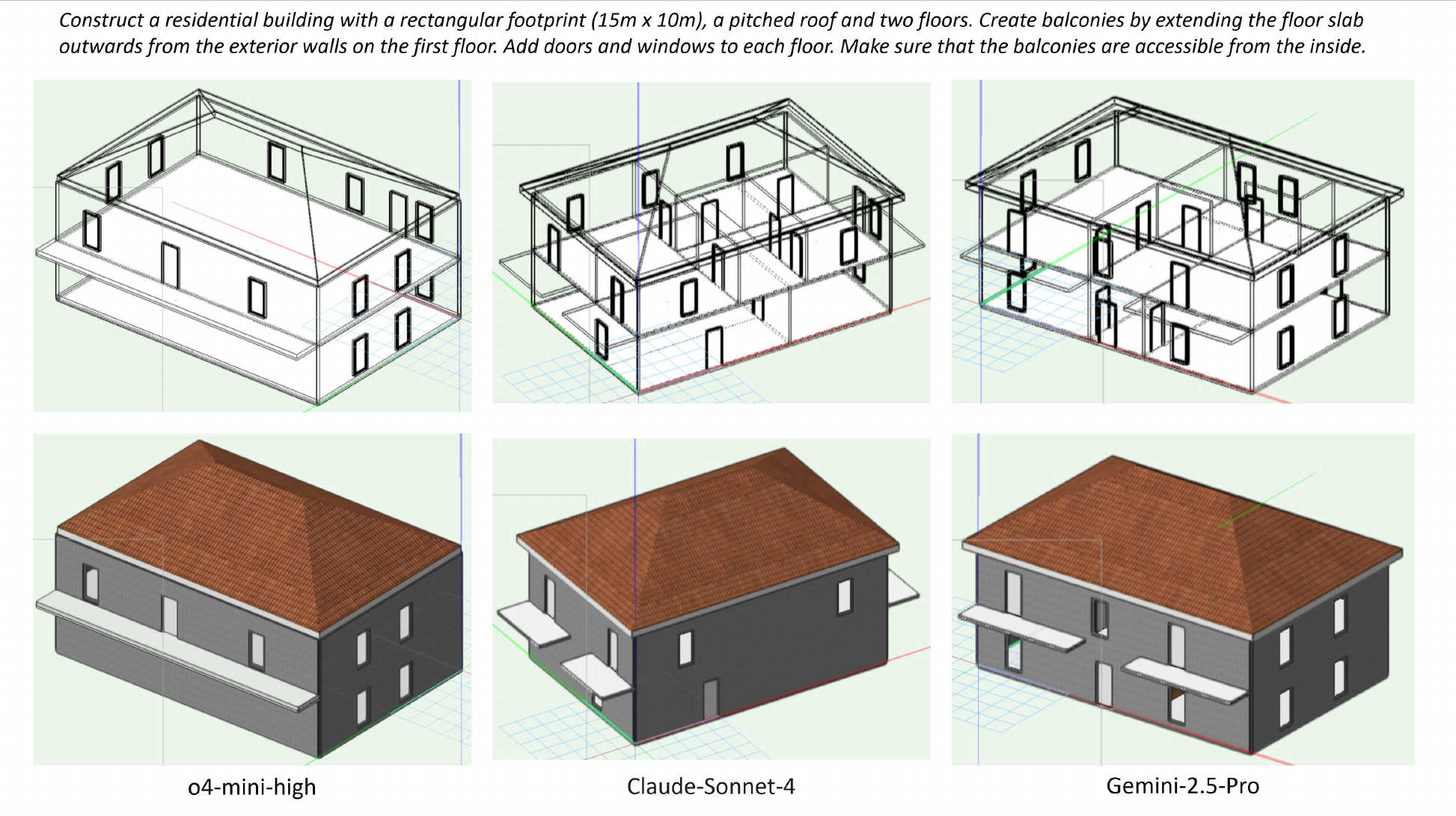}
\caption{The building model in wireframe/rendering mode generated by different LLMs through the proposed framework according to the corresponding text description (Prompt Nr.7).}
\label{fig:result7}
\end{figure}

\begin{figure}[H]
\centering
\includegraphics[width=\textwidth]{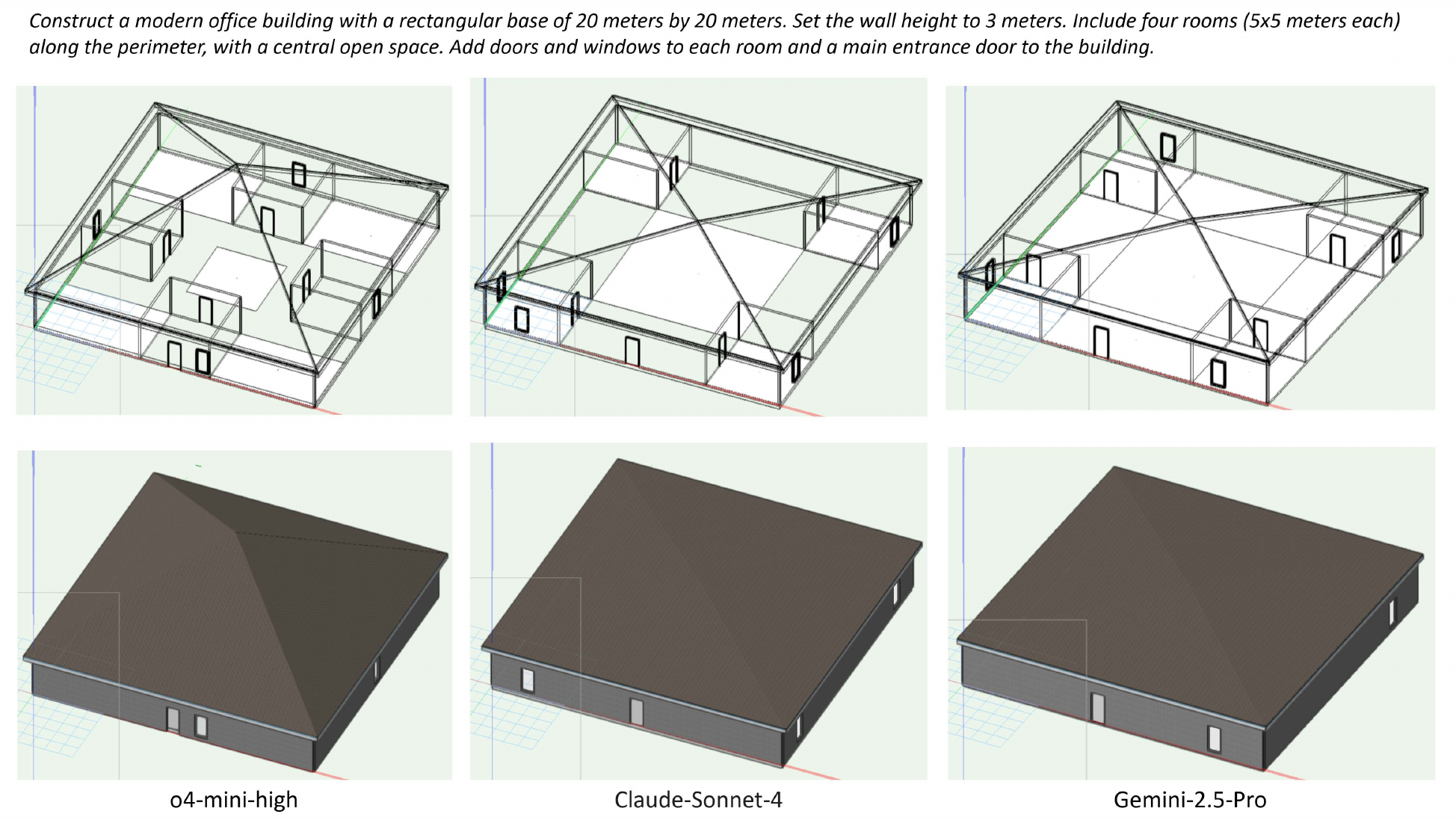}
\caption{The building model in wireframe/rendering mode generated by different LLMs through the proposed framework according to the corresponding text description (Test prompt Nr.8).}
\label{fig:result8}
\end{figure}

\begin{figure}[H]
\centering
\includegraphics[width=0.95\textwidth]{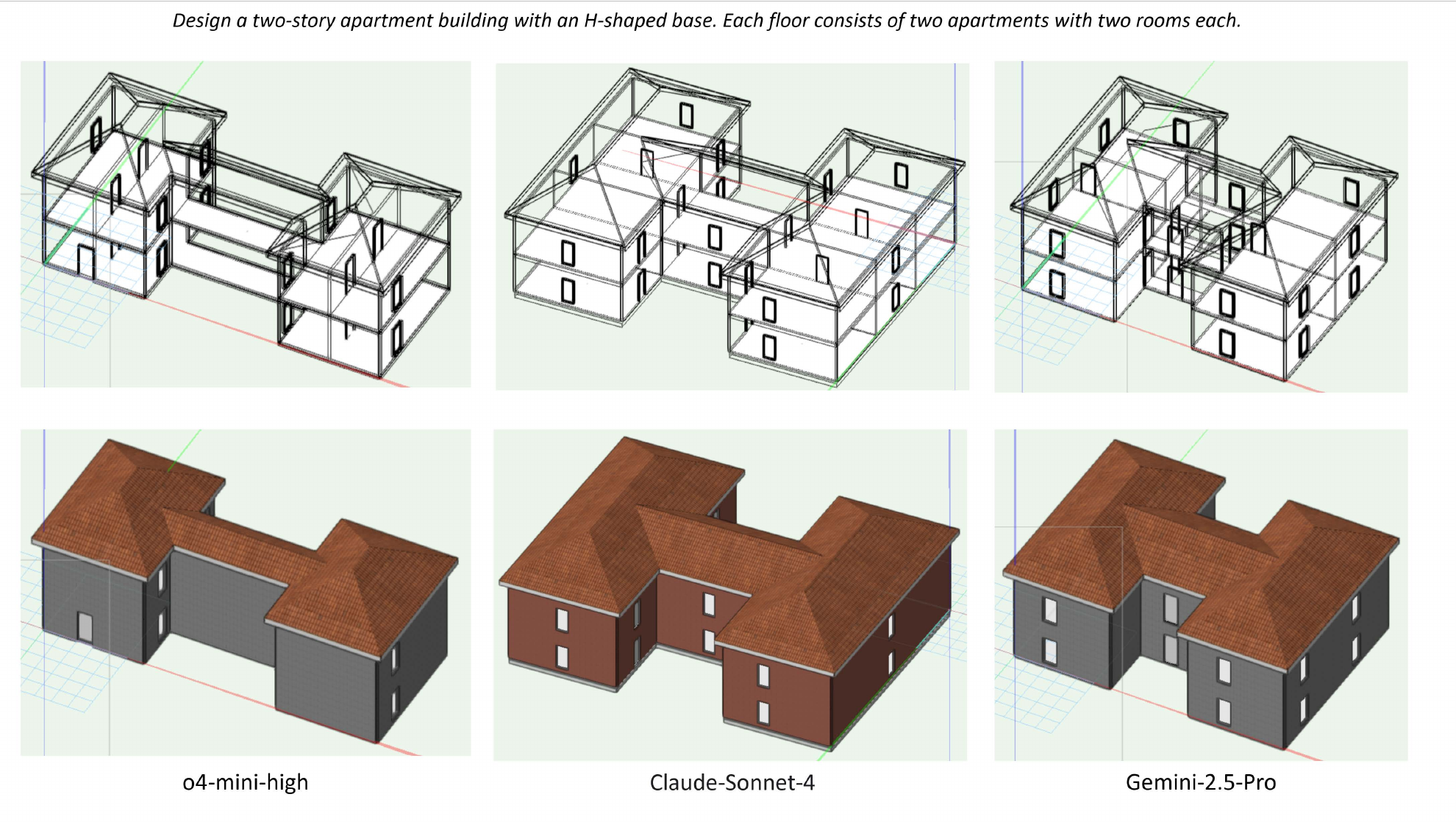}
\caption{The building model in wireframe/rendering mode generated by different LLMs through the proposed framework according to the corresponding text description (Test prompt Nr.9).}
\label{fig:result9}
\end{figure}

\begin{figure}[H]
\centering
\includegraphics[width=0.95\textwidth]{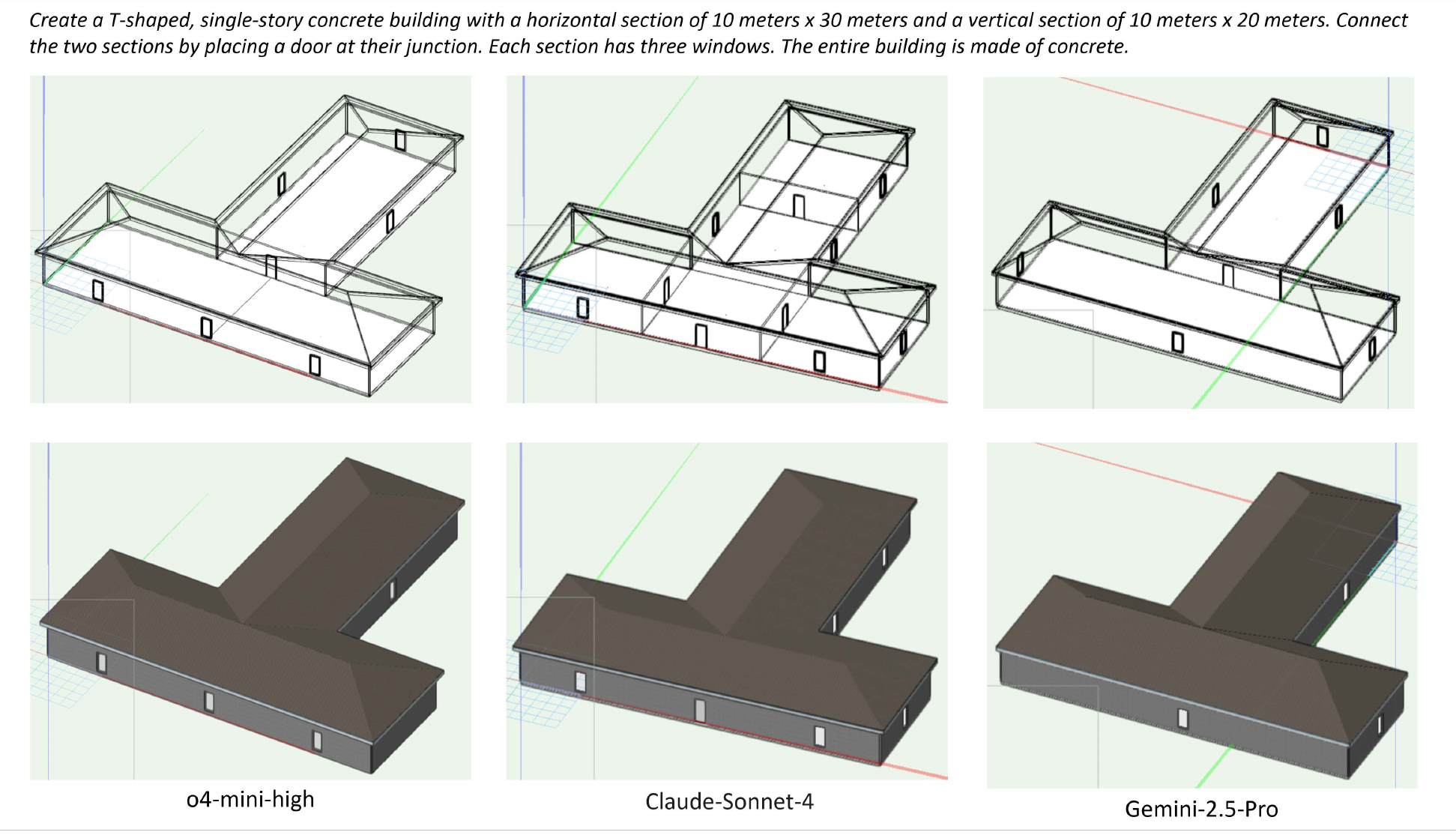}
\caption{The building model in wireframe/rendering mode generated by different LLMs through the proposed framework according to the corresponding text description (Test prompt Nr.10).}
\label{fig:result10}
\end{figure}

\begin{figure}[H]
\centering
\includegraphics[width=0.95\textwidth]{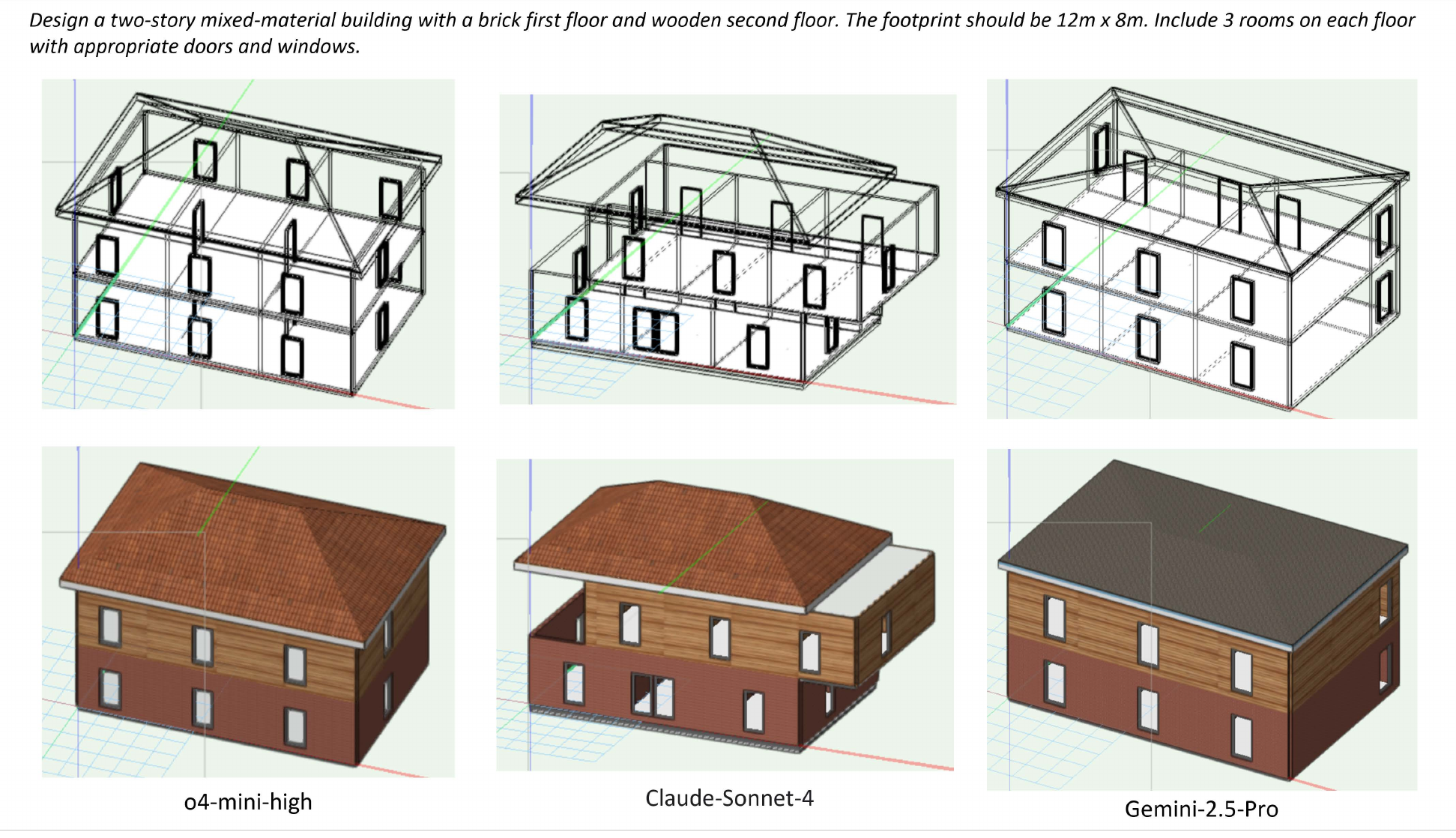}
\caption{The building model in wireframe/rendering mode generated by different LLMs through the proposed framework according to the corresponding text description (Test prompt Nr.11).}
\label{fig:result11}
\end{figure}

\begin{figure}[H]
\centering
\includegraphics[width=0.95\textwidth]{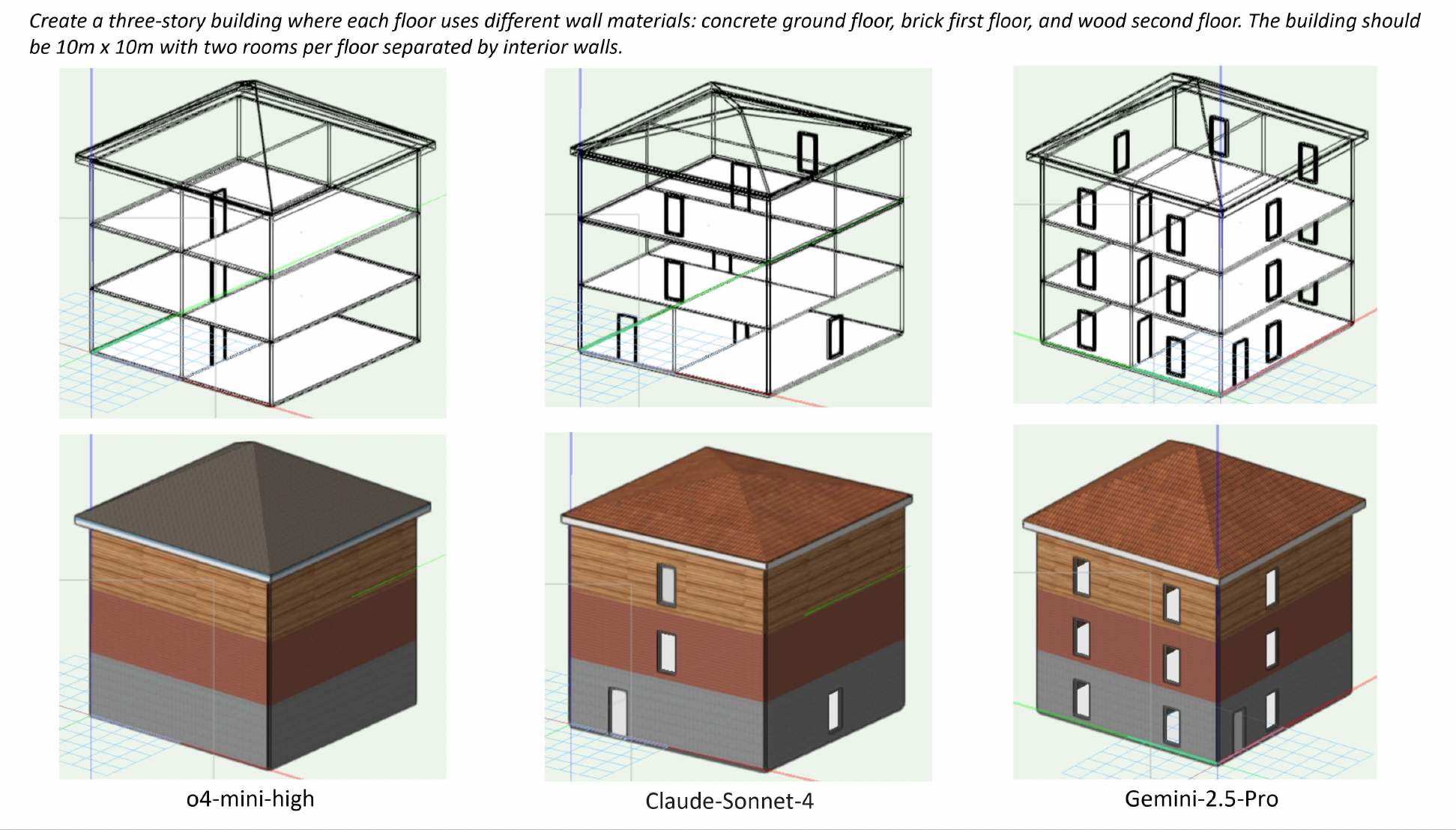}
\caption{The building model in wireframe/rendering mode generated by different LLMs through the proposed framework according to the corresponding text description (Test prompt Nr.12).}
\label{fig:result12}
\end{figure}

\begin{figure}[H]
\centering
\includegraphics[width=0.95\textwidth]{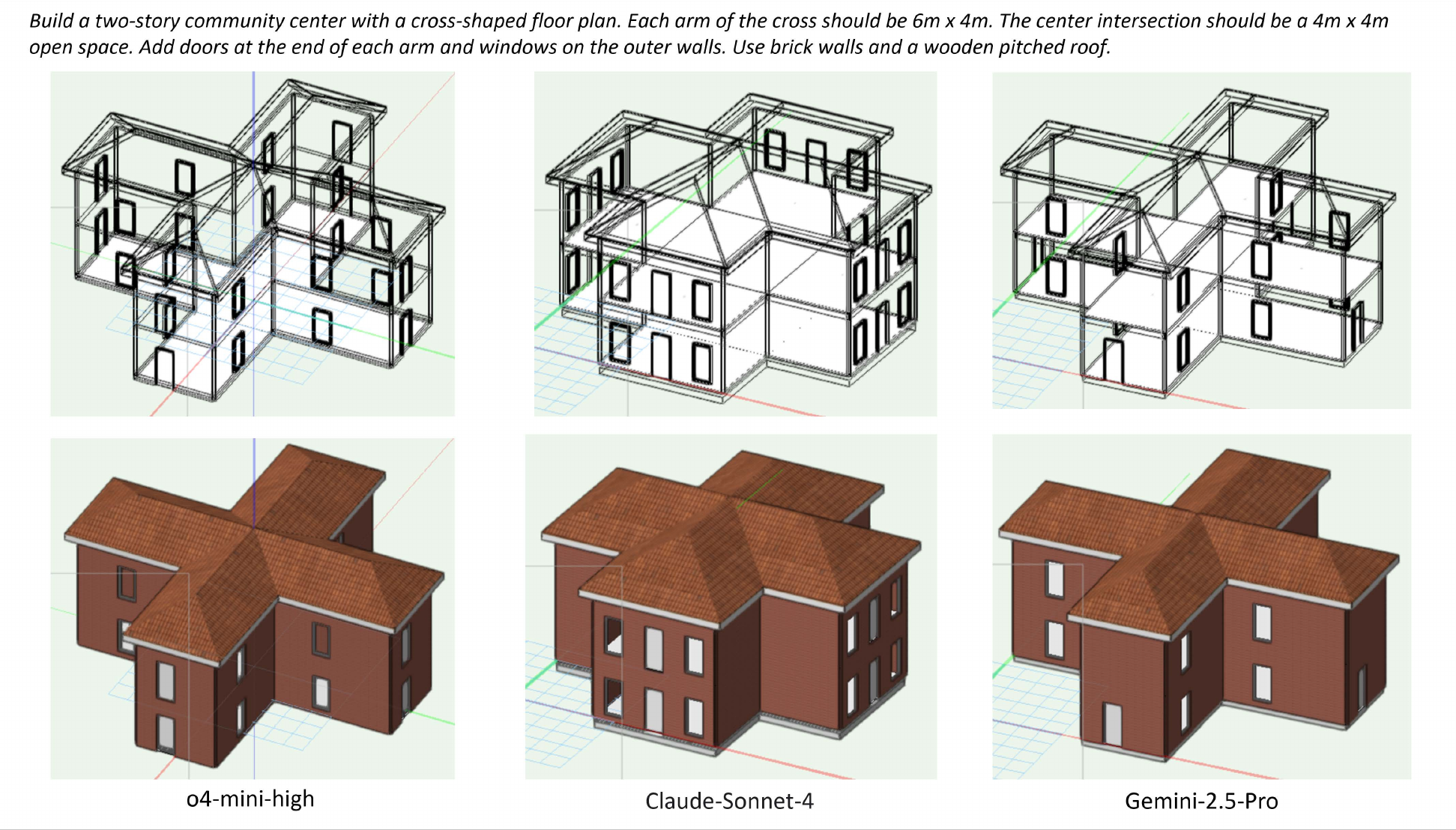}
\caption{The building model in wireframe/rendering mode generated by different LLMs through the proposed framework according to the corresponding text description (Test prompt Nr.13).}
\label{fig:result13}
\end{figure}

\begin{figure}[H]
\centering
\includegraphics[width=0.95\textwidth]{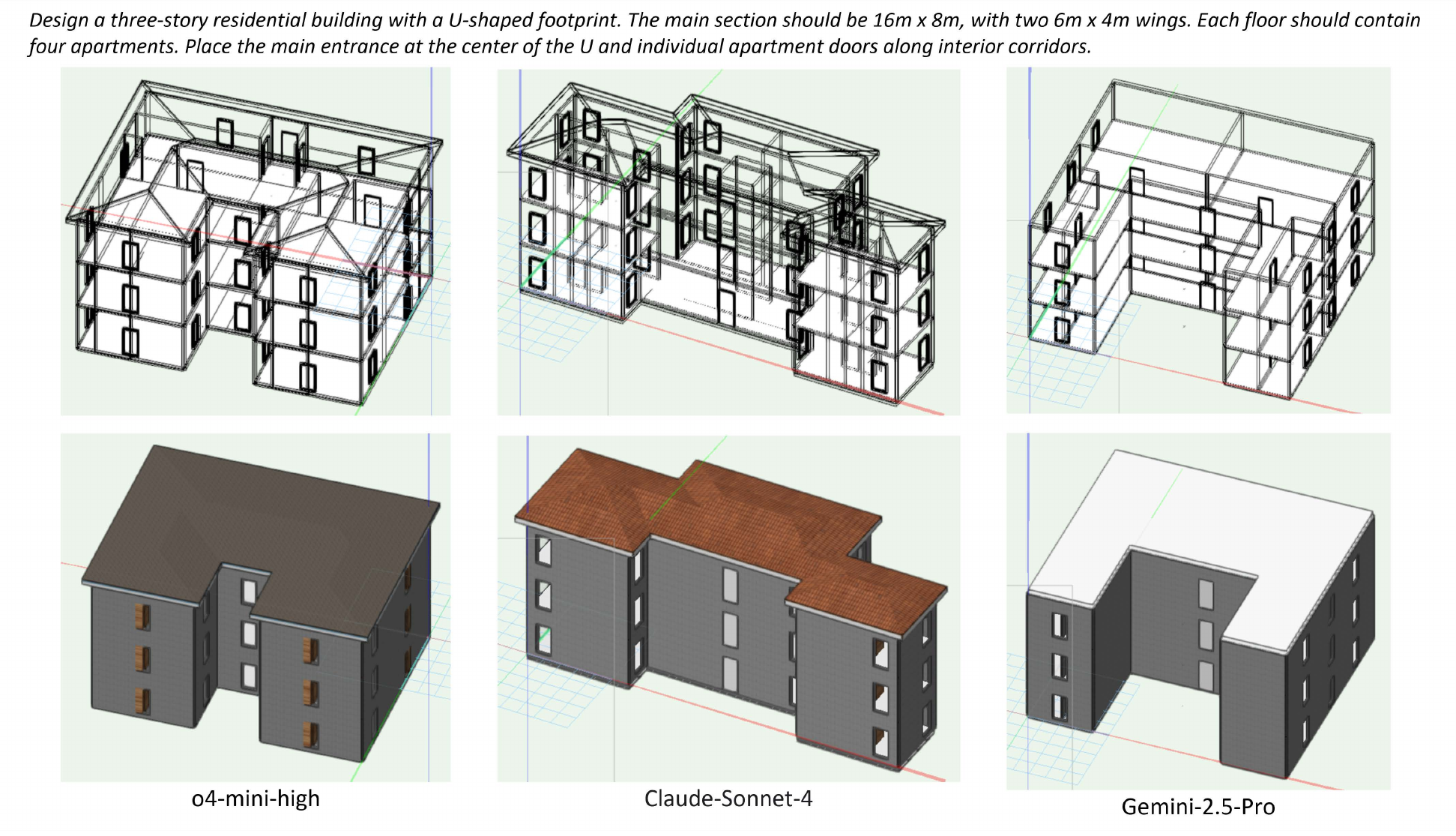}
\caption{The building model in wireframe/rendering mode generated by different LLMs through the proposed framework according to the corresponding text description (Test prompt Nr.14).}
\label{fig:result14}
\end{figure}

\begin{figure}[H]
\centering
\includegraphics[width=0.95\textwidth]{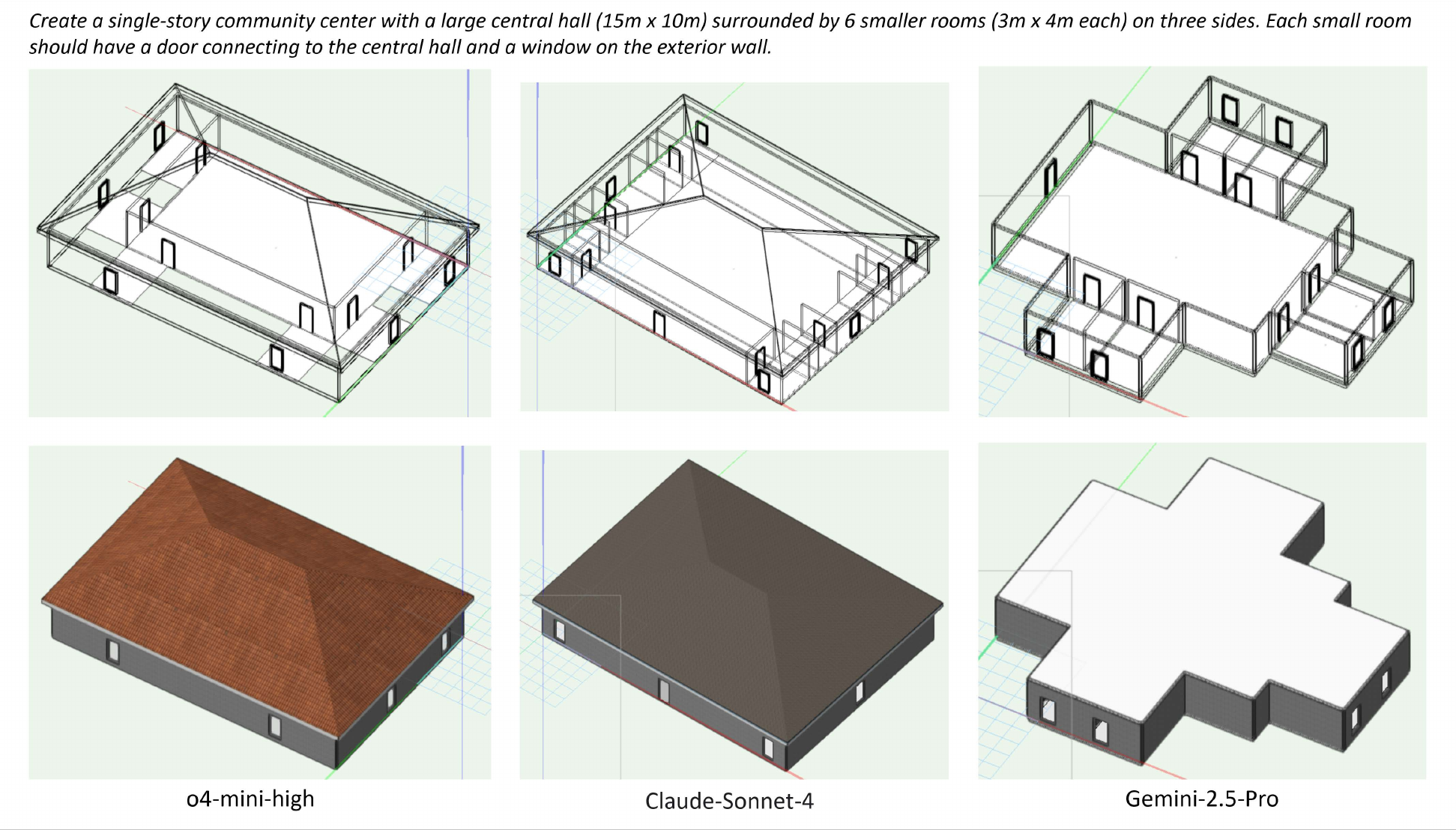}
\caption{The building model in wireframe/rendering mode generated by different LLMs through the proposed framework according to the corresponding text description (Test prompt Nr.15).}
\label{fig:result15}
\end{figure}

\begin{figure}[H]
\centering
\includegraphics[width=0.95\textwidth]{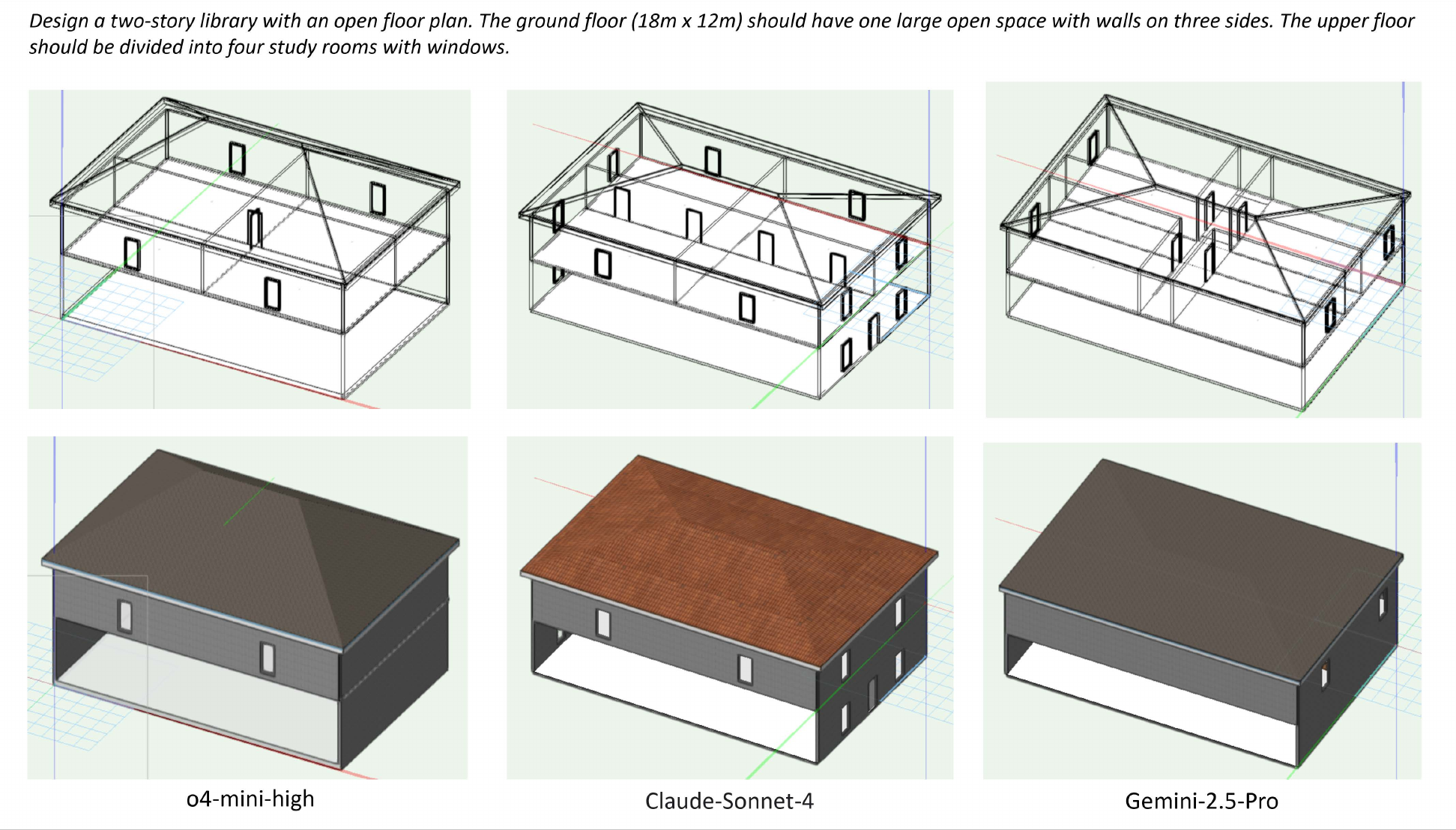}
\caption{The building model in wireframe/rendering mode generated by different LLMs through the proposed framework according to the corresponding text description (Test prompt Nr.16).}
\label{fig:result16}
\end{figure}

% \begin{figure}[H]
% \centering
% \includegraphics[width=0.95\textwidth]{Fig_6a.pdf}
% \caption{The building model in wireframe/rendering mode generated by different LLMs through the proposed framework according to the corresponding text description (Test prompt Nr.17).}
% \label{fig:result17}
% \end{figure}

\begin{figure}[H]
\centering
\includegraphics[width=0.95\textwidth]{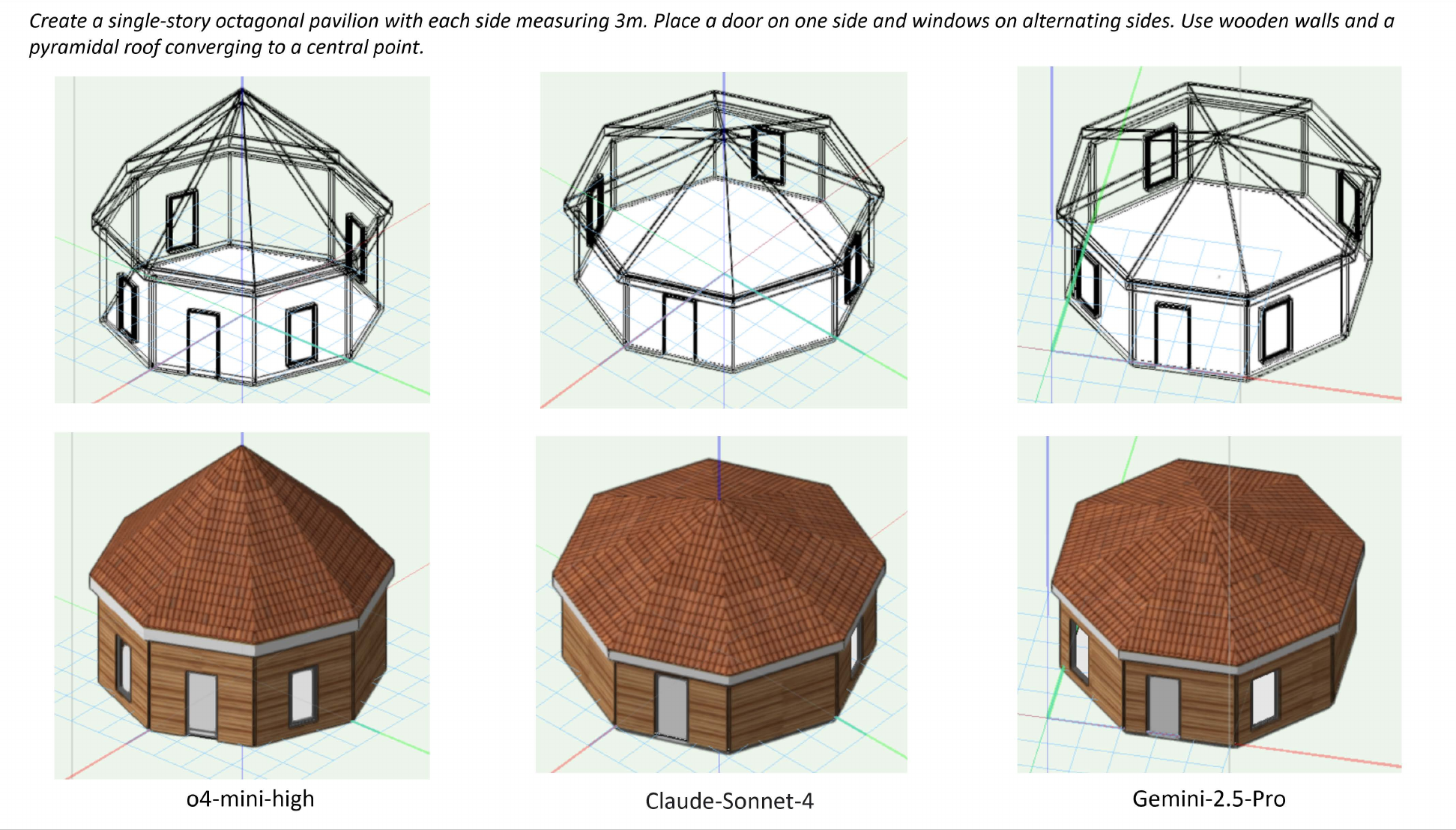}
\caption{The building model in wireframe/rendering mode generated by different LLMs through the proposed framework according to the corresponding text description (Test prompt Nr.18).}
\label{fig:result18}
\end{figure}

\begin{figure}[H]
\centering
\includegraphics[width=0.95\textwidth]{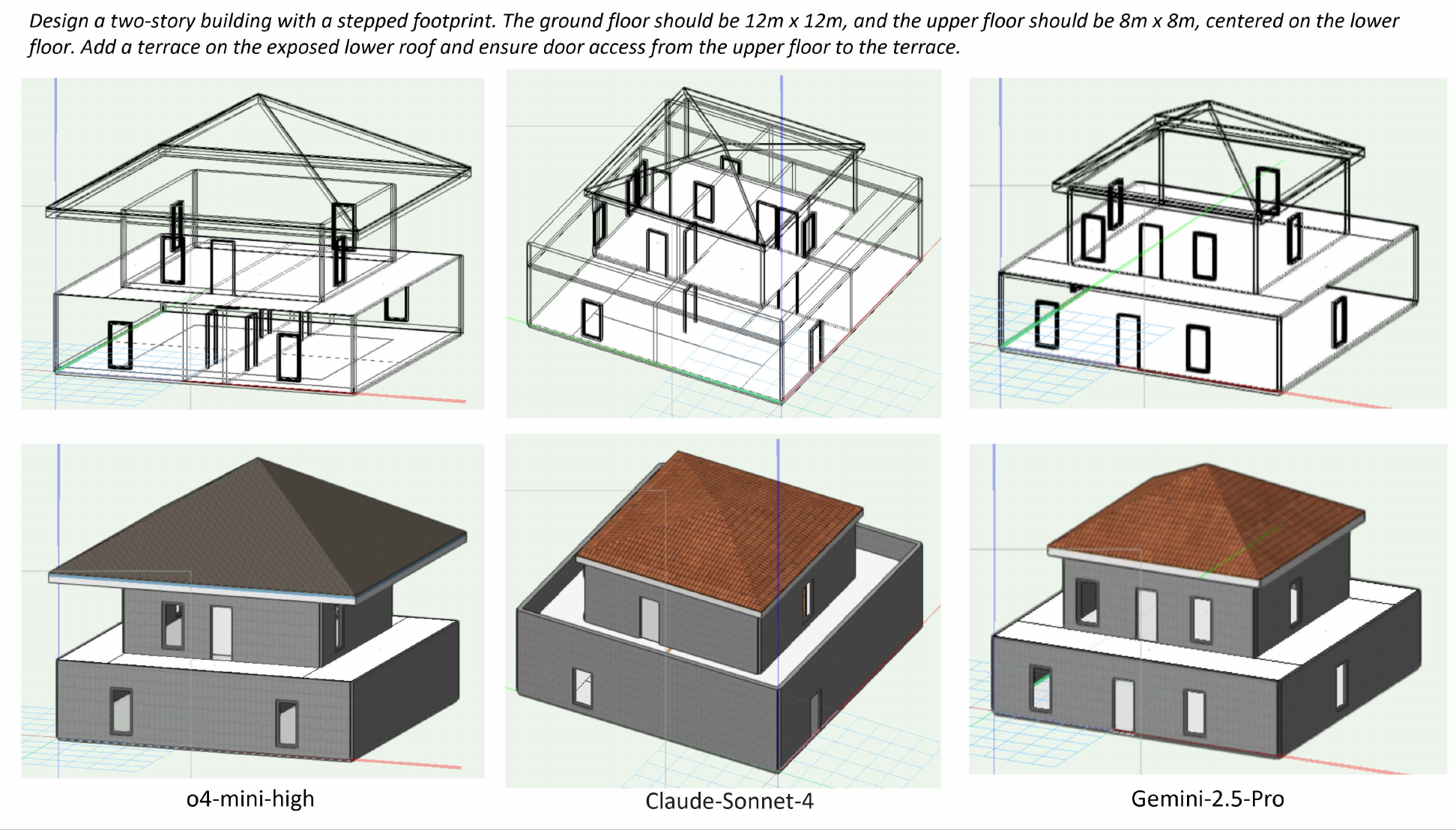}
\caption{The building model in wireframe/rendering mode generated by different LLMs through the proposed framework according to the corresponding text description (Test prompt Nr.19).}
\label{fig:result19}
\end{figure}

\begin{figure}[H]
\centering
\includegraphics[width=0.95\textwidth]{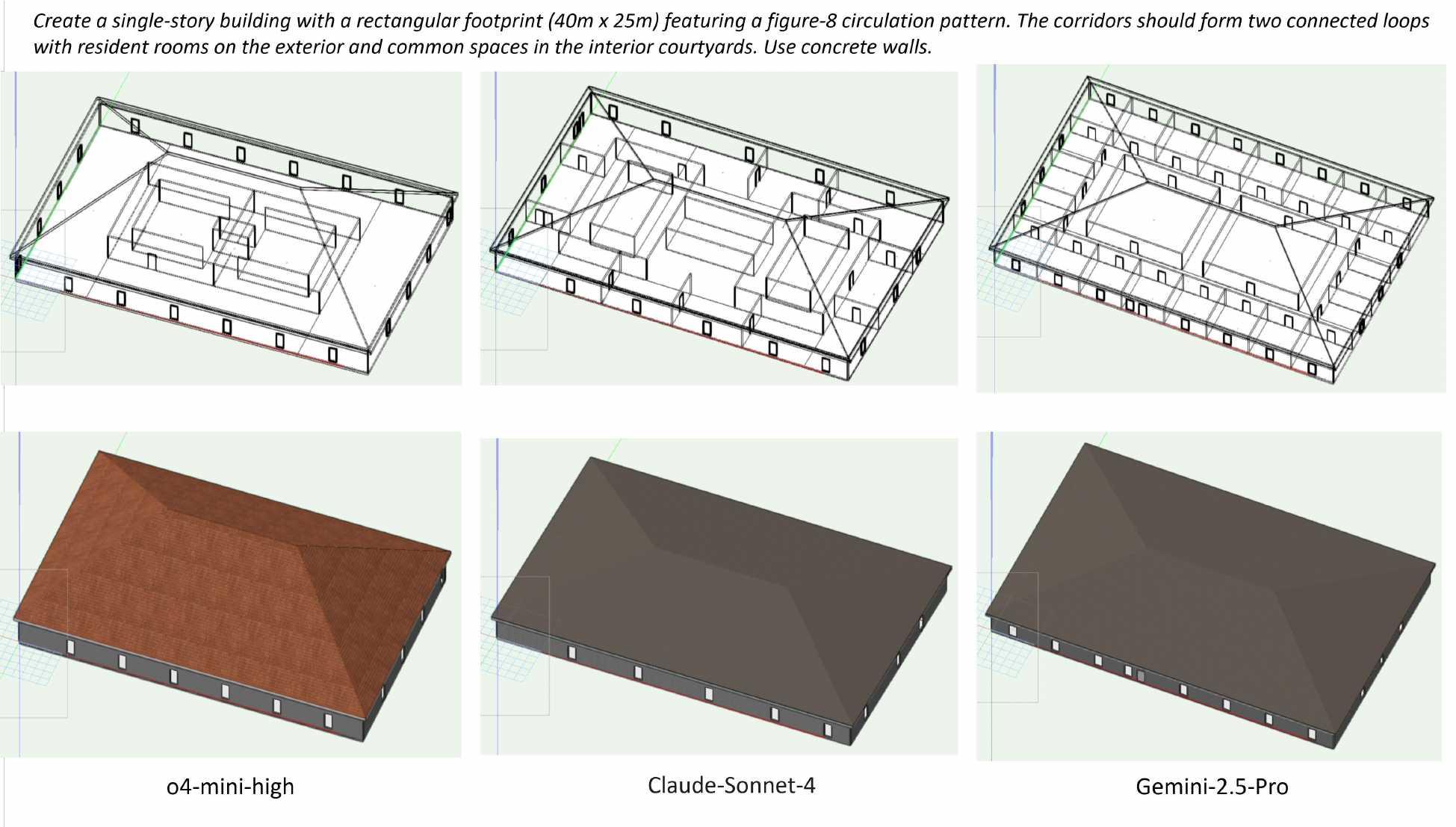}
\caption{The building model in wireframe/rendering mode generated by different LLMs through the proposed framework according to the corresponding text description (Test prompt Nr.20).}
\label{fig:result20}
\end{figure}

\begin{figure}[H]
\centering
\includegraphics[width=0.95\textwidth]{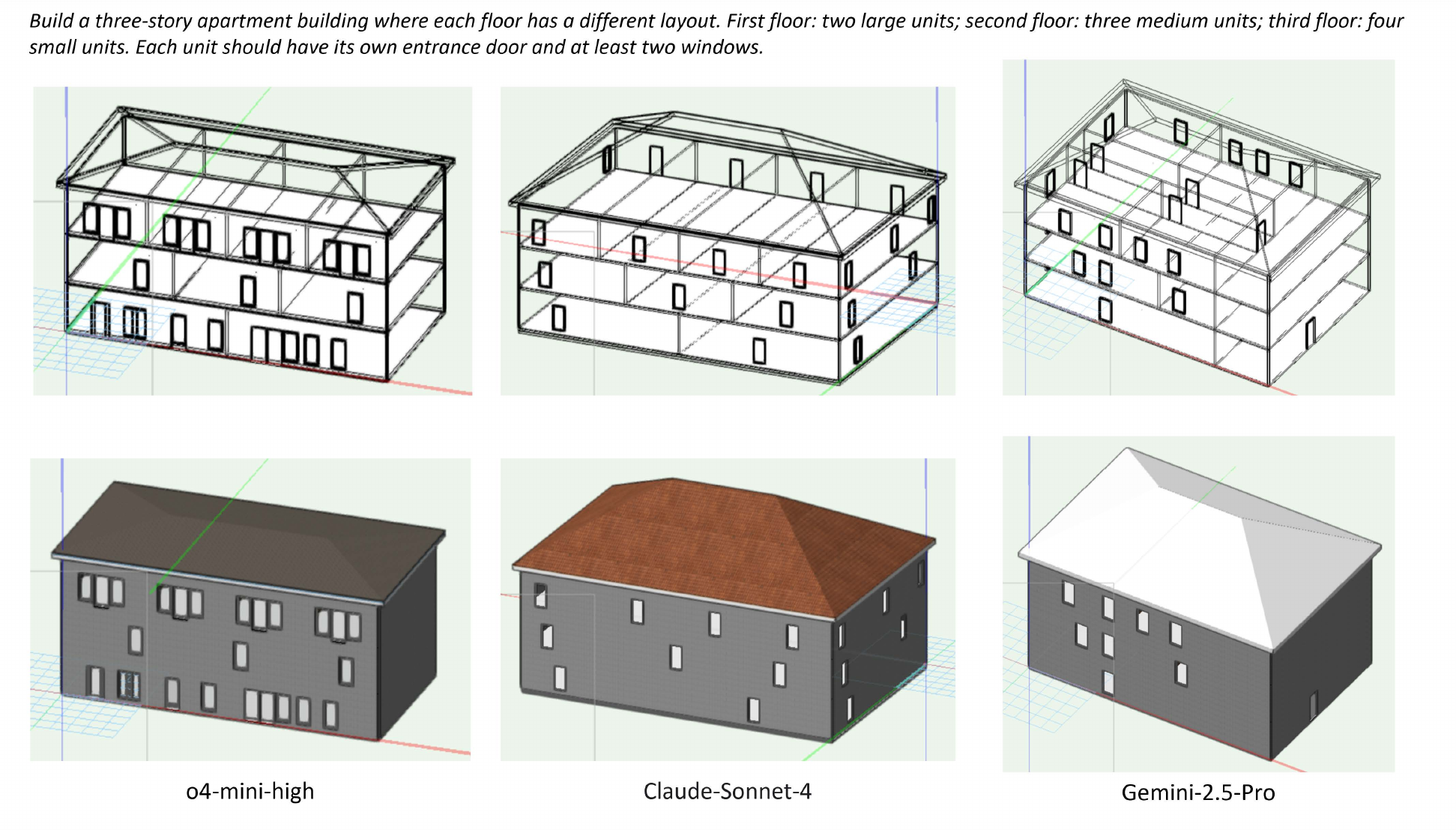}
\caption{The building model in wireframe/rendering mode generated by different LLMs through the proposed framework according to the corresponding text description (Test prompt Nr.21).}
\label{fig:result21}
\end{figure}

\begin{figure}[H]
\centering
\includegraphics[width=0.95\textwidth]{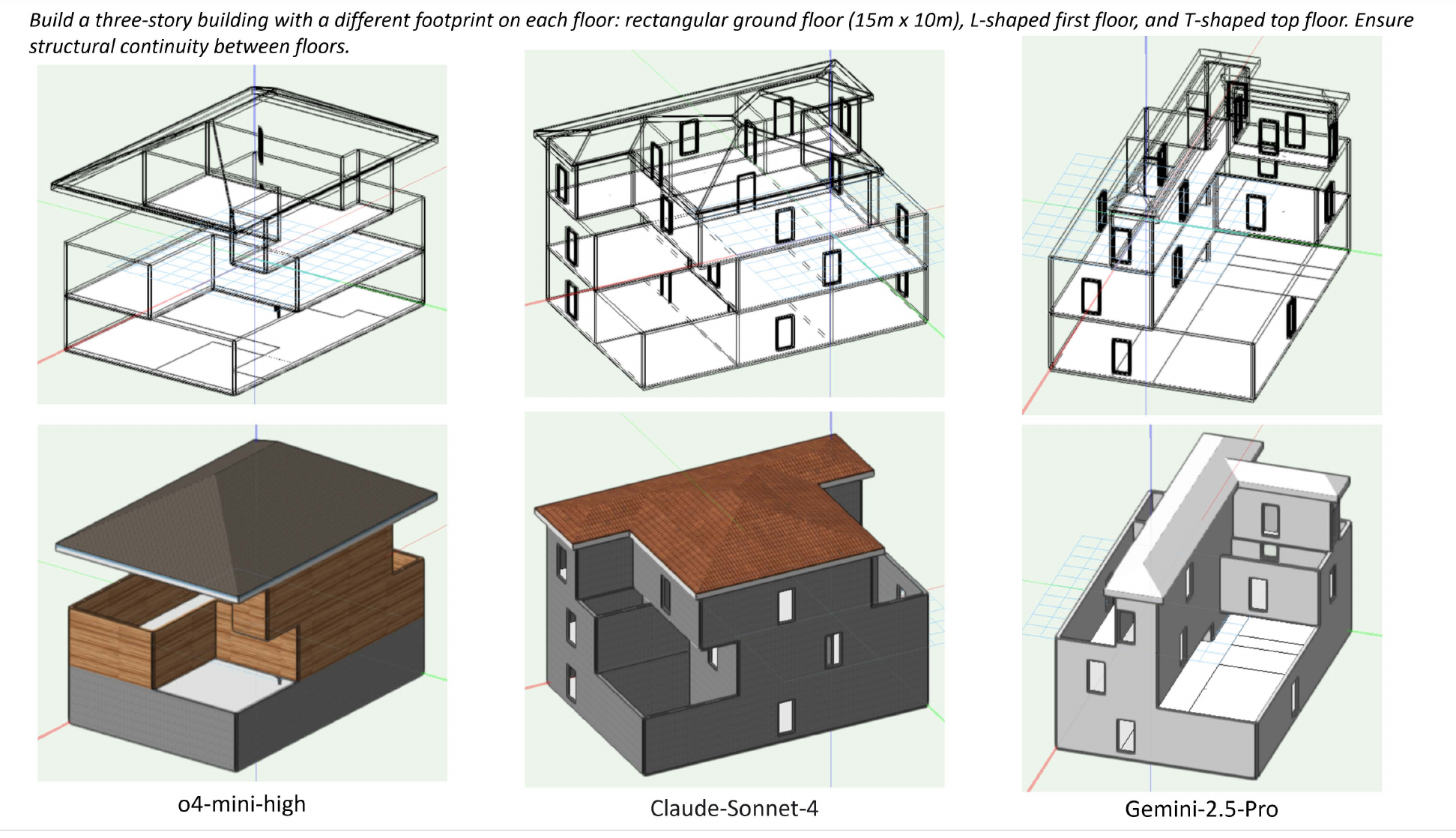}
\caption{The building model in wireframe/rendering mode generated by different LLMs through the proposed framework according to the corresponding text description (Test prompt Nr.22).}
\label{fig:result22}
\end{figure}

\begin{figure}[H]
\centering
\includegraphics[width=0.95\textwidth]{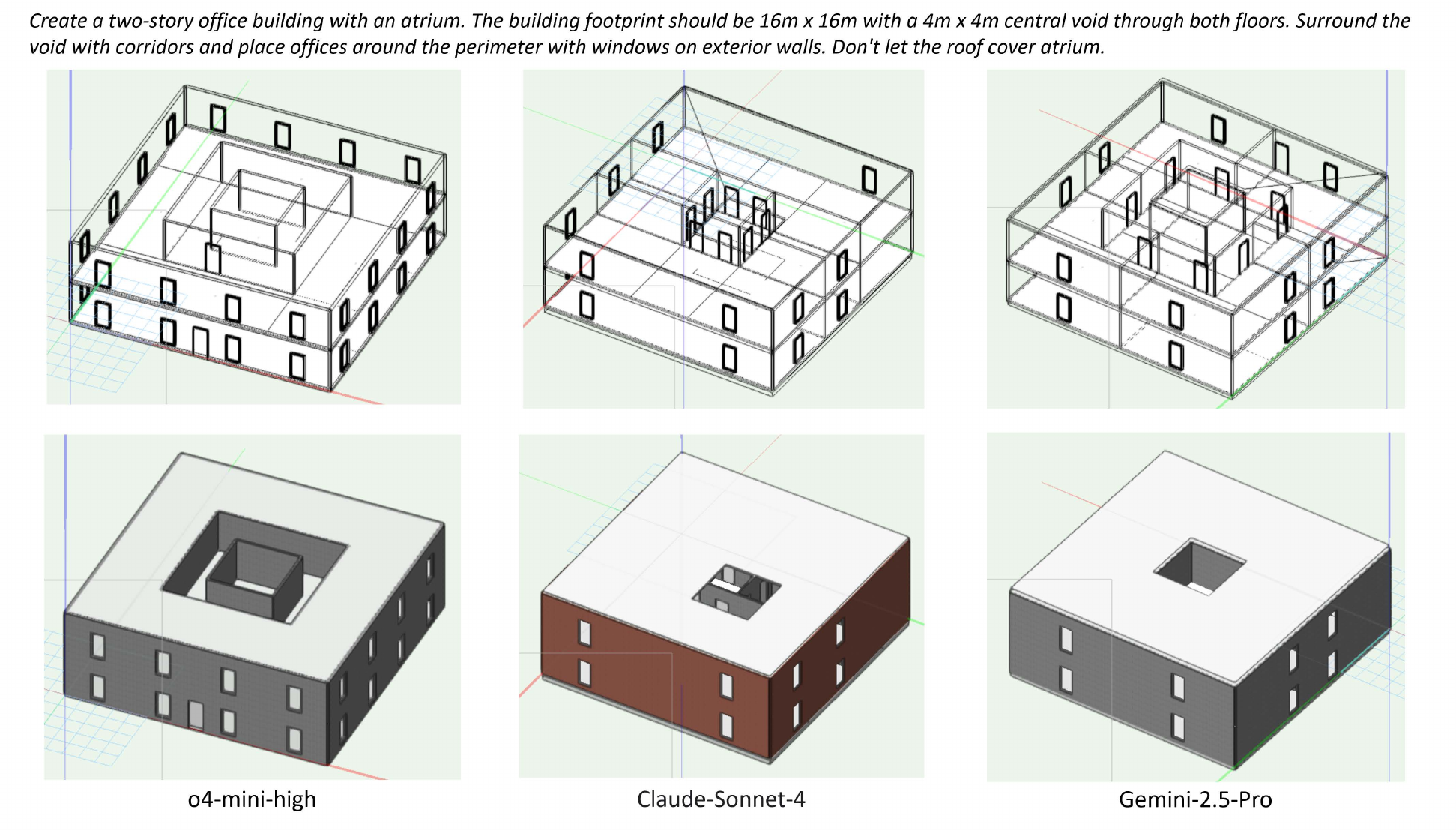}
\caption{The building model in wireframe/rendering mode generated by different LLMs through the proposed framework according to the corresponding text description (Test prompt Nr.23).}
\label{fig:result23}
\end{figure}

\begin{figure}[H]
\centering
\includegraphics[width=0.95\textwidth]{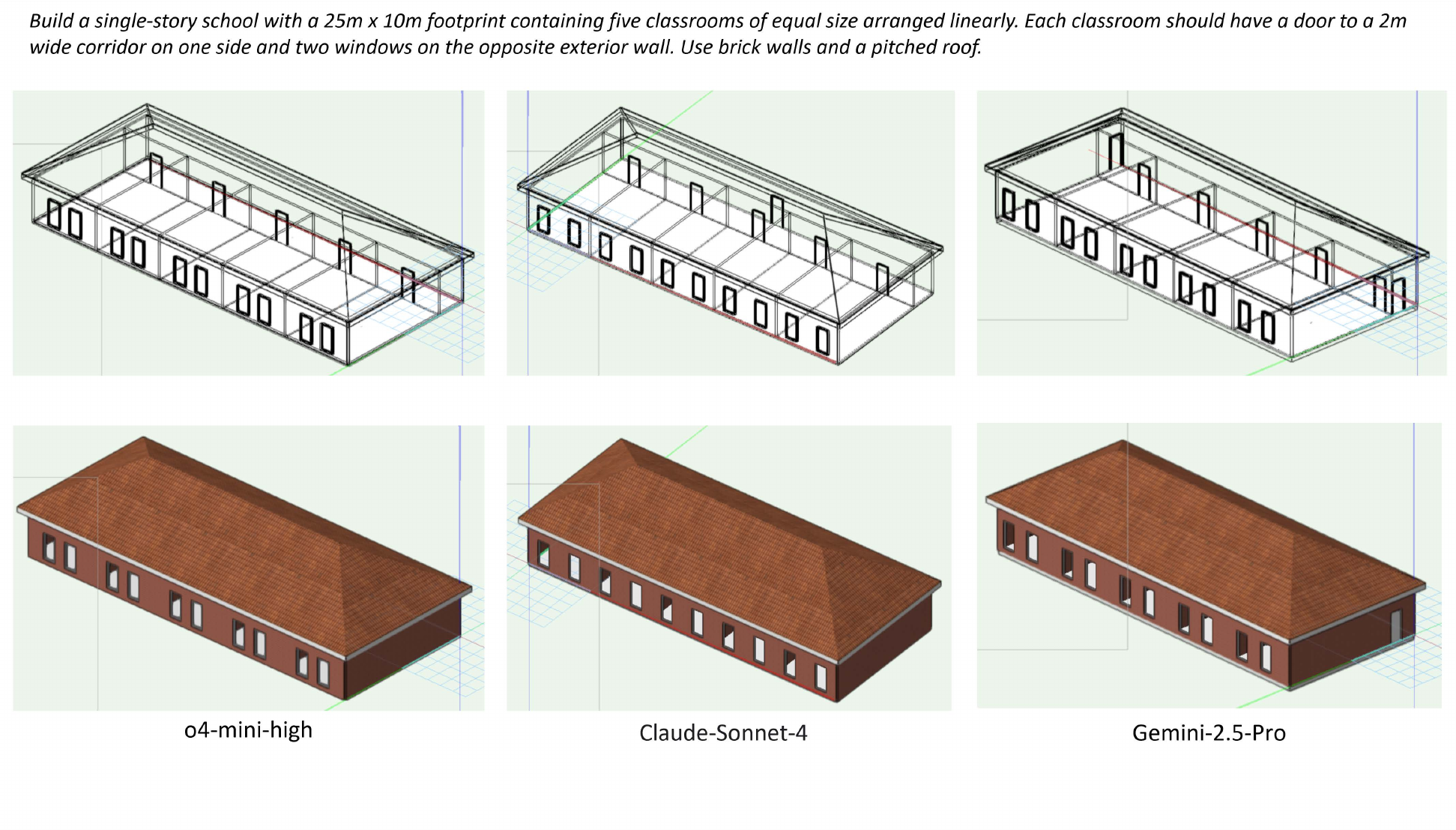}
\caption{The building model in wireframe/rendering mode generated by different LLMs through the proposed framework according to the corresponding text description (Test prompt Nr.24).}
\label{fig:result24}
\end{figure}

\begin{figure}[H]
\centering
\includegraphics[width=0.95\textwidth]{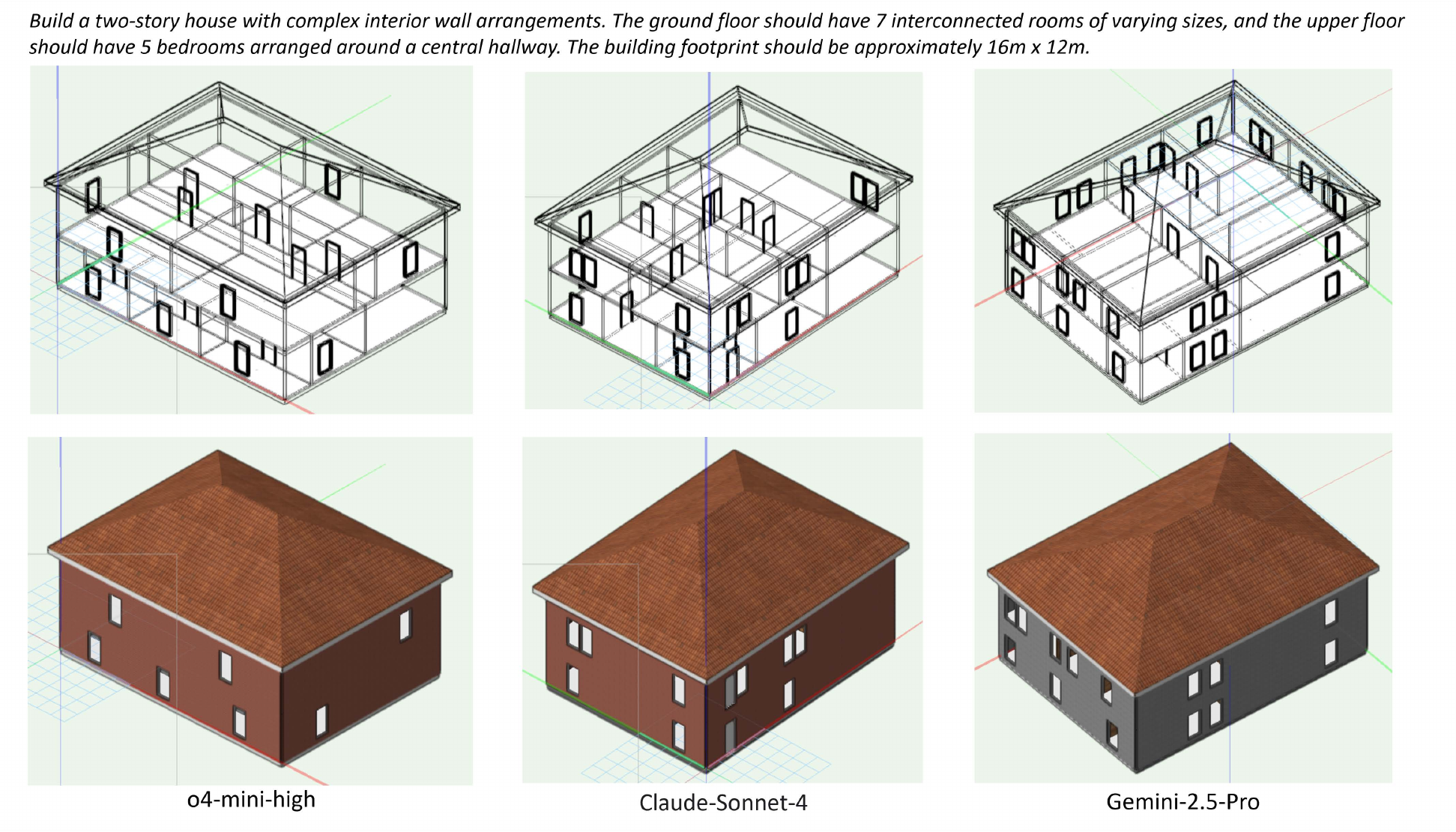}
\caption{The building model in wireframe/rendering mode generated by different LLMs through the proposed framework according to the corresponding text description (Test prompt Nr.25).}
\label{fig:result25}
\end{figure}

\section{Sample conversation log of agents}
\label{app:chat_log}
To better illustrate the internal workings of the proposed LLM-based multi-agent framework, this section presents a detailed conversation log between the agents based on a given sample user instruction (test prompt Nr.6). The BIM model generated by this example can be found in Fig.~\ref{fig:result6}.

In this example, the User wants to create a hexagonal building (1). The \textsc{Instruction Enhancer} agent obtains building design information from the \textsc{Architect} agent through function-calling (2 - 6). It then refines and enhances the user's original instructions into a detailed guidance document by referencing the toolset documentation (7). The \textsc{Programmer} agent analyzes this document and calls the relevant tool functions to write the necessary code (8 - 9). A custom Python interpreter evaluates and runs the code to generate a model in the BIM authoring software (10). The model is then checked by the model checker, and the results (11) are fed back to the \textsc{Reviewer} agent, who generates model optimization suggestions (12 - 13). The \textsc{Programmer} agent applies these suggestions by writing patches to fix issues in the model code (14 -15). The patch code is executed again by the custom Python interpreter (16), and the updated model is sent back to the model checker for re-inspection. The process concludes when the model checker finds no further errors (17).

\lstset{
    breaklines=true, 
    breakatwhitespace=false,  
    basicstyle=\ttfamily\scriptsize,
    escapeinside={@}{@},
}

(1) User Instruction:
\begin{lstlisting}
Design a building with a complex polygonal footprint (e.g., hexagonal). Each side of the hexagon should be 5 meters. Add a slab for the floor and a pitched roof. Include a door on one side and a window on each of the other sides.
\end{lstlisting}

(2) Input prompt for the \textsc{Instruction Enhancer} agent (the filled prompt template and the description of the available function for \textit{function-calling}):
\begin{lstlisting}
You are an experienced Instruction Enhancer who can instruct an AI programmer to write Python scripts in Vectorworks to accomplish user-specified tasks. You are a master at understanding API documentation, adept at breaking down the user's original instructions into subtasks and sub-logic based on the functions available, and able to express the user's description in a more rigorous and detailed manner.
Please give detailed hints on coordinates and dimensions when instructing the programmer. Consult the Architect when you feel you need a floor/building design plan or more architectural context to assist your instructions. If functional areas are given in the design plan, make sure to instruct the programmer to create them. You must refer to architectural knowledge to ensure that your content is spatially and geometrically rational. Use millimeter units. Think step by step. Minimize other prose.
For reference, the following are the API functions available to the programmer, try to give hints on how to use them effectively:
"""
- create_wall: 
    This tool is use to create a wall on a specified layer. By default, the wall is created with a bottom_elevation of 0 and a top_elevation of 3000 relative to this layer.
    Input:
        - st_pt: tuple, the 2D coordinate of the starting point of the wall.
        - ed_pt: tuple, the 2D coordinate of the end point of the wall.
        - layer_uuid: str, the uuid of the story layer where the wall will be created.
    Return:
        - str, the uuid of the new created wall.
- set_wall_elevation: 
    This tool is used to set the top/bottom elevation of a wall. Subtracting these two is the height of the wall itself.
    Input:
        - uuid: str, the uuid of the wall object.
        - top_elevation: float, the vertical distance from the top of the wall to the story layer where the wall was originally created.
        - bottom_elevation: float, the vertical distance from the bottom of the wall to the story layer where the wall was originally created.
    Return:
        - str, the uuid of the wall object that has been modified.
...
@{\textit{(... The remaining content is identical to the toolset documentation in Appendix I)}}@
"""

Now, based on the conversation given below, please relay the user's instruction and complete your answer. You don't need to write code, just guide the programmer step by step. Let's think step by step.

@{\textbf{User:}}@ Design a building with a complex polygonal footprint (e.g., hexagonal). Each side of the hexagon should be 5 meters. Add a slab for the floor and a pitched roof. Include a door on one side and a window on each of the other sides.
Instruction Enhancer: 
\end{lstlisting}

\begin{lstlisting}
 "tools": [{
    "type": "function",
    "function": {
        "name": "Architect",
        "description": "Generate a floor or building design in structured text format based on the query, providing architectural knowledge.",
        "parameters": {
            "type": "object",
            "properties": {
                "query": {
                    "type": "string",
                    "description": "The requirements for the design, the more detailed the better."
                },
            },
            "required": ["query"],
        },
    }
}]
\end{lstlisting}

(3) The intermediate output (JSON) generated when the \textsc{Instruction Enhancer} agent consults the \textsc{Architect} agent using function-calling:
\begin{lstlisting}
{
  "id": "call_vUg7MUxZHpPy1WIJhszmFID8",
  "type": "function",
  "function": {
    "name": "Architect",
    "arguments": "{\"query\":\"Design a building with a hexagonal footprint. Each side of the hexagon should be 5000 mm. Add a slab for the floor and a pitched roof. Include a door on one side and a window on each of the other sides.\"}"
  }
}
\end{lstlisting}

(4) Input prompt for the \textsc{Architect} agent:
\begin{lstlisting}
You are an experienced architect who can design floors/buildings based on the user's needs. You will use your extensive architectural knowledge to expand and supplement the user's original description and ultimately express your design in structured text format.
Depending on the user's specific needs, try to include in the output the starting and ending points of each wall, the location of windows and doors (offset relative to the start of the wall), boundaries of interior rooms/functional areas and the position and geometric details of other components required for a complete building.
Please refer to basic architectural rules, such like: 
- Foundation: Ensure a solid foundation slab that can support the entire structure.
- Wall Configuration: Arrange walls to define the building's perimeter and internal spaces. Ensure that load-bearing walls are adequately spaced and placed to distribute the weight of the structure evenly. Set the wall elevations properly for each floor.
- Slab Design: Place slabs for each floor. They should be level and supported by the walls, providing stability and separating different floors.
- Roof Construction: Design the roof to cover the entire building, protecting it from weather conditions.
- Window Placement: Install windows strategically to provide natural light and ventilation to rooms. Ensure window locations are proportionate to the room size.
- Door Placement: Position doors for easy access to different rooms and areas. Main entrance to the building should be prominent and easy to locate, with interior doors facilitating smooth movement.
- Interior Layout: Organize and define the interior room layout logically. Use interior walls to separate different functional rooms and ensure easy flow between them with appropriately placed doors.
- Structural Integrity: Ensure all elements (walls, slabs, roof) are securely connected and stable.
- Compliance: Avoid clashing/overlapping building components, such as overlapping partitions between different areas and overlapping window and door locations. Adjacent rooms can share internal partitions. Rooms can also utilize exterior walls.

Make your design spatially and geometrically rational. Use millimeter units. Minimize other prose.

Here is a sample conversation:
"""
User: I want to build an office building. I want the building to have 3 floors and the layout of each floor to be the same. Each floor has 6 rooms, 3 on each side, separated by a 3 meter wide corridor. Each room has a door and a window. The door to each room should be on the wall on one side of the corridor and the window should be on the outside wall of the building.
Architect: **3-Floor Office Building Design**
**Foundation:**
- Rectangular foundation slab: 30000mm x 15000mm

**Ground Floor Plan:**
1. **Perimeter Walls:**
- Wall A: (0,0) to (30000,0)
- Wall B: (30000,0) to (30000,15000)
- Wall C: (30000,15000) to (0,15000)
- Wall D: (0,15000) to (0,0)

2. **Functional Areas**
Boundary in format (x_min,y_min),(x_max,y_max):
- Room 1: (0,0), (10000,6000)
- Room 2: (10000,0), (20000,6000)
- Room 3: (20000,0), (30000,6000)
- Room 4: (0,9000), (10000,15000)
- Room 5: (10000,9000), (20000,15000)
- Room 6: (20000,9000), (30000,15000)
- Central corridor: (0,6000), (30000,9000)

3. **Internal Corridor Walls:**
- Wall E: (0,6000) to (30000,6000)
- Wall F: (0,9000) to (30000,9000)

4. **Room Dividing Walls:**
- Wall G: (10000,0) to (10000,6000)
- Wall H: (20000,0) to (20000,6000)
- Wall I: (10000,9000) to (10000,15000)
- Wall J: (20000,9000) to (20000,15000)

5. **Doors:**
Insertion offset of each room door relative to the start of the corresponding wall:
- Room 1 door on corridor wall E: 5000
- Room 2 door on corridor wall E: 15000
- Room 3 door on corridor wall E: 25000
- Room 4 door on corridor wall F: 5000
- Room 5 door on corridor wall F: 15000
- Room 6 door on corridor wall F: 25000

6. **Windows:**
Insertion offset of each room window relative to the start of the corresponding wall:
- Room 1 window on wall A: 5000
- Room 2 window on wall A: 15000
- Room 3 window on wall A: 25000
- Room 4 window on wall C: 5000
- Room 5 window on wall C: 15000
- Room 6 window on wall C: 25000

**First Floor Plan:**
- Identical to Ground Floor Plan

**Second Floor Plan:**
- Identical to Ground Floor Plan

**Roof Construction:**
- Roof covering entire building: (0,0) to (30000,0) to (30000,15000) to (0,15000) to (0,0).

**Slab Design:**
- Create slabs for each floor supported by perimeter and internal walls. Slabs covering entire floor area with same dimensions as foundation.

**Summary:**
- Building dimensions: 30000mm x 15000mm x 3 floors.
- Each floor has 6 rooms, 3 on each side of a central corridor.
"""

The user now provides the following instruction, please generate the plan as an architect. Let's think step by step.
User: Design a building with a hexagonal footprint. Each side of the hexagon should be 5000 mm. Add a slab for the floor and a pitched roof. Include a door on one side and a window on each of the other sides.
Architect: 
\end{lstlisting}

(5) Output from the \textsc{Architect} agent:
\begin{lstlisting}
**Hexagonal Building Design**

**Foundation:**
- Hexagonal foundation slab with each side 5000mm.

**Ground Floor Plan:**
1. **Perimeter Walls:**
- Wall A: (0,0) to (5000,0)
- Wall B: (5000,0) to (7500,4330)
- Wall C: (7500,4330) to (5000,8660)
- Wall D: (5000,8660) to (0,8660)
- Wall E: (0,8660) to (-2500,4330)
- Wall F: (-2500,4330) to (0,0)

2. **Functional Areas:**
- Single open space within the hexagonal boundary.

3. **Doors:**
Insertion offset of the door relative to the start of the corresponding wall:
- Door on Wall A: 2500mm

4. **Windows:**
Insertion offset of each window relative to the start of the corresponding wall:
- Window on Wall B: 2500mm
- Window on Wall C: 2500mm
- Window on Wall D: 2500mm
- Window on Wall E: 2500mm
- Window on Wall F: 2500mm

**Roof Construction:**
- Pitched roof covering the entire hexagonal building. The roof will have a central ridge running from the midpoint of Wall A to the midpoint of Wall D, with slopes extending to the edges of the hexagon.

**Slab Design:**
- Hexagonal slab covering the entire floor area with the same dimensions as the foundation.

**Summary:**
- Building dimensions: Hexagonal footprint with each side 5000mm.
- Single open space within the hexagonal boundary.
- One door on Wall A.
- One window on each of the other walls (B, C, D, E, F).
- Pitched roof covering the entire building.
\end{lstlisting}

(6) After executing the function-calling, the intermediate input for the \textsc{Instruction Enhancer} agent considering the \textsc{Architect}'s output:
\begin{lstlisting}
You are an experienced Instruction Enhancer who can instruct an AI programmer...
@{\textit{(... The remaining content is identical to the (2) Input prompt for the Instruction Enhancer agent)}}@

{
    'tool_call_id': 'call_vUg7MUxZHpPy1WIJhszmFID8',
    'role': 'tool', 
    'name': 'Architect', 
    'content': '**Hexagonal Building Design**\n\n**Foundation:**\n- Hexagonal foundation slab with each side 5000mm...
    @{\textit{(... The remaining content is identical to the (5) Output from the Architect agent)}}@'
}
\end{lstlisting}

(7) Output from the \textsc{Instruction Enhancer} agent:
\begin{lstlisting}
### Step-by-Step Instructions for the Programmer

#### 1. Create the Story Layer
- **Action**: Create a new story layer for the ground floor.
- **Function**: `create_story_layer`
- **Parameters**:
  - `layer_name`: "Ground Floor"
  - `elevation`: 0
  - `floor_index`: 1

#### 2. Create the Hexagonal Footprint
- **Action**: Create a hexagonal polygon with each side 5000mm.
- **Function**: `create_polygon`
- **Parameters**:
  - `vertices`: [(0,0), (5000,0), (7500,4330), (5000,8660), (0,8660), (-2500,4330)]
  - `layer_uuid`: UUID of "Ground Floor"

#### 3. Create the Slab
- **Action**: Create a slab using the hexagonal polygon.
- **Function**: `create_slab`
- **Parameters**:
  - `profile_id`: UUID of the hexagonal polygon
  - `layer_uuid`: UUID of "Ground Floor"

#### 4. Create Perimeter Walls
- **Action**: Create walls along the edges of the hexagon.
- **Function**: `create_wall`
- **Parameters**:
  - `st_pt`: Starting point of the wall
  - `ed_pt`: Ending point of the wall
  - `layer_uuid`: UUID of "Ground Floor"

- **Walls**:
  - Wall A: (0,0) to (5000,0)
  - Wall B: (5000,0) to (7500,4330)
  - Wall C: (7500,4330) to (5000,8660)
  - Wall D: (5000,8660) to (0,8660)
  - Wall E: (0,8660) to (-2500,4330)
  - Wall F: (-2500,4330) to (0,0)

#### 5. Add Door to Wall A
- **Action**: Add a door to Wall A.
- **Function**: `add_door_to_wall`
- **Parameters**:
  - `wall_uuid`: UUID of Wall A
  - `door_elevation`: 0
  - `door_offset`: 2500
  - `door_name`: "Main Door"

#### 6. Add Windows to Other Walls
- **Action**: Add a window to each of the other walls (B, C, D, E, F).
- **Function**: `add_window_to_wall`
- **Parameters**:
  - `wall_uuid`: UUID of the wall
  - `window_elevation`: 1000 (assuming standard window height)
  - `window_offset`: 2500
  - `window_name`: "Standard Window"

- **Windows**:
  - Wall B: UUID of Wall B
  - Wall C: UUID of Wall C
  - Wall D: UUID of Wall D
  - Wall E: UUID of Wall E
  - Wall F: UUID of Wall F

#### 7. Create the Pitched Roof
- **Action**: Create a pitched roof covering the entire hexagonal building.
- **Function**: `create_pitched_roof`
- **Parameters**:
  - `profile_id`: UUID of the hexagonal polygon
  - `layer_uuid`: UUID of "Ground Floor"
  - `slope`: 30 (example slope)
  - `eave_overhang`: 500 (example overhang)
  - `eave_height`: 3000 (height of the walls)
  - `roof_thickness`: 200 (example thickness)

#### 8. Set Roof Style
- **Action**: Set the style of the pitched roof.
- **Function**: `set_pitched_roof_style`
- **Parameters**:
  - `roof_id`: UUID of the created roof
  - `style_name`: "Sloped Wood Struct Insul Flat Clay Tile"

### Summary
1. Create the story layer.
2. Create the hexagonal footprint.
3. Create the slab.
4. Create perimeter walls.
5. Add a door to Wall A.
6. Add windows to Walls B, C, D, E, and F.
7. Create the pitched roof.
8. Set the roof style.

This step-by-step guide ensures that the building is designed with a hexagonal footprint, a slab for the floor, a pitched roof, a door on one side, and windows on the other sides.
\end{lstlisting}

(8) Input prompt for the \textsc{Programmer} agent:
\begin{lstlisting}
You are an AI programmer that specialized in coding. Your job is to write Python code that invoke suitable pre-defined tool functions to implement Instruction Enhancer's requirement. Your code style is concise, efficient and accurate.
To help with that, you have access to a set of tools. Each tool is a Python function and has a description explaining the task it performs, the inputs it expects and the outputs it returns. You should think step by step and write all your Python code in a single code block. You can only give one reply for each conversation turn. Minimize other prose. Do not make any assumptions or examples in your code.
Accessible tools:
"""
- create_wall: 
    This tool is use to create a wall on a specified layer. By default, the wall is created with a bottom_elevation of 0 and a top_elevation of 3000 relative to this layer.
    Input:
        - st_pt: tuple, the 2D coordinate of the starting point of the wall.
        - ed_pt: tuple, the 2D coordinate of the end point of the wall.
        - layer_uuid: str, the uuid of the story layer where the wall will be created.
    Return:
        - str, the uuid of the new created wall.
- set_wall_elevation: 
    This tool is used to set the top/bottom elevation of a wall. Subtracting these two is the height of the wall itself.
    Input:
        - uuid: str, the uuid of the wall object.
        - top_elevation: float, the vertical distance from the top of the wall to the story layer where the wall was originally created.
        - bottom_elevation: float, the vertical distance from the bottom of the wall to the story layer where the wall was originally created.
    Return:
        - str, the uuid of the wall object that has been modified. 
...
@{\textit{(... The remaining content is identical to the toolset documentation in Appendix I)}}@
"""
In your code, you should only use the tool functions that are accessible to you, rather than attempting to invent new tools.
Use millimeter units. If necessary, you can import and use the Python standard library. You do not need to import tools.

Now, based on the conversation given below, please complete your answer as a programmer:

Instruction Enhancer: ### Step-by-Step Instructions for the Programmer
#### 1. Create the Story Layer
- **Action**: Create a new story layer for the ground floor...
@{\textit{(... The remaining content is identical to the (7) Output from the Instruction Enhancer agent)}}@
Programmer: 
\end{lstlisting}

(9) Output from the \textsc{Programmer} agent:
\begin{lstlisting}
```py
# Step 1: Create the Story Layer
ground_floor_uuid = create_story_layer(layer_name="Ground Floor", elevation=0, floor_index=1)

# Step 2: Create the Hexagonal Footprint
hexagon_vertices = [(0, 0), (5000, 0), (7500, 4330), (5000, 8660), (0, 8660), (-2500, 4330)]
hexagon_uuid = create_polygon(vertices=hexagon_vertices, layer_uuid=ground_floor_uuid)

# Step 3: Create the Slab
slab_uuid = create_slab(profile_id=hexagon_uuid, layer_uuid=ground_floor_uuid)

# Step 4: Create Perimeter Walls
wall_A_uuid = create_wall(st_pt=(0, 0), ed_pt=(5000, 0), layer_uuid=ground_floor_uuid)
wall_B_uuid = create_wall(st_pt=(5000, 0), ed_pt=(7500, 4330), layer_uuid=ground_floor_uuid)
wall_C_uuid = create_wall(st_pt=(7500, 4330), ed_pt=(5000, 8660), layer_uuid=ground_floor_uuid)
wall_D_uuid = create_wall(st_pt=(5000, 8660), ed_pt=(0, 8660), layer_uuid=ground_floor_uuid)
wall_E_uuid = create_wall(st_pt=(0, 8660), ed_pt=(-2500, 4330), layer_uuid=ground_floor_uuid)
wall_F_uuid = create_wall(st_pt=(-2500, 4330), ed_pt=(0, 0), layer_uuid=ground_floor_uuid)

# Step 5: Add Door to Wall A
door_uuid = add_door_to_wall(wall_uuid=wall_A_uuid, door_elevation=0, door_offset=2500, door_name="Main Door")

# Step 6: Add Windows to Other Walls
window_B_uuid = add_window_to_wall(wall_uuid=wall_B_uuid, window_elevation=1000, window_offset=2500, window_name="Standard Window")
window_C_uuid = add_window_to_wall(wall_uuid=wall_C_uuid, window_elevation=1000, window_offset=2500, window_name="Standard Window")
window_D_uuid = add_window_to_wall(wall_uuid=wall_D_uuid, window_elevation=1000, window_offset=2500, window_name="Standard Window")
window_E_uuid = add_window_to_wall(wall_uuid=wall_E_uuid, window_elevation=1000, window_offset=2500, window_name="Standard Window")
window_F_uuid = add_window_to_wall(wall_uuid=wall_F_uuid, window_elevation=1000, window_offset=2500, window_name="Standard Window")

# Step 7: Create the Pitched Roof
roof_uuid = create_pitched_roof(profile_id=hexagon_uuid, layer_uuid=ground_floor_uuid, slope=30, eave_overhang=500, eave_height=3000, roof_thickness=200)

# Step 8: Set Roof Style
set_pitched_roof_style(roof_id=roof_uuid, style_name="Sloped Wood Struct Insul Flat Clay Tile")
```
\end{lstlisting}

(10) Output from the custom Python interpreter:
\begin{lstlisting}
==Result==
Code executed successfully!
\end{lstlisting}

(11) Processed feedback from the Solibri model checker:
\begin{lstlisting}
- Issue: No Space Components in prompt_27c53c4b-dd9b-478b-9ead-e1f4e69d18a9
- Issue description: Model Should Have Spaces
The rule checks if functional areas (spaces) are present in the model
Desired resolution: Create meaningful functional areas in the building by considering building requirements
No Space Components in prompt_27c53c4b-dd9b-478b-9ead-e1f4e69d18a9
prompt_27c53c4b-dd9b-478b-9ead-e1f4e69d18a9 doesn't  contain any Space components
- Related element uuids: []
\end{lstlisting}

(12) Input prompt for the \textsc{Reviewer} agent (entering quality optimization loop):
\begin{lstlisting}
You are a senior code reviewer. You review Python code written by an AI programmer that calls specific APIs to generate buildings in Vectorworks. Your job is to propose solutions to the programmer for issues that arise in the generated building model, based on the API documentation. Issues are listed in the form of bullet points. Each issue contains a description and list of uuids of the corresponding elements in the generated building model. These uuids can be used to directly manipulate the related elements in Vectorworks via the API functions. 

Here is the information you can refer to:
Code from the AI programmer:
```py
# Step 1: Create the Story Layer
ground_floor_uuid = create_story_layer(layer_name="Ground Floor", elevation=0, floor_index=1)
...
@{\textit{(... The remaining content is identical to the (9) Output from the Programmer agent)}}@
```

Issues in the generated buildings:
- Issue: No Space Components in prompt_27c53c4b-dd9b-478b-9ead-e1f4e69d18a9
- Issue description: Model Should Have Spaces
The rule checks if functional areas (spaces) are present in the model
Desired resolution: Create meaningful functional areas in the building by considering building requirements
No Space Components in prompt_27c53c4b-dd9b-478b-9ead-e1f4e69d18a9
prompt_27c53c4b-dd9b-478b-9ead-e1f4e69d18a9 doesn't  contain any Space components
- Related element uuids: []

API documentation:
"""
- create_wall: 
    This tool is use to create a wall on a specified layer. By default, the wall is created with a bottom_elevation of 0 and a top_elevation of 3000 relative to this layer.
    Input:
        - st_pt: tuple, the 2D coordinate of the starting point of the wall.
        - ed_pt: tuple, the 2D coordinate of the end point of the wall.
        - layer_uuid: str, the uuid of the story layer where the wall will be created.
    Return:
        - str, the uuid of the new created wall.
- set_wall_elevation: 
    This tool is used to set the top/bottom elevation of a wall. Subtracting these two is the height of the wall itself.
    Input:
        - uuid: str, the uuid of the wall object.
        - top_elevation: float, the vertical distance from the top of the wall to the story layer where the wall was originally created.
        - bottom_elevation: float, the vertical distance from the bottom of the wall to the story layer where the wall was originally created.
    Return:
        - str, the uuid of the wall object that has been modified.
...
@{\textit{(... The remaining content is identical to the toolset documentation in Appendix I)}}@
"""
Please provide your guidance below as a reviewer. Use millimeter units. Think step by step. Try to avoid duplicating the programmer's code, but rather suggest code fixes.
Reviewer: 
\end{lstlisting}

(13) Output from the \textsc{Reviewer} agent:
\begin{lstlisting}
To address the issue of missing space components in the generated building model, we need to create functional areas (spaces) within the model. This will help define the interior layout of the building. Here's a step-by-step guide on how to achieve this:

### Step-by-Step Solution:

1. **Identify the Interior Layout:**
   - Since the building has a hexagonal footprint, we can divide the interior into functional areas. For simplicity, let's assume we want to create a central living area and six surrounding rooms.

2. **Define the Vertices for Functional Areas:**
   - We will use the vertices of the hexagon to define the boundaries of the functional areas. The central area can be defined by a smaller hexagon inside the main hexagon, and the surrounding rooms can be defined by the remaining space.

3. **Create Functional Areas:**
   - Use the `create_functional_area` API to create the central living area and the surrounding rooms.

### Suggested Code Fixes:

```python
# Step 9: Create Functional Areas

# Define the vertices for the central living area (smaller hexagon inside the main hexagon)
central_hexagon_vertices = [(1250, 2165), (3750, 2165), (5000, 4330), (3750, 6495), (1250, 6495), (0, 4330)]
central_area_uuid = create_functional_area(vertices=central_hexagon_vertices, name="Central Living Area", layer_uuid=ground_floor_uuid)

# Define the vertices for the surrounding rooms
# Room 1 (Top-Left)
room_1_vertices = [(0, 0), (2500, 0), (1250, 2165), (0, 2165)]
room_1_uuid = create_functional_area(vertices=room_1_vertices, name="Room 1", layer_uuid=ground_floor_uuid)

# Room 2 (Top-Right)
room_2_vertices = [(2500, 0), (5000, 0), (3750, 2165), (1250, 2165)]
room_2_uuid = create_functional_area(vertices=room_2_vertices, name="Room 2", layer_uuid=ground_floor_uuid)

# Room 3 (Right)
room_3_vertices = [(5000, 0), (7500, 4330), (5000, 4330), (3750, 2165)]
room_3_uuid = create_functional_area(vertices=room_3_vertices, name="Room 3", layer_uuid=ground_floor_uuid)

# Room 4 (Bottom-Right)
room_4_vertices = [(7500, 4330), (5000, 8660), (3750, 6495), (5000, 4330)]
room_4_uuid = create_functional_area(vertices=room_4_vertices, name="Room 4", layer_uuid=ground_floor_uuid)

# Room 5 (Bottom-Left)
room_5_vertices = [(5000, 8660), (0, 8660), (1250, 6495), (3750, 6495)]
room_5_uuid = create_functional_area(vertices=room_5_vertices, name="Room 5", layer_uuid=ground_floor_uuid)

# Room 6 (Left)
room_6_vertices = [(0, 8660), (-2500, 4330), (0, 4330), (1250, 6495)]
room_6_uuid = create_functional_area(vertices=room_6_vertices, name="Room 6", layer_uuid=ground_floor_uuid)
```

### Explanation:

1. **Central Living Area:**
   - We defined a smaller hexagon inside the main hexagon to represent the central living area. The vertices are calculated to be proportionally smaller than the main hexagon.

2. **Surrounding Rooms:**
   - We divided the remaining space into six rooms, each defined by a set of vertices. These rooms are created around the central living area.

3. **API Usage:**
   - The `create_functional_area` API is used to create each functional area with the specified vertices and names.

By following these steps, we ensure that the building model now includes functional areas, addressing the issue of missing space components. This approach maintains the integrity of the original design while adding the necessary interior layout.
\end{lstlisting}

(14) Input for the \textsc{Programmer} agent:
\begin{lstlisting}
You are an AI programmer that specialized in coding. Your job is to write Python code that invoke suitable pre-defined tool functions to implement Instruction Enhancer's requirement. Your code style is concise, efficient and accurate.
To help with that, you have access to a set of tools. Each tool is a Python function and has a description explaining the task it performs, the inputs it expects and the outputs it returns. You should think step by step and write all your Python code in a single code block. You can only give one reply for each conversation turn. Minimize other prose. Do not make any assumptions or examples in your code.
Accessible tools:
"""
- create_wall: 
    This tool is use to create a wall on a specified layer. By default, the wall is created with a bottom_elevation of 0 and a top_elevation of 3000 relative to this layer.
    Input:
        - st_pt: tuple, the 2D coordinate of the starting point of the wall.
        - ed_pt: tuple, the 2D coordinate of the end point of the wall.
        - layer_uuid: str, the uuid of the story layer where the wall will be created.
    Return:
        - str, the uuid of the new created wall.
- set_wall_elevation: 
    This tool is used to set the top/bottom elevation of a wall. Subtracting these two is the height of the wall itself.
    Input:
        - uuid: str, the uuid of the wall object.
        - top_elevation: float, the vertical distance from the top of the wall to the story layer where the wall was originally created.
        - bottom_elevation: float, the vertical distance from the bottom of the wall to the story layer where the wall was originally created.
    Return:
        - str, the uuid of the wall object that has been modified.
...
@{\textit{(... The remaining content is identical to the toolset documentation in Appendix I)}}@
"""
In your code, you should only use the tool functions that are accessible to you, rather than attempting to invent new tools.
Use millimeter units. If necessary, you can import and use the Python standard library. You do not need to import tools.

Now, based on the conversation given below, please complete your answer as a programmer:

Instruction Enhancer: To address the issue of missing space components in the generated building model, we need to create functional areas (spaces) within the model...
@{\textit{(... The remaining content is identical to the (13) Output from the Reviewer agent)}}@
Programmer: 
\end{lstlisting}

(15) Output from the \textsc{Programmer} agent:
\begin{lstlisting}
```python
# Step 1: Define the vertices for the central living area (smaller hexagon inside the main hexagon)
central_hexagon_vertices = [(1250, 2165), (3750, 2165), (5000, 4330), (3750, 6495), (1250, 6495), (0, 4330)]

# Step 2: Create the central living area
central_area_uuid = create_functional_area(vertices=central_hexagon_vertices, name="Central Living Area", layer_uuid=ground_floor_uuid)

# Step 3: Define the vertices for the surrounding rooms
# Room 1 (Top-Left)
room_1_vertices = [(0, 0), (2500, 0), (1250, 2165), (0, 2165)]
room_1_uuid = create_functional_area(vertices=room_1_vertices, name="Room 1", layer_uuid=ground_floor_uuid)

# Room 2 (Top-Right)
room_2_vertices = [(2500, 0), (5000, 0), (3750, 2165), (1250, 2165)]
room_2_uuid = create_functional_area(vertices=room_2_vertices, name="Room 2", layer_uuid=ground_floor_uuid)

# Room 3 (Right)
room_3_vertices = [(5000, 0), (7500, 4330), (5000, 4330), (3750, 2165)]
room_3_uuid = create_functional_area(vertices=room_3_vertices, name="Room 3", layer_uuid=ground_floor_uuid)

# Room 4 (Bottom-Right)
room_4_vertices = [(7500, 4330), (5000, 8660), (3750, 6495), (5000, 4330)]
room_4_uuid = create_functional_area(vertices=room_4_vertices, name="Room 4", layer_uuid=ground_floor_uuid)

# Room 5 (Bottom-Left)
room_5_vertices = [(5000, 8660), (0, 8660), (1250, 6495), (3750, 6495)]
room_5_uuid = create_functional_area(vertices=room_5_vertices, name="Room 5", layer_uuid=ground_floor_uuid)

# Room 6 (Left)
room_6_vertices = [(0, 8660), (-2500, 4330), (0, 4330), (1250, 6495)]
room_6_uuid = create_functional_area(vertices=room_6_vertices, name="Room 6", layer_uuid=ground_floor_uuid)
```
\end{lstlisting}

(16) Output from the custom Python interpreter:
\begin{lstlisting}
==Result==
Code executed successfully!
\end{lstlisting}

(17) Processed feedback from the Solibri model checker:
\begin{lstlisting}
No issue was found in the model!
\end{lstlisting}

\section{Supplemental materials}
\begin{enumerate}
    \item Demo video is available online in the ASCE Library (www.ascelibrary.org).
    \item Expert assessment protocol is available online in the ASCE Library (www.ascelibrary.org).
\end{enumerate}

%
% Here's the list of references:
%
% \label{section:references}
\bibliography{ascexmpl-new}

\begin{thebibliography}{}

\bibitem[\protect\citeauthoryear{}{Amor and Dimyadi}{2021}]{AMOR2021}
Amor, R. and Dimyadi, J. (2021).
\newblock ``The promise of automated compliance checking.''\ {\em Developments in the Built Environment}, 5, 100039.

\bibitem[\protect\citeauthoryear{}{Anthropic}{2025}]{anthropic2025claudesonnet4}
Anthropic (2025).
\newblock ``Claude sonnet 4, $<$https://www.anthropic.com/claude/sonnet$>$\ (May).
\newblock Accessed: 2025-05-29.

\bibitem[\protect\citeauthoryear{}{Bloch et~al.\@}{2024}]{EC32024_247}
Bloch, T., Wu, J., Alnuzha, M., and Borrmann, A. (2024).
\newblock ``Leveraging graph based semantic enrichment for enhanced automated code-checking.''\ {\em Proceedings of the 2024 European Conference on Computing in Construction}, Vol.~5 of {\em Computing in Construction}, Chania, Greece, European Council on Computing in Construction\ (July).

\bibitem[\protect\citeauthoryear{}{Borrmann et~al.\@}{2018}]{Borrmann:2018:BIMIntro}
Borrmann, A., König, M., Koch, C., and Beetz, J. (2018).
\newblock ``Building information modeling: Why? what? how?.''\ {\em Building Information Modeling - Technology foundations and industry practice}, A. Borrmann, M. König, C. Koch, and J. Beetz, eds., Vol.~1, Springer,  1--24.

\bibitem[\protect\citeauthoryear{}{Brown et~al.\@}{2020}]{brown2020languagemodelsfewshotlearners}
Brown, T.~B., Mann, B., Ryder, N., Subbiah, M., Kaplan, J., Dhariwal, P., Neelakantan, A., Shyam, P., Sastry, G., Askell, A., Agarwal, S., Herbert-Voss, A., Krueger, G., Henighan, T., Child, R., Ramesh, A., Ziegler, D.~M., Wu, J., Winter, C., Hesse, C., Chen, M., Sigler, E., Litwin, M., Gray, S., Chess, B., Clark, J., Berner, C., McCandlish, S., Radford, A., Sutskever, I., and Amodei, D. (2020).
\newblock ``Language models are few-shot learners, $<$https://arxiv.org/abs/2005.14165$>$.

\bibitem[\protect\citeauthoryear{}{Chen et~al.\@}{2023}]{buildings13071861}
Chen, J., Shao, Z., and Hu, B. (2023).
\newblock ``Generating interior design from text: A new diffusion model-based method for efficient creative design.''\ {\em Buildings}, 13(7).

\bibitem[\protect\citeauthoryear{}{de~Miguel~Rodríguez et~al.\@}{2020}]{Caparrini_2020}
de~Miguel~Rodríguez, J., Villafañe, M.~E., Piškorec, L., and Sancho~Caparrini, F. (2020).
\newblock ``Generation of geometric interpolations of building types with deep variational autoencoders.''\ {\em Design Science}, 6, e34.

\bibitem[\protect\citeauthoryear{}{DeepMind}{2025}]{google2025gemini25pro}
DeepMind, G. (2025).
\newblock ``Gemini 2.5 pro: Our most intelligent ai model, $<$https://blog.google/technology/google-deepmind/gemini-model-thinking-updates-march-2025/$>$\ (March).
\newblock Accessed: 2025-05-29.

\bibitem[\protect\citeauthoryear{}{Dong et~al.\@}{2024}]{dong2024surveyincontextlearning}
Dong, Q., Li, L., Dai, D., Zheng, C., Ma, J., Li, R., Xia, H., Xu, J., Wu, Z., Chang, B., Sun, X., Li, L., and Sui, Z. (2024).
\newblock ``A survey on in-context learning, $<$https://arxiv.org/abs/2301.00234$>$.

\bibitem[\protect\citeauthoryear{}{Du et~al.\@}{2024a}]{Du:2024:command_recommender}
Du, C., Deng, Z., Nousias, S., and Borrmann, A. (2024a).
\newblock ``Towards commands recommender system in bim authoring tool using transformers.''\ {\em Proc. of the 31th Int. Conference on Intelligent Computing in Engineering (EG-ICE)}\ (Jul).

\bibitem[\protect\citeauthoryear{}{Du et~al.\@}{2024b}]{Du:2024:Copilot_BIM}
Du, C., Nousias, S., and Borrmann, A. (2024b).
\newblock ``Towards a copilot in {BIM} authoring tool using large language model based agent for intelligent human-machine interaction.''\ {\em Proc. of the 31th Int. Conference on Intelligent Computing in Engineering (EG-ICE)}\ (Jul).

\bibitem[\protect\citeauthoryear{}{Eastman et~al.\@}{2009}]{Eastman2009}
Eastman, C., min Lee, J., suk Jeong, Y., and kook Lee, J. (2009).
\newblock ``Automatic rule-based checking of building designs.''\ {\em Automation in Construction}, 18, 1011--1033.

\bibitem[\protect\citeauthoryear{}{Edge et~al.\@}{2024}]{edge2024localglobalgraphrag}
Edge, D., Trinh, H., Cheng, N., Bradley, J., Chao, A., Mody, A., Truitt, S., and Larson, J. (2024).
\newblock ``From local to global: A graph rag approach to query-focused summarization, $<$https://arxiv.org/abs/2404.16130$>$.

\bibitem[\protect\citeauthoryear{}{Ennemoser and Mayrhofer-Hufnagl}{2023}]{Ennemoser:2023}
Ennemoser, B. and Mayrhofer-Hufnagl, I. (2023).
\newblock ``Design across multi-scale datasets by developing a novel approach to 3dgans.''\ {\em International Journal of Architectural Computing}, 21(2), 358--373.

\bibitem[\protect\citeauthoryear{}{Feng et~al.\@}{2020}]{feng-etal-2020-codebert}
Feng, Z., Guo, D., Tang, D., Duan, N., Feng, X., Gong, M., Shou, L., Qin, B., Liu, T., Jiang, D., and Zhou, M. (2020).
\newblock ``{C}ode{BERT}: A pre-trained model for programming and natural languages.''\ {\em Findings of the Association for Computational Linguistics: EMNLP 2020}, T. Cohn, Y. He, and Y. Liu, eds., Online, Association for Computational Linguistics,  1536--1547, $<$https://aclanthology.org/2020.findings-emnlp.139/$>$\ (November).

\bibitem[\protect\citeauthoryear{}{Fernandes et~al.\@}{2024}]{buildings14082499}
Fernandes, D., Garg, S., Nikkel, M., and Guven, G. (2024).
\newblock ``A gpt-powered assistant for real-time interaction with building information models.''\ {\em Buildings}, 14(8).

\bibitem[\protect\citeauthoryear{}{Fischer et~al.\@}{2007}]{4299919}
Fischer, G., Lusiardi, J., and Wolff~von Gudenberg, J. (2007).
\newblock ``Abstract syntax trees - and their role in model driven software development.''\ {\em International Conference on Software Engineering Advances (ICSEA 2007)},  38--38.

\bibitem[\protect\citeauthoryear{}{Fuchs et~al.\@}{2022}]{Fuchs2022}
Fuchs, S., Witbrock, M., Dimyadi, J., and Amor, R. (2022).
\newblock ``Neural semantic parsing of building regulations for compliance checking.''\ {\em IOP Conference Series: Earth and Environmental Science}, 1101, 092022.

\bibitem[\protect\citeauthoryear{}{Fuchs et~al.\@}{2024}]{fuchs2024usinglargelanguagemodels}
Fuchs, S., Witbrock, M., Dimyadi, J., and Amor, R. (2024).
\newblock ``Using large language models for the interpretation of building regulations, $<$https://arxiv.org/abs/2407.21060$>$.

\bibitem[\protect\citeauthoryear{}{Gade and Svidt}{2021}]{Gade2021}
Gade, P.~N. and Svidt, K. (2021).
\newblock ``Exploration of practitioner experiences of flexibility and transparency to improve bim-based model checking systems.''\ {\em ITcon}, 26, 1041--1060.

\bibitem[\protect\citeauthoryear{}{Gemini}{2024}]{gemini_function_calling}
Gemini (2024).
\newblock ``Intro to function calling with gemini api, $<$https://ai.google.dev/gemini-api/docs/function-calling$>$.
\newblock Accessed: 2024-07-16.

\bibitem[\protect\citeauthoryear{}{Graphisoft}{2024}]{Archicad:2024}
Graphisoft (2024).
\newblock ``Archicad ai visualizer, $<$https://graphisoft.com/solutions/innovation/archicad-ai-visualizer$>$.
\newblock accessed June 27, 2024.

\bibitem[\protect\citeauthoryear{}{Guo et~al.\@}{2024}]{guo2024largelanguagemodelbased}
Guo, T., Chen, X., Wang, Y., Chang, R., Pei, S., Chawla, N.~V., Wiest, O., and Zhang, X. (2024).
\newblock ``Large language model based multi-agents: A survey of progress and challenges, $<$https://arxiv.org/abs/2402.01680$>$.

\bibitem[\protect\citeauthoryear{}{Harode et~al.\@}{2022}]{doi:10.1061/9780784483961.071}
Harode, A., Thabet, W., and Gao, X. (2022).
\newblock {\em An Integrated Supervised Reinforcement Machine Learning Approach for Automated Clash Resolution}.
\newblock  679--688.

\bibitem[\protect\citeauthoryear{}{Harode et~al.\@}{2024}]{doi:10.1061/JCCEE5.CPENG-5548}
Harode, A., Thabet, W., and Gao, X. (2024).
\newblock ``Developing a machine-learning model to predict clash resolution options.''\ {\em Journal of Computing in Civil Engineering}, 38(2), 04024005.

\bibitem[\protect\citeauthoryear{}{He et~al.\@}{2023}]{He2023generative}
He, Z., Wang, Y.-H., and Zhang, J. (2023).
\newblock ``Generative structural design integrating bim and diffusion model, $<$https://synthical.com/article/bb21e837-1ed0-4489-8a33-768e6d0882fb$>$\ (10).

\bibitem[\protect\citeauthoryear{}{Hong et~al.\@}{2023}]{hong2023metagptmetaprogrammingmultiagent}
Hong, S., Zhuge, M., Chen, J., Zheng, X., Cheng, Y., Zhang, C., Wang, J., Wang, Z., Yau, S. K.~S., Lin, Z., Zhou, L., Ran, C., Xiao, L., Wu, C., and Schmidhuber, J. (2023).
\newblock ``Metagpt: Meta programming for a multi-agent collaborative framework, $<$https://arxiv.org/abs/2308.00352$>$.

\bibitem[\protect\citeauthoryear{}{Hu et~al.\@}{2024}]{hu2024scenecraftllmagentsynthesizing}
Hu, Z., Iscen, A., Jain, A., Kipf, T., Yue, Y., Ross, D.~A., Schmid, C., and Fathi, A. (2024).
\newblock ``Scenecraft: An llm agent for synthesizing 3d scene as blender code, $<$https://arxiv.org/abs/2403.01248$>$.

\bibitem[\protect\citeauthoryear{}{Häußler et~al.\@}{2021}]{Haeussler2020}
Häußler, M., Esser, S., and Borrmann, A. (2021).
\newblock ``{Code compliance checking of railway designs by integrating BIM, BPMN and DMN}.''\ {\em Automation in Construction}, 121, 103427.

\bibitem[\protect\citeauthoryear{}{ISO}{2024}]{ISO16739-1:2024}
ISO (2024).
\newblock ``{ISO 16739-1:2024: Industry Foundation Classes (IFC) for data sharing in the construction and facility management industries - Part 1: Data schema}, $<$https://www.iso.org/standard/84123.html$>$.
\newblock https://www.iso.org/standard/84123.html (last access: 2024-08-08).

\bibitem[\protect\citeauthoryear{}{Jang et~al.\@}{2024}]{JANG2024102532}
Jang, S., Lee, G., Oh, J., Lee, J., and Koo, B. (2024).
\newblock ``Automated detailing of exterior walls using {NADIA}: Natural-language-based architectural detailing through interaction with {AI}.''\ {\em Advanced Engineering Informatics}, 61, 102532.

\bibitem[\protect\citeauthoryear{}{Jiang et~al.\@}{2024}]{JIANG2024123431}
Jiang, G., Ma, Z., Zhang, L., and Chen, J. (2024).
\newblock ``Eplus-llm: A large language model-based computing platform for automated building energy modeling.''\ {\em Applied Energy}, 367, 123431.

\bibitem[\protect\citeauthoryear{}{Lewis et~al.\@}{2020}]{10.5555/3495724.3496517}
Lewis, P., Perez, E., Piktus, A., Petroni, F., Karpukhin, V., Goyal, N., K\"{u}ttler, H., Lewis, M., Yih, W.-t., Rockt\"{a}schel, T., Riedel, S., and Kiela, D. (2020).
\newblock ``Retrieval-augmented generation for knowledge-intensive {NLP} tasks.''\ {\em Proceedings of the 34th International Conference on Neural Information Processing Systems}, NIPS '20, Red Hook, NY, USA, Curran Associates Inc.

\bibitem[\protect\citeauthoryear{}{Li et~al.\@}{2024a}]{li2024generativeaiarchitecturaldesign}
Li, C., Zhang, T., Du, X., Zhang, Y., and Xie, H. (2024a).
\newblock ``Generative ai for architectural design: A literature review, $<$https://arxiv.org/abs/2404.01335$>$.

\bibitem[\protect\citeauthoryear{}{Li et~al.\@}{2024b}]{li2024advancing}
Li, F., Hogg, D.~C., and Cohn, A.~G. (2024b).
\newblock ``Advancing spatial reasoning in large language models: An in-depth evaluation and enhancement using the stepgame benchmark.''\ {\em Proceedings of the AAAI Conference on Artificial Intelligence}, Vol.~38,  18500--18507.

\bibitem[\protect\citeauthoryear{}{Li et~al.\@}{2024c}]{li2024sketchtoarchitecturegenerativeaiaidedarchitectural}
Li, P., Li, B., and Li, Z. (2024c).
\newblock ``Sketch-to-architecture: Generative ai-aided architectural design, $<$https://arxiv.org/abs/2403.20186$>$.

\bibitem[\protect\citeauthoryear{}{Liao et~al.\@}{2024}]{LIAO2024105187}
Liao, W., Lu, X., Fei, Y., Gu, Y., and Huang, Y. (2024).
\newblock ``Generative ai design for building structures.''\ {\em Automation in Construction}, 157, 105187.

\bibitem[\protect\citeauthoryear{}{Lin et~al.\@}{2023}]{lin2023magic3d}
Lin, C.-H., Gao, J., Tang, L., Takikawa, T., Zeng, X., Huang, X., Kreis, K., Fidler, S., Liu, M.-Y., and Lin, T.-Y. (2023).
\newblock ``Magic3d: High-resolution text-to-3d content creation.''\ {\em IEEE Conference on Computer Vision and Pattern Recognition ({CVPR})}.

\bibitem[\protect\citeauthoryear{}{Liu et~al.\@}{2023}]{liu2023lostmiddlelanguagemodels}
Liu, N.~F., Lin, K., Hewitt, J., Paranjape, A., Bevilacqua, M., Petroni, F., and Liang, P. (2023).
\newblock ``Lost in the middle: How language models use long contexts.''\ {\em arXiv preprint arXiv:2307.03172}, $<$https://arxiv.org/abs/2307.03172$>$.

\bibitem[\protect\citeauthoryear{}{Luo and Huang}{2022}]{LUO2022104470}
Luo, Z. and Huang, W. (2022).
\newblock ``Floorplangan: Vector residential floorplan adversarial generation.''\ {\em Automation in Construction}, 142, 104470.

\bibitem[\protect\citeauthoryear{}{Marvin et~al.\@}{2024}]{10.1007/978-981-99-7962-2_30}
Marvin, G., Hellen, N., Jjingo, D., and Nakatumba-Nabende, J. (2024).
\newblock ``Prompt engineering in large language models.''\ {\em Data Intelligence and Cognitive Informatics}, I.~J. Jacob, S. Piramuthu, and P. Falkowski-Gilski, eds., Singapore, Springer Nature Singapore,  387--402.

\bibitem[\protect\citeauthoryear{}{Mehta et~al.\@}{2024}]{mehta-etal-2024-improving}
Mehta, N., Teruel, M., Deng, X., Figueroa~Sanz, S., Awadallah, A., and Kiseleva, J. (2024).
\newblock ``Improving grounded language understanding in a collaborative environment by interacting with agents through help feedback.''\ {\em Findings of the Association for Computational Linguistics: EACL 2024}, Y. Graham and M. Purver, eds., St. Julian{'}s, Malta, Association for Computational Linguistics,  1306--1321, $<$https://aclanthology.org/2024.findings-eacl.87$>$\ (March).

\bibitem[\protect\citeauthoryear{}{Mescheder et~al.\@}{2019}]{8953655}
Mescheder, L., Oechsle, M., Niemeyer, M., Nowozin, S., and Geiger, A. (2019).
\newblock ``Occupancy networks: Learning 3d reconstruction in function space.''\ {\em 2019 IEEE/CVF Conference on Computer Vision and Pattern Recognition (CVPR)},  4455--4465.

\bibitem[\protect\citeauthoryear{}{Nuyts et~al.\@}{2024}]{Nuyts2024}
Nuyts, E., Bonduel, M., and Verstraeten, R. (2024).
\newblock ``Comparative analysis of approaches for automated compliance checking of construction data.''\ {\em Advanced Engineering Informatics}, 60, 102443.

\bibitem[\protect\citeauthoryear{}{Oleynikova et~al.\@}{2016}]{Oleynikova2016SignedDF}
Oleynikova, H., Millane, A., Taylor, Z., Galceran, E., Nieto, J.~I., and Siegwart, R.~Y. (2016).
\newblock ``Signed distance fields: A natural representation for both mapping and planning, $<$https://api.semanticscholar.org/CorpusID:28083959$>$.

\bibitem[\protect\citeauthoryear{}{OpenAI}{2024a}]{openai2024gpt4technicalreport}
OpenAI (2024a).
\newblock ``Gpt-4 technical report.''\ {\em arXiv preprint arXiv:2303.08774}, $<$https://arxiv.org/abs/2303.08774$>$.

\bibitem[\protect\citeauthoryear{}{OpenAI}{2024b}]{OpenAI_Function_Calling}
OpenAI (2024b).
\newblock ``Openai function calling guide, $<$https://platform.openai.com/docs/guides/function-calling$>$.
\newblock Accessed: 2024-07-16.

\bibitem[\protect\citeauthoryear{}{OpenAI}{2025}]{openai2025o3o4mini}
OpenAI (2025).
\newblock ``Introducing openai o3 and o4-mini, $<$https://openai.com/index/introducing-o3-and-o4-mini/$>$\ (April).
\newblock Accessed: 2025-05-29.

\bibitem[\protect\citeauthoryear{}{Pan et~al.\@}{2024}]{pan-etal-2024-automatically}
Pan, L., Saxon, M., Xu, W., Nathani, D., Wang, X., and Wang, W.~Y. (2024).
\newblock ``Automatically correcting large language models: Surveying the landscape of diverse automated correction strategies.''\ {\em Transactions of the Association for Computational Linguistics}, 12, 484--506.

\bibitem[\protect\citeauthoryear{}{Park et~al.\@}{2023}]{park2023generativeagentsinteractivesimulacra}
Park, J.~S., O'Brien, J.~C., Cai, C.~J., Morris, M.~R., Liang, P., and Bernstein, M.~S. (2023).
\newblock ``Generative agents: Interactive simulacra of human behavior, $<$https://arxiv.org/abs/2304.03442$>$.

\bibitem[\protect\citeauthoryear{}{Pauwels et~al.\@}{2011}]{Pauwels2011}
Pauwels, P., Deursen, D.~V., Verstraeten, R., Roo, J.~D., Meyer, R.~D., de~Walle, R.~V., and Campenhout, J.~V. (2011).
\newblock ``A semantic rule checking environment for building performance checking.''\ {\em Automation in Construction}, 20, 506--518.

\bibitem[\protect\citeauthoryear{}{Poole et~al.\@}{2022}]{Poole2022DreamFusionTU}
Poole, B., Jain, A., Barron, J.~T., and Mildenhall, B. (2022).
\newblock ``Dreamfusion: Text-to-3d using 2d diffusion.''\ {\em ArXiv}, abs/2209.14988.

\bibitem[\protect\citeauthoryear{}{Pouliou et~al.\@}{2023}]{Panagiota}
Pouliou, P., Horvath, A.-S., and Palamas, G. (2023).
\newblock ``Speculative hybrids: Investigating the generation of conceptual architectural forms through the use of 3d generative adversarial networks.''\ {\em International Journal of Architectural Computing}, 21(2), 315--336.

\bibitem[\protect\citeauthoryear{}{Preidel and Borrmann}{2018}]{Preidel2018}
Preidel, C. and Borrmann, A. (2018).
\newblock ``{BIM-Based Code Compliance Checking}.''\ {\em Building Information Modeling}, Springer International Publishing,  367--381.

\bibitem[\protect\citeauthoryear{}{Radford et~al.\@}{2022}]{radford2022robustspeechrecognitionlargescale}
Radford, A., Kim, J.~W., Xu, T., Brockman, G., McLeavey, C., and Sutskever, I. (2022).
\newblock ``Robust speech recognition via large-scale weak supervision, $<$https://arxiv.org/abs/2212.04356$>$.

\bibitem[\protect\citeauthoryear{}{Radford et~al.\@}{2015}]{Radford2015UnsupervisedRL}
Radford, A., Metz, L., and Chintala, S. (2015).
\newblock ``Unsupervised representation learning with deep convolutional generative adversarial networks.''\ {\em CoRR}, abs/1511.06434.

\bibitem[\protect\citeauthoryear{}{Reynolds and McDonell}{2021}]{10.1145/3411763.3451760}
Reynolds, L. and McDonell, K. (2021).
\newblock ``Prompt programming for large language models: Beyond the few-shot paradigm.''\ {\em Extended Abstracts of the 2021 CHI Conference on Human Factors in Computing Systems}, CHI EA '21, New York, NY, USA, Association for Computing Machinery, $<$https://doi.org/10.1145/3411763.3451760$>$.

\bibitem[\protect\citeauthoryear{}{Salewski et~al.\@}{2023}]{10.5555/3666122.3669274}
Salewski, L., Alaniz, S., Rio-Torto, I., Schulz, E., and Akata, Z. (2023).
\newblock ``In-context impersonation reveals large language models' strengths and biases.''\ {\em Proceedings of the 37th International Conference on Neural Information Processing Systems}, NIPS '23, Red Hook, NY, USA, Curran Associates Inc.

\bibitem[\protect\citeauthoryear{}{Schulhoff et~al.\@}{2024}]{schulhoff2024promptreportsystematicsurvey}
Schulhoff, S., Ilie, M., Balepur, N., Kahadze, K., Liu, A., Si, C., Li, Y., Gupta, A., Han, H., Schulhoff, S., Dulepet, P.~S., Vidyadhara, S., Ki, D., Agrawal, S., Pham, C., Kroiz, G., Li, F., Tao, H., Srivastava, A., Costa, H.~D., Gupta, S., Rogers, M.~L., Goncearenco, I., Sarli, G., Galynker, I., Peskoff, D., Carpuat, M., White, J., Anadkat, S., Hoyle, A., and Resnik, P. (2024).
\newblock ``The prompt report: A systematic survey of prompting techniques, $<$https://arxiv.org/abs/2406.06608$>$.

\bibitem[\protect\citeauthoryear{}{Shabani et~al.\@}{2023}]{10204517}
Shabani, M.~A., Hosseini, S., and Furukawa, Y. (2023).
\newblock ``Housediffusion: Vector floorplan generation via a diffusion model with discrete and continuous denoising.''\ {\em 2023 IEEE/CVF Conference on Computer Vision and Pattern Recognition (CVPR)},  5466--5475.

\bibitem[\protect\citeauthoryear{}{Shinn et~al.\@}{2023}]{shinn2023reflexionlanguageagentsverbal}
Shinn, N., Cassano, F., Berman, E., Gopinath, A., Narasimhan, K., and Yao, S. (2023).
\newblock ``Reflexion: Language agents with verbal reinforcement learning, $<$https://arxiv.org/abs/2303.11366$>$.

\bibitem[\protect\citeauthoryear{}{Solihin and Eastman}{2015}]{Solihin2015}
Solihin, W. and Eastman, C. (2015).
\newblock ``Classification of rules for automated {BIM} rule checking development.''\ {\em Automation in Construction}, 53, 69--82.

\bibitem[\protect\citeauthoryear{}{Stigsen et~al.\@}{2023}]{inproceedings}
Stigsen, M., Moisi, A., Rasoulzadeh, S., Schinegger, K., and Rutzinger, S. (2023).
\newblock ``Ai diffusion as design vocabulary - investigating the use of ai image generation in early architectural design and education.''\  587--596\ (01).

\bibitem[\protect\citeauthoryear{}{Sun et~al.\@}{2024}]{sun20243dgptprocedural3dmodeling}
Sun, C., Han, J., Deng, W., Wang, X., Qin, Z., and Gould, S. (2024).
\newblock ``3d-gpt: Procedural 3d modeling with large language models, $<$https://arxiv.org/abs/2310.12945$>$.

\bibitem[\protect\citeauthoryear{}{Sun et~al.\@}{2022}]{SUN2022108781}
Sun, C., Zhou, Y., and Han, Y. (2022).
\newblock ``Automatic generation of architecture facade for historical urban renovation using generative adversarial network.''\ {\em Building and Environment}, 212, 108781.

\bibitem[\protect\citeauthoryear{}{Sydora and Stroulia}{2020}]{Sydora2020}
Sydora, C. and Stroulia, E. (2020).
\newblock ``Rule-based compliance checking and generative design for building interiors using {BIM}.''\ {\em Automation in Construction}, 120, 103368.

\bibitem[\protect\citeauthoryear{}{Tomczak et~al.\@}{2022}]{Tomczak2022}
Tomczak, A., v~Berlo, L., Krijnen, T., Borrmann, A., and Bolpagni, M. (2022).
\newblock ``A review of methods to specify information requirements in digital construction projects.''\ {\em IOP Conference Series: Earth and Environmental Science}, 1101, 092024.

\bibitem[\protect\citeauthoryear{}{Tono and Fischer}{2022}]{tono22vitruvio}
Tono, A. and Fischer, M. (2022).
\newblock ``Vitruvio: 3d building meshes via single perspective sketches, $<$https://arxiv.org/abs/2210.13634$>$.

\bibitem[\protect\citeauthoryear{}{Vectorworks}{2024}]{SDKExamples}
Vectorworks, D. (2024).
\newblock ``Sdk examples, $<$https://github.com/VectorworksDeveloper/SDKExamples$>$.
\newblock Accessed: 2024-07-16.

\bibitem[\protect\citeauthoryear{}{vom Brocke et~al.\@}{2020}]{vomBrocke2020}
vom Brocke, J., Hevner, A., and Maedche, A. (2020).
\newblock {\em Introduction to Design Science Research}.
\newblock Springer International Publishing, Cham,  1--13.

\bibitem[\protect\citeauthoryear{}{Wang et~al.\@}{2024a}]{Wang_2024}
Wang, L., Ma, C., Feng, X., Zhang, Z., Yang, H., Zhang, J., Chen, Z., Tang, J., Chen, X., Lin, Y., Zhao, W.~X., Wei, Z., and Wen, J. (2024a).
\newblock ``A survey on large language model based autonomous agents.''\ {\em Frontiers of Computer Science}, 18(6).

\bibitem[\protect\citeauthoryear{}{Wang et~al.\@}{2021}]{article}
Wang, S., Zeng, W., Chen, X., Ye, Y., Qiao, Y., and Fu, C.-W. (2021).
\newblock ``Actfloor-gan: Activity-guided adversarial networks for human-centric floorplan design.''\ {\em IEEE Transactions on Visualization and Computer Graphics}, PP, 1--1.

\bibitem[\protect\citeauthoryear{}{Wang et~al.\@}{2024b}]{doi:10.1061/JCCEE5.CPENG-5487}
Wang, Z., Sacks, R., Ouyang, B., Ying, H., and Borrmann, A. (2024b).
\newblock ``A framework for generic semantic enrichment of bim models.''\ {\em Journal of Computing in Civil Engineering}, 38(1), 04023038.

\bibitem[\protect\citeauthoryear{}{Wei et~al.\@}{2022a}]{wei2022finetunedlanguagemodelszeroshot}
Wei, J., Bosma, M., Zhao, V.~Y., Guu, K., Yu, A.~W., Lester, B., Du, N., Dai, A.~M., and Le, Q.~V. (2022a).
\newblock ``Finetuned language models are zero-shot learners, $<$https://arxiv.org/abs/2109.01652$>$.

\bibitem[\protect\citeauthoryear{}{Wei et~al.\@}{2022b}]{wei2022emergentabilitieslargelanguage}
Wei, J., Tay, Y., Bommasani, R., Raffel, C., Zoph, B., Borgeaud, S., Yogatama, D., Bosma, M., Zhou, D., Metzler, D., Chi, E.~H., Hashimoto, T., Vinyals, O., Liang, P., Dean, J., and Fedus, W. (2022b).
\newblock ``Emergent abilities of large language models, $<$https://arxiv.org/abs/2206.07682$>$.

\bibitem[\protect\citeauthoryear{}{Wei et~al.\@}{2024}]{10.5555/3600270.3602070}
Wei, J., Wang, X., Schuurmans, D., Bosma, M., Ichter, B., Xia, F., Chi, E.~H., Le, Q.~V., and Zhou, D. (2024).
\newblock ``Chain-of-thought prompting elicits reasoning in large language models.''\ {\em Proceedings of the 36th International Conference on Neural Information Processing Systems}, NIPS '22, Red Hook, NY, USA, Curran Associates Inc.

\bibitem[\protect\citeauthoryear{}{Wu et~al.\@}{2023a}]{Wu:2023:EGICE}
Wu, J., Nousias, S., and Borrmann, A. (2023a).
\newblock ``Parametrization-based solution space exploration for model healing.''\ {\em Proc. of the 30th Int. Conference on Intelligent Computing in Engineering (EG-ICE)}\ (Jul).

\bibitem[\protect\citeauthoryear{}{Wu et~al.\@}{2023b}]{wu2023autogenenablingnextgenllm}
Wu, Q., Bansal, G., Zhang, J., Wu, Y., Li, B., Zhu, E., Jiang, L., Zhang, X., Zhang, S., Liu, J., Awadallah, A.~H., White, R.~W., Burger, D., and Wang, C. (2023b).
\newblock ``Autogen: Enabling next-gen llm applications via multi-agent conversation.''\ {\em arXiv preprint arXiv:2308.08155}, $<$https://arxiv.org/abs/2308.08155$>$.

\bibitem[\protect\citeauthoryear{}{Xu et~al.\@}{2024}]{xu2024surveygameplayingagents}
Xu, X., Wang, Y., Xu, C., Ding, Z., Jiang, J., Ding, Z., and Karlsson, B.~F. (2024).
\newblock ``A survey on game playing agents and large models: Methods, applications, and challenges.''\ {\em arXiv preprint arXiv:2403.10249}, $<$https://arxiv.org/abs/2403.10249$>$.

\bibitem[\protect\citeauthoryear{}{Yang and Zhang}{2024}]{YANG2024105817}
Yang, F. and Zhang, J. (2024).
\newblock ``Prompt-based automation of building code information transformation for compliance checking.''\ {\em Automation in Construction}, 168, 105817.

\bibitem[\protect\citeauthoryear{}{Yang et~al.\@}{2021}]{Yang_Wu_Zhang_Jin_2021}
Yang, X., Wu, Y., Zhang, K., and Jin, C. (2021).
\newblock ``Cpcgan: A controllable 3d point cloud generative adversarial network with semantic label generating.''\ {\em Proceedings of the AAAI Conference on Artificial Intelligence}, 35(4), 3154--3162.

\bibitem[\protect\citeauthoryear{}{Zhang et~al.\@}{2024a}]{zhang2024restmctsllmselftrainingprocess}
Zhang, D., Zhoubian, S., Hu, Z., Yue, Y., Dong, Y., and Tang, J. (2024a).
\newblock ``Rest-mcts*: Llm self-training via process reward guided tree search, $<$https://arxiv.org/abs/2406.03816$>$.

\bibitem[\protect\citeauthoryear{}{Zhang and El-Gohary}{2017}]{Zhang2017}
Zhang, J. and El-Gohary, N.~M. (2017).
\newblock ``Integrating semantic nlp and logic reasoning into a unified system for fully-automated code checking.''\ {\em Automation in Construction}, 73, 45--57.

\bibitem[\protect\citeauthoryear{}{Zhang et~al.\@}{2024b}]{ZHANG2024114788}
Zhang, L., Chen, Z., and Ford, V. (2024b).
\newblock ``Advancing building energy modeling with large language models: Exploration and case studies.''\ {\em Energy and Buildings}, 323, 114788.

\bibitem[\protect\citeauthoryear{}{Zhang et~al.\@}{2022}]{Zhanglei:2022}
Zhang, L., Zheng, L., Chen, Y., Huang, L., and Zhou, S. (2022).
\newblock ``Cgan-assisted renovation of the styles and features of street facades—a case study of the wuyi area in fujian, china.''\ {\em Sustainability}, 14, 16575.

\bibitem[\protect\citeauthoryear{}{Zhang et~al.\@}{2023}]{zhang2023automatic}
Zhang, Z., Zhang, A., Li, M., and Smola, A. (2023).
\newblock ``Automatic chain of thought prompting in large language models.''\ {\em The Eleventh International Conference on Learning Representations (ICLR 2023)}.

\bibitem[\protect\citeauthoryear{}{Zheng and Fischer}{2023}]{ZHENG2023105067}
Zheng, J. and Fischer, M. (2023).
\newblock ``Dynamic prompt-based virtual assistant framework for bim information search.''\ {\em Automation in Construction}, 155, 105067.

\bibitem[\protect\citeauthoryear{}{Zhou et~al.\@}{2023}]{zhou-etal-2023-codebertscore}
Zhou, S., Alon, U., Agarwal, S., and Neubig, G. (2023).
\newblock ``{C}ode{BERTS}core: Evaluating code generation with pretrained models of code.''\ {\em Proceedings of the 2023 Conference on Empirical Methods in Natural Language Processing}, H. Bouamor, J. Pino, and K. Bali, eds., Singapore, Association for Computational Linguistics,  13921--13937, $<$https://aclanthology.org/2023.emnlp-main.859/$>$\ (December).

\bibitem[\protect\citeauthoryear{}{Zhou et~al.\@}{2022}]{Zhou2022}
Zhou, Y.~C., Zheng, Z., Lin, J.~R., and Lu, X.~Z. (2022).
\newblock ``Integrating {NLP} and context-free grammar for complex rule interpretation towards automated compliance checking.''\ {\em Computers in Industry}, 142.

\bibitem[\protect\citeauthoryear{}{Zhu et~al.\@}{2024}]{zhu2024llava}
Zhu, C., Wang, T., Zhang, W., Pang, J., and Liu, X. (2024).
\newblock ``Llava-3d: A simple yet effective pathway to empowering lmms with 3d-awareness.''\ {\em arXiv preprint arXiv:2409.18125}.

\bibitem[\protect\citeauthoryear{}{Zhuang et~al.\@}{2023}]{XinweiZhuang:2023}
Zhuang, X., Ju, Y., Yang, A., and Caldas, L. (2023).
\newblock ``Synthesis and generation for 3d architecture volume with generative modeling.''\ {\em International Journal of Architectural Computing}, 21(2), 297--314.

\bibitem[\protect\citeauthoryear{}{Çelen et~al.\@}{2024}]{çelen2024idesignpersonalizedllminterior}
Çelen, A., Han, G., Schindler, K., Gool, L.~V., Armeni, I., Obukhov, A., and Wang, X. (2024).
\newblock ``{I-Design: Personalized LLM Interior Designer}, $<$https://arxiv.org/abs/2404.02838$>$.

\end{thebibliography}

\end{document}